\documentclass[10pt, hidelinks]{article} 

\usepackage[colorlinks,linkcolor=blue,citecolor=magenta]{hyperref}
\usepackage{url}
\usepackage{fullpage}
\usepackage[pdftex,dvipsnames]{xcolor}
\usepackage{enumitem, comment, xifthen}
\usepackage{graphicx}
\usepackage{booktabs}       
\usepackage{nicefrac}       
\usepackage{microtype}      

\usepackage{subfigure}
\usepackage{amsmath,amssymb,amsthm, amsfonts}
\usepackage[T1]{fontenc}
\usepackage[utf8]{inputenc}
\usepackage{mathtools}
\usepackage[english]{babel}
\usepackage{mwe}
\usepackage{tcolorbox}

\usepackage[toc,page,header]{appendix}
\usepackage{algpseudocode}
\usepackage{algorithm}
\usepackage[ruled,algo2e]{algorithm2e}
\usepackage{setspace}
\usepackage{framed}
\usepackage{mdframed}
\usepackage{cite}
\usepackage{framed}
\usepackage{mdframed}
\usepackage{float}
\usepackage{wrapfig}

\usepackage{blindtext}
\usepackage{enumitem}
\usepackage{amssymb}
\usepackage{pifont}
\usepackage{xargs}    
\usepackage{soul}
\usepackage{scrwfile}
\usepackage[stable]{footmisc}
\usepackage{multirow}
\usepackage{makecell}

\newtheorem{theorem}{Theorem}[section]
\newtheorem{lemma}[theorem]{Lemma}
\newtheorem{remark}{Remark}
\newtheorem{assumption}{Assumption}

\newtheorem{corollary}[theorem]{Corollary}

\allowdisplaybreaks

\DeclarePairedDelimiter{\floor}{\lfloor}{\rfloor}

\newcommand{\pl}{Polyak-\L{}ojasiewicz}

\def\@fnsymbol#1{\ensuremath{\ifcase#1\or *\or \dagger\or \ddagger\or
   \mathsection\or \mathparagraph\or \|\or **\or \dagger\dagger
   \or \ddagger\ddagger \else\@ctrerr\fi}}
\newcommand{\ssymbol}[1]{^{\@fnsymbol{#1}}}

\definecolor{my-green}{cmyk}{0.2, 0.04, 0.1, 0.04, 0.8}
\definecolor{classicrose}{rgb}{0.98, 0.8, 0.91}

\newcommand\mycommfont[1]{\footnotesize\ttfamily\textcolor{blue}{#1}}
\newcommand{\algcolor}[3]{\hspace*{-\fboxsep}\colorbox{#1}{\parbox{#2\linewidth}{#3}}}
\newcommand{\algemph}[3]{\algcolor{#1}{#2}{#3}}
\definecolor{blizzardblue}{rgb}{0.67, 0.9, 0.93}

\newcommand{\upstairs}[1]{\textsuperscript{#1}}
\newcommand{\affilone}{\dag}
\newcommand{\affiltwo}{\ddag}

\TOCclone[Table of Contents]{toc}{atoc}
\newcommand\StartAppendixEntries{}
\AfterTOCHead[toc]{%
  \renewcommand\StartAppendixEntries{\value{tocdepth}=-10000\relax}%
}
\AfterTOCHead[atoc]{%
  \edef\maintocdepth{\the\value{tocdepth}}%
  \value{tocdepth}=-10000\relax%
  \renewcommand\StartAppendixEntries{\value{tocdepth}=\maintocdepth\relax}%
}
\newcommand*\appendixwithtoc{%
\hypersetup{linkcolor=black}
  \addtocontents{toc}{\protect\StartAppendixEntries}
  \listofatoc
}

\begin{document}

\begin{center}

  {\bf{\LARGE Federated Learning with Compression:\\[4pt]
Unified Analysis and Sharp Guarantees}} \\

  \vspace*{.2in}
  
  \begin{tabular}{cc}
    Farzin Haddadpour\upstairs{\affilone}
    ~~Mohammad Mahdi Kamani\upstairs{\affiltwo}
    ~~Aryan Mokhtari\upstairs{\S}
    ~~Mehrdad Mahdavi\upstairs{\affilone}
    \\[5pt]
    \\
    \upstairs{\affilone}School of  Electrical Engineering and Computer Science \\
    \upstairs{\affiltwo}College of Information Sciences and Technology \\
    The Pennsylvania State University\\
    \texttt{\{fxh18, mqk5591, mzm616\}@psu.edu}\\
    \\
    \upstairs{\S}Department of Electrical and Computer Engineering \\
    The University of Texas at Austin\\
   \texttt{mokhtari@austin.utexas.edu}
  \end{tabular}
  
  \vspace*{.2in}
\begin{abstract}
In federated learning, communication cost is often a critical bottleneck to scale up distributed optimization algorithms to collaboratively learn a model from millions of devices with potentially unreliable or limited communication and  heterogeneous data distributions. Two notable trends to deal with the communication overhead of federated algorithms are \emph{gradient compression} and \emph{local computation with periodic communication}. Despite many attempts, characterizing the relationship between these two approaches has proven elusive. We address this by proposing a set of algorithms with periodical compressed (quantized or sparsified) communication and analyze their convergence properties in both homogeneous and heterogeneous local data distributions settings. For the homogeneous setting, our analysis improves existing bounds by providing tighter convergence rates for both \emph{strongly convex} and \emph{non-convex} objective functions. To mitigate data heterogeneity, we introduce a \emph{local gradient tracking} scheme and obtain sharp convergence rates that match the best-known communication complexities without compression for convex, strongly convex, and nonconvex settings. We complement our theoretical results by demonstrating the effectiveness of our proposed methods on real-world datasets.
\end{abstract}

\end{center}

\section{Introduction}\label{sec:intro}
The primary obstacle towards scaling distributed optimization algorithms is the significant communication cost both in terms of the number of communication rounds and the amount of exchanged data per round. To significantly reduce the number of communication rounds, a practical solution is to trade-off local computation for less communication via periodic averaging~\cite{zhang2016parallel,stich2018local}. In particular, the local SGD algorithm~\cite{stich2018local,wang2018cooperative,yu2018parallel} alternates between a fixed number of local updates and one step of synchronization which is shown to enjoy the same convergence rate as its fully synchronous counterpart, while significantly reducing the number of communication rounds.

A fundamentally different solution to scale up distributed optimization algorithms is to reduce the size of the communicated message per communication round.  This problem is especially exacerbated in edge computing where the worker devices (e.g., smartphones or IoT devices) are remotely connected, and communication bandwidth and power resources are limited. For instance, ResNet~\cite{he2016deep} has more than 25 million parameters, so the communication cost of sending local models through a computer network could be prohibitive.  The current methodology towards reducing the size of messages is to communicate compressed local gradients or models to the central server by utilizing a quantization operator~\cite{alistarh2017qsgd,bernstein2018signsgd,tang2018communication,wu2018error,wen2017terngrad,suresh2017distributed,reisizadeh2019fedpaq}, sparsification schema~\cite{stich2018sparsified,LinHM0D18,wu2018error,alistarh2018convergence}, or composition of both~\cite{basu2019qsparse}. 


\begin{table}[t]
    \centering
    \resizebox{0.8\linewidth}{!}{
    \begin{tabular}{llll}
        \toprule
                    &  \multicolumn{3}{c}{Objective function}
        \\ \cmidrule(r){2-4}
        Reference        & Nonconvex      & PL/Strongly Convex                                  & General Convex  
        \\
        \midrule
        \makecell{QSPARSE\cite{basu2019qsparse}}  & \makecell[l]{$R\!=\!O\left(\frac{q+1}{\epsilon^{{3}/{2}}}\right)$ \\ $\tau\!=\!O\left(\frac{1}{m(q+1)\sqrt{\epsilon}}\right)$}   & \makecell[l]{$R=O\left({\color{black}\kappa}\frac{q+1}{\sqrt{\epsilon}}\right)$ \\ $\tau=O\left(\frac{1}{m\left(q+1\right)\sqrt{\epsilon}}\right)$}               & \makecell{$-$}  \\
        \midrule
        \makecell{FedPAQ\cite{reisizadeh2019fedpaq}}  & \makecell[l]{$R=O\left(\frac{1}{\epsilon}\right)$ \\ $\tau=O\left(\frac{(\frac{q}{m})^2+1}{\epsilon}\right)$}   & \makecell[l]{$R=O\left(m+\frac{q+1}{m\epsilon}\right)$ \\ $\tau=O\left(1\right)$}               & \makecell{$-$}   \\
        \midrule
       \makecell{\textbf{Theorem~\ref{thm:homog_case}}} & \makecell[l]{$\boldsymbol{R=O\left(\frac{1}{\epsilon}\right)}$ \\[3pt] $\boldsymbol{\tau=O\left(\frac{q+1}{m\epsilon}\right)}$}   & \makecell[l]{$\boldsymbol{R=O\left(\kappa\left(\frac{q}{m}+1\right)\log\left(\frac{1}{\epsilon}\right)\right)}$ \\[3pt] $\boldsymbol{\tau=O\left(\frac{\left(q+1\right)}{m\left(\frac{q}{m}+1\right)\epsilon}\right)}$}               & \makecell[l]{$\boldsymbol{R\!=\!O\left(\frac{1+\frac{q}{m}}{\epsilon}{\color{black}\log\left(\frac{1}{\epsilon}\right)}\right)}$\\[3pt]
       $\boldsymbol{\tau\!=\!O\left(\frac{\left(q+1\right)^2}{m\left(\frac{q}{m}+1\right)^2\epsilon^2}\right)}$}         
   \\
        \bottomrule
    \end{tabular}
    }
\caption{Comparison of results with compression and periodic averaging in the homogeneous setting. Here, $m$ is the number of devices, $q$ is compression distortion constant, $\kappa$ is condition number, $\epsilon$ is target accuracy, $R$ is  the number of communication rounds, and $\tau$ is the number of local updates. QSPARSE~\cite{basu2019qsparse} has the assumption of bounded gradient, while FedPAQ~\cite{reisizadeh2019fedpaq} and our proposed algorithm do not have such assumption.}
\label{table:1}
\vspace{-2mm}
\end{table}

Despite significant progress in improving both aspects of communication efficiency~\cite{karimireddy2019scaffold,stich2018sparsified,bernstein2018signsgd,stich2019error}, there still exists a huge gap in our understanding of these approaches in federated learning, in particular for the cases that both compression and periodic averaging techniques are applied simultaneously. In terms of reducing communication rounds, a few recent attempts were able to reduce the frequency of synchronizing locally evolving models~\cite{haddadpour2019local,khaled2020tighter}, which are not improvable in general~\cite{zhang2020fedpd}. This necessitates that further improvement in communication efficiency needs to be explored by reducing the size of communicated messages. We highlight that compressed communication is of further importance to accelerate training non-convex objectives as it requires significantly more communication rounds to converge compared to distributed convex optimization. Furthermore, most existing methods are analyzed for homogeneous data and our understanding of the efficiency of these methods in the heterogeneous case is lacking.

In light of the above issues, the key contribution of this paper is the introduction and analysis of simple variants of \textit{local SGD with compressed communication}P~\footnote{Based on the literature, noting  the algorithmic similarity of  Federated Averaging~\cite{mcmahan2016communication} and Local SGD~\cite{stich2018local},  the main differences between them are the participation of clients and heterogeneity of local data distributions. In local SGD it is usually assumed that all of the clients are involved in communication, whereas in federated averaging a randomly selected subset of clients participate at averaging. Also, federated averaging is commonly used to reflect the data heterogeneity, which is a key ingredient in our analysis as well. For simplicity, we do not differentiate between these two terms and use them interchangeably.} 
  without compromising the attainable guarantees. The proposed algorithmic ideas accommodate both homogeneous and heterogeneous data distribution settings with the obtained rates summarized in Table~\ref{table:1} and Table~\ref{table:2}, respectively. In the homogeneous case, with a tight analysis of a simple quantized variant of local SGD, we show that not only our proposed method improves the complexity bounds for algorithms with compression (Table~\ref{table:1}), but also outperforms the complexity bounds for non-compressed counterparts in terms of the number of communication rounds (Table~\ref{table:3}). In the heterogeneous case, we argue that in the presence of compression (quantization or sparsification), locally updating models via local gradient information could lead to a significant drift among local models, which shed light on designing a quantized variant of local SGD that tracks local gradient information at local devices. We show that this simple gradient tracking idea leads to a method that outperforms state-of-the-art methods with compression for the heterogeneous setting (Table~\ref{table:2}) and it can even compensate for the noise introduced by compression and lead to the best-known convergence rates for convex and non-convex settings under perfect communication, i.e., no compression (Table~\ref{table:4}). 

\begin{table}[t]
	\centering
	\resizebox{0.8\linewidth}{!}{
		\begin{tabular}{llll}
			\toprule
			&  \multicolumn{3}{c}{Objective function}
			\\ \cmidrule(r){2-4}
			Reference        & Nonconvex      & PL/Strongly Convex                                  & General Convex 	\\
			\midrule
			\makecell{QSPARSE\cite{basu2019qsparse}}  & \makecell[l]{$R=O\left(\frac{q+1}{\epsilon^{{3}/{2}}}\right)$ \\ $\tau=O\left(\frac{1}{m(q+1)\sqrt{\epsilon}}\right)$}   & \makecell[l]{$R=O\left({\color{black}\kappa}\frac{q+1}{\sqrt{\epsilon}}\right)$ \\ $\tau=O\left(\frac{1}{m\left(q+1\right)\sqrt{\epsilon}}\right)$}               & \makecell{$-$}
			\\
			\midrule
			\makecell{\textbf{Theorem~\ref{thm:non-homogen}} } & \makecell[l]{$\boldsymbol{R=O\left(\frac{q+1}{\epsilon}\right)}$ \\[3pt] $\boldsymbol{\tau=O\left(\frac{1}{m\epsilon}\right)}$}   & \makecell[l]{$\boldsymbol{R=O\left(\kappa\left({q}+1\right)\log\left(\frac{1}{\epsilon}\right)\right)}$ \\[3pt] $\boldsymbol{\tau=O\left(\frac{1}{m\epsilon}\right)}$}               & \makecell[l]{$\boldsymbol{R=O\left(\frac{1+{q}}{\epsilon}{\color{black}\log\left(\frac{1}{\epsilon}\right)}\right)}$\\[3pt]
				$\boldsymbol{\tau=O\left(\frac{1}{m\epsilon^2}\right)}$}    
			\\
			\bottomrule
		\end{tabular}
	}
	\caption{Comparison of results with compression and periodic averaging in the heterogeneous setting. QSPARSE~\cite{basu2019qsparse} has the assumption of bounded gradient, while our proposed algorithm does not.}
	\label{table:2}
	\vspace{-2mm}
\end{table}

\noindent\textbf{Contributions.} We summarize the main contributions of this paper below:
\begin{itemize}[leftmargin=*]
 \item \textit{Homogeneous local distributions:} To keep the analysis simple yet insightful, we start with a quantized variant of federated averaging algorithm and analyze its convergence for non-convex, strongly convex and general convex objectives. As demonstrated in Table~\ref{table:1}, the obtained rates  is novel for convex objectives to the best of our knowledge, and improves the best known bounds in~\cite{reisizadeh2019fedpaq} and ~\cite{basu2019qsparse} for general non-convex and strongly convex objectives, respectively.

\item \textit{Heterogeneous local distributions:} For the heterogeneous setting, we propose federated averaging with compression and local gradient tracking, dubbed as \texttt{FedCOMGATE} algorithm, and establish its convergence rates for general non-convex, strongly convex or PL, and convex objectives. The obtained rates improve upon the results reported in~\cite{basu2019qsparse} for general non-convex and strongly-convex objectives. The obtained rates for general convex functions are novel to the best of our knowledge. 

 \item We verify our theoretical results through various extensive experiments on different  real federated datasets that demonstrate the practical efficacy of our methods.
\end{itemize}

\section{Problem Setup}

In this paper we focus on a federated architecture, where $m$ users aim to learn a global model in a collaborative manner without exchanging their data points with each other. Moreover, we assume that users (computing units) can only exchange information via a central unit (server) which is connected to all users. The optimization problem that the users try to solve can be written as
\begin{equation}\label{eq:global-cost}
    \min_{\boldsymbol{w}\in \mathbb{R}^d}f(\boldsymbol{w})\triangleq \frac{1}{m}\sum_{j=1}^m f_j(\boldsymbol{w})
\end{equation}
where  $f_j: \mathbb{R}^d \to \mathbb{R}$ is the loss function corresponding to user $j$. We further assume that the local objective function of each user $j$ is the expected loss over the set of data points of node $j$, i.e., 
\begin{equation}
    f_j(\boldsymbol{w})=\mathbb{E}_{\boldsymbol{z}\sim\mathcal{P}_j}[\ell_j(\boldsymbol{w},\boldsymbol{z})],
\end{equation}
where $\boldsymbol{z}$ is a random variable with probability distribution $\mathcal{P}_j$ and the loss function $\ell_j$ measures how well the model performs. $\mathcal{P}_j$ can be considered as the underlying distribution of node $j$ for generating data points, and realizations of the random variable $\boldsymbol{z}$ are the data points of node $j$. For instance, in a supervised learning case each element sample point $\boldsymbol{z}_i$ corresponds to a pair of input (feature) vector $\boldsymbol{x}_i$ and its label $y_i$. In this case, $\ell_j(\boldsymbol{w},\boldsymbol{z}_i)=\ell_j(\boldsymbol{w},(\boldsymbol{x}_i,y_i))$ measures how well the model $\boldsymbol{w}$ performs in predicting the label of $\boldsymbol{x}_i$ which is $y_i$. Note that the probability distributions of users may not be necessarily identical. In fact, through the paper, we study two settings (i) homogeneous setting in which all the probability distributions and loss functions are identical, i.e.,  $(\mathcal{P}_1\!=\!\dots\!=\!\mathcal{P}_m)$ and $(\ell_1\!=\!\dots\!=\!\ell_m)$; and (ii) heterogeneous setting in which the users' distributions and loss functions could be different.
\section{Federated Averaging with Compression\footnote{Generalized Compressed Local SGD}}\label{sec:fplg}

In this section, we propose a generalized version of the local stochastic gradient descent (SGD) method for federated learning which uses compressed signals to reduce the overall communication overhead of solving problem~\eqref{eq:global-cost}. The proposed federated averaging with compression (\texttt{FedCOM}) is designed for homogeneous settings where the probability distributions and loss functions of the users are identical. \texttt{FedCOM} differs from standard local SGD methods \cite{stich2018local,yu2018parallel,wang2018cooperative} in two major aspects. First, it uses compressed messages for uplink communication. Second, at the central node, the new global model is a convex combination of the previous global model and the average of updated local models of users.  We show that \texttt{FedCOM}  converges faster than state-of-the-art methods in a homogeneous setting by periodic averaging, local and global learning rates, and compressed communications.

To formally present the steps of \texttt{FedCOM}, consider $R$ as the rounds of communication between server and users, and $\tau$ as the number of local updates performed between two consecutive communication rounds. Further,  define $\boldsymbol{w}^{(r)}$ as the model at the master at the $r$-th round of communication. At each round $r$, the server sends the global model $\boldsymbol{w}^{(r)}$ to the users (clients). Then, each user $j$ computes its local stochastic gradient and updates the model by following the update of SGD for $\tau$ iterations. Specifically, at communication round $r$, user $j$ follows the update
\begin{equation}
    \boldsymbol{w}^{(c+1,r)}_{j}=\boldsymbol{w}^{(c,r)}_j-\eta~ \tilde{\mathbf{g}}_{j}^{(c,r)},\qquad \text{for}\quad  c=0,\dots,\tau-1.
\end{equation}
Here, $\boldsymbol{w}^{(c,r)}_{j}$ is the model at node $j$ and round $r$ after $c$ local updates,
$\tilde{\mathbf{g}}_{j}^{(c,r)}:=\nabla f_j (\boldsymbol{w}^{(c,r)}_j;\mathcal{Z}_j^{(c,r)}):= \frac{1}{b_j}\sum_{\mathbf{z}\in \mathcal{Z}_j^{(c,r)}}\nabla \ell_j (\boldsymbol{w}^{(c,r)}_j,\mathbf{z})$ is a stochastic gradient of $f_j$ evaluated using the mini-batch $\mathcal{Z}_j^{(c,r)}:=\{\boldsymbol{z}_{j,1}^{(c,r)},\dots, \boldsymbol{z}_{j,b_j}^{(c,r)}\}$ of size $b_j$, and  $\eta$ is the learning rate. The output of this $\tau$ recursive updates for node $j$ at round $r$ is $\boldsymbol{w}^{(\tau,r)}_{j}$. After computing the local models, each user $j$ sends a compressed version of $(\boldsymbol{w}^{(\tau,r)}_{j}-\boldsymbol{w}^{(r)}_{j})/\eta$ to the central node by applying a compression operator $Q(\cdot)$. Note that the compressed signal $\boldsymbol{\Delta}_{j,q}^{(r)}\triangleq Q((\boldsymbol{w}^{(\tau,r)}_{j}-\boldsymbol{w}^{(r)}_{j})/\eta)$ indicates a normalized version of the difference between the input and output of the local SGD process at round $r$ at node $j$, which is equal to the aggregation of all local SGD directions, i.e., $(\boldsymbol{w}^{(\tau,r)}_{j}-\boldsymbol{w}^{(r)}_{j})/\eta=\sum_{c=0}^\tau \tilde{\mathbf{g}}_{j}^{(c,r)}$. 
Once, the server receives the compressed signals $\{\boldsymbol{\Delta}_{j,q}^{(r)}\}_{j=1}^{j=m}$, it computes the new global model according to 
\begin{equation}\label{eq:lfsgd-update}
  {\boldsymbol{w}}^{(r+1)}=\boldsymbol{w}^{(r)}-\frac{\eta\gamma}{m}\sum_{j=1}^m\boldsymbol{\Delta}_{j,q}^{(r)},
\end{equation}
where $\gamma$ is the global learning rate. The steps of the \texttt{FedCOM} algorithm are summarized in Algorithm~\ref{Alg:one-shot-using data samoples-b}.

\begin{algorithm2e}[t]
\DontPrintSemicolon
\SetNoFillComment
\setstretch{0.5}
\LinesNumbered
\caption{\texttt{FedCOM}($R$, $\tau, \eta, \gamma$)}\label{Alg:one-shot-using data samoples-b}
\SetKwFor{ForPar}{for}{do in parallel}{end forpar}
\textbf{Inputs:} Number of communication rounds $R$, number of local updates $\tau$, learning rates $\gamma$ and $\eta$, initial  global model $\boldsymbol{w}^{(0)}$\\[2pt]
\For{$r=0, \ldots, R-1$}{
    \ForPar{each client $j\in[m]$}{
        Set $\boldsymbol{w}_j^{(0,r)}={\boldsymbol{w}}^{(r)}$ \\
        \For{ $c=0,\ldots,\tau-1$}{
            Sample a minibatch $\mathcal{Z}_j^{(c,r)}$ and compute $\tilde{\mathbf{g}}_{j}^{(c,r)}\triangleq\nabla f_j (\boldsymbol{w}^{(c,r)}_j;\mathcal{Z}_j^{(c,r)})$\\
            $\boldsymbol{w}^{(c+1,r)}_{j}=\boldsymbol{w}^{(c,r)}_j-\eta~ \tilde{\mathbf{g}}_{j}^{(c,r)}$ \label{eq:update-rule-alg}
        }
        Device \textbf{sends} $  \boldsymbol{\Delta}_{j,q}^{(r)} = Q(({\boldsymbol{w}^{(r)}-\boldsymbol{w}^{(\tau,r)}_j})/\eta)$  back to the server\\
    }
    Server \textbf{computes} $\boldsymbol{\Delta}_q^{(r)}=\frac{1}{m}\sum_{j=1}^m \boldsymbol{\Delta}_{j,q}^{(r)}$\\
    Server \textbf{computes} $\boldsymbol{w}^{(r+1)}=\boldsymbol{w}^{(r)}-\eta\gamma\boldsymbol{\Delta}_q^{(r)}$ and \textbf{broadcasts} to all devices\\
}
\end{algorithm2e}

\begin{remark}
Note that by setting $\gamma=1$ in  \eqref{eq:lfsgd-update}, \texttt{FedCOM} boils down to the FedPAQ algorithm proposed in \cite{reisizadeh2019fedpaq}, and if we further remove the compression scheme then we recover  FedAvg \cite{wang2018cooperative}. Note that in both FedAvg and its vanilla quantized variant FedPAQ, the new global model is the average of local models (if we ignore the error of compression for FedPAQ), while in \texttt{FedCOM} the new global model is a linear combination of the previous global model and the average of updated local models, due to the extra parameter $\gamma$.
We show that by adding this modification and properly choosing $\gamma$, \texttt{FedCOM} improves the complexity bounds of FedPAQ for both strongly convex and non-convex settings. Note that the update in \eqref{eq:lfsgd-update} can also be interpreted as running a global SGD update on master's model by descending towards the average of aggregated local gradient directions with stepsize~$\eta \gamma$. Specifically, if we assume perfect communication (ignoring the quantization) then we obtain that the new global model is given by
${\boldsymbol{w}}^{(r+1)}={\boldsymbol{w}}^{(r)}-\eta\gamma\frac{1}{m}\sum_{j=1}^m \sum_{c=0}^\tau \tilde{\mathbf{g}}_{j}^{(c,r)}$. 
\end{remark}

\section{Compressed Local SGD with Local Gradient Tracking}
In the previous section, we introduced a relatively simple algorithm called \texttt{FedCOM}  for homogeneous settings, where the probability distributions of the users are identical. Although \texttt{FedCOM} both theoretically (Section~\ref{Sec:convergence}) and numerically (Section~\ref{sec:exp}) performs well for homogeneous settings, its performance is not satisfactory in heterogeneous settings where the probability distributions of users are different. This is due to the fact that the updates of  \texttt{FedCOM} heavily depend on the local SGD directions. In a homogeneous setting, following local gradient directions leads to a good global model as all samples are drawn from the same distribution and the local gradient direction is a good estimate of the global function gradient. However, in a heterogeneous setting, updating local models only based on local gradient information could lead to an arbitrary poor performance as the local gradient directions could be very different from the global gradient direction. 

To address this issue, in this section we propose a novel variant of federated averaging with compression and local gradient tracking (\texttt{FedCOMGATE}) for heterogeneous settings. The main difference between \texttt{FedCOM} and \texttt{FedCOMGATE} is the idea of local gradient tracking that ensures that each node uses an estimate of the global gradient direction to locally update its model. To estimate global gradient direction nodes also require access to the average of local models which means that in \texttt{FedCOMGATE} in addition to sending the global updates master also needs to broadcast the average of $\boldsymbol{\Delta}_{j,q}^{(r)}\triangleq Q((\boldsymbol{w}^{(\tau,r)}_{j}-\boldsymbol{w}^{(r)}_{j})/\eta)$, shown by $\boldsymbol{\Delta}_{q}^{(r)}=\frac{1}{m}\sum_{j=1}^m\boldsymbol{\Delta}_{j,q}^{(r)}$ to devices. 

To present \texttt{FedCOMGATE}, consider $\delta_j$ as a sequence at node $j$ that is designed to track the difference between the local gradient direction and the global gradient direction (the direction obtained by incorporating gradient information of all users). At round $r$, each worker $j$ updates its local sequence $\delta_j$ based on the update
\begin{equation}
    \delta_j^{(r+1)}=\delta_j^{(r)}+\frac{1}{\tau}\left(\boldsymbol{\Delta}_{j,q}^{(r)}-\boldsymbol{\Delta}_{q}^{(r)}\right),
\end{equation}
where $\boldsymbol{\Delta}_{j,q}^{(r)}$ is the quantized version of the accumulation of the gradients  at node $j$ from the previous round and $\boldsymbol{\Delta}_{q}^{(r)}$ is the average of $\boldsymbol{\Delta}_{j,q}^{(r)}$. Once the correction vector $\delta_j^{(r)}$ is computed, each node $j$ runs a corrected local update for $\tau$ rounds based on the update
\begin{align}
    \boldsymbol{w}^{(c+1,r)}_{j}=\boldsymbol{w}^{(c,r)}_j-\eta~ \tilde{\boldsymbol{d}}^{(c,r)}_{j,q}
    &=\boldsymbol{w}^{(c,r)}_j-\eta (\tilde{\mathbf{g}}_{j}^{(c,r)}-\delta_j^{(r)}), \quad \text{for}\quad c=0,\dots,\tau-1,
\end{align}
where $\tilde{\mathbf{g}}_{j}^{(c,r)}\triangleq\nabla{f}_j(\boldsymbol{w}^{(c,r)}_j\mathcal{Z}_j^{(c,r)})$ is the stochastic gradient of node $j$ at round $r$ for the $c$-th local update. In the above update the local descent direction $\tilde{\boldsymbol{d}}^{(c,r)}_{j,q}$ is defined as the difference the local stochastic gradient $\tilde{\mathbf{g}}_{j}^{(c,r)}$ and the correction vector $\delta_j^{(r)}$ which aims to track the difference between local and global gradient directions. Note that for all $\tau$ local updates at round $r$, the vector $\delta_j^{(r)}$ is fixed while the local stochastic gradient $\tilde{\mathbf{g}}_{j}^{(c,r)}$ is computed via fresh samples for each local update. 
Once the local models $\boldsymbol{w}_j^{(\tau,r)}$ are computed, nodes send their quantized accumulation of gradients $\boldsymbol{\Delta}_{j,q}^{(r)}$ to the server. Then, the server uses this information to compute the average update $\boldsymbol{\Delta}_{q}^{(r)}\triangleq\frac{1}{m}\sum_{j=1}^m\boldsymbol{\Delta}_{j,q}^{(r)}$ and broadcasts it to the devices. Moreover, the server utilizes $\boldsymbol{\Delta}_{q}^{(r)}$ to compute the new global model ${\boldsymbol{w}}^{(r+1)}$ according to \eqref{eq:lfsgd-update}. The steps of \texttt{FedCOMGATE} are outlined in Algorithm~\ref{Alg:VRFLDL}.

\begin{algorithm2e}[t]
\DontPrintSemicolon
\SetNoFillComment
\setstretch{0.5}
\LinesNumbered
\caption{\texttt{FedCOMGATE}($R, \tau, \eta, \gamma$)}\label{Alg:VRFLDL}
\SetKwFor{ForPar}{for}{do in parallel}{end forpar}
\textbf{Inputs:} Number of communication rounds $R$, number of local updates $\tau$, learning rates $\gamma$ and $\eta$, initial  global model $\boldsymbol{w}^{(0)}$, initial gradient tracking $\mathbf{\delta}_j^{(0)}=\boldsymbol{0}, \; \forall j\in[m]$\\[2pt] 
\For{$r=0, \ldots, R-1$}{
    \ForPar{each client $j\in[m]$}{
        Set $\boldsymbol{w}_j^{(0,r)}={\boldsymbol{w}}^{(r)}$ \\
        \For{ $c=0,\ldots,\tau-1$}{
            Set $\tilde{\boldsymbol{d}}^{(c,r)}_{j,q}=\tilde{\mathbf{g}}_{j}^{(c,r)}-\delta_j^{(r)}$ where $\tilde{\mathbf{g}}_{j}^{(c,r)}\triangleq\nabla f_j (\boldsymbol{w}^{(c,r)}_j;\mathcal{Z}_j^{(c,r)})$\\
             $\boldsymbol{w}^{(c+1,r)}_{j}=\boldsymbol{w}^{(c,r)}_j-\eta~ \tilde{\boldsymbol{d}}^{(c,r)}_{j,q}$
        }
        Device \textbf{sends} $  \boldsymbol{\Delta}_{j,q}^{(r)} = Q(({\boldsymbol{w}^{(r)}-\boldsymbol{w}^{(\tau,r)}_j})/\eta)$  to the server \\
        Device \textbf{updates}  $\delta_j^{(r+1)}=\delta_j^{(r)}+\frac{1}{\tau}( \boldsymbol{\Delta}_{j,q}^{(r)}-\boldsymbol{\Delta}_{q}^{(r)})$ 
    }
Server  \textbf{computes} $\boldsymbol{\Delta}_q^{(r)}=\frac{1}{m}\sum_{j=1}^m \boldsymbol{\Delta}_{j,q}^{(r)}$ and \textbf{broadcasts} back to all devices\\
    Server \textbf{computes} $\boldsymbol{w}^{(r+1)}=\boldsymbol{w}^{(r)}-\eta\gamma\boldsymbol{\Delta}_q^{(r)}$ and \textbf{broadcasts} to all devices\\
}
\end{algorithm2e}

\paragraph{Comparison with SCAFFOLD~\cite{karimireddy2019scaffold} and VRL-SGD in \cite{liang2019variance}.} From an algorithmic standpoint, in comparison to the SCAFFOLD method proposed in~\cite{karimireddy2019scaffold} , in addition to the fact that we use compressed signals to further reduce the communication overhead, we would like to highlight that our algorithm is much simpler and does not require any extra control variable (see Eq.~(4) and Eq.~(5) in \cite{karimireddy2019scaffold} for more details). Also, since we do not use an extra control variable, the extension of our convergence analysis to the case where a subset of devices participate at each communication round is straightforward and for clarity, we do not include analysis with device sampling. Yet, we shall study the impact of device sampling empirically (see  Figure~\ref{sampl_comp_sim_l10} in Section~\ref{sec:exp} and Algorithm~\ref{Alg:CLGTS} in Appendix~\ref{app:variant}). In comparison to \cite{liang2019variance} which employs an explicit variance reduction component, if we let $Q(\boldsymbol{x})=\boldsymbol{x}$ (case of no quantization), our algorithm reduces to a generalization of algorithm in \cite{liang2019variance} with distinct local and global learning rates. We note that for the case of $\gamma=1$ and $Q(\boldsymbol{x})=\boldsymbol{x}$ the \texttt{FedCOMGATE}($\tau, \eta, \gamma=1$) reduces to the federated algorithm proposed in \cite{liang2019variance} with minor distinction that our algorithm's output is the global model at the server.

\paragraph{Comparison with DIANA \cite{horvath2019stochastic}}\label{sec:compar-diana}
\begin{table*}[t]
    \centering
    \resizebox{0.9\linewidth}{!}{%
    \begin{tabular}{llllll}
        \toprule
                    &  \multicolumn{3}{c}{Objective function} &
        \\ \cmidrule(r){2-5}
        Reference        & Nonconvex      & PL  & Strongly Convex                                  & General Convex   & F.S.
        \\
        \midrule
        \makecell{DIANA~\cite{horvath2019stochastic}}  & \makecell[l]{$-$}   &                 \makecell[l]{$-$} & \makecell[l]{$R=\tilde{O}\left({\kappa+\frac{\kappa q}{m}+q}\right)$ \\ $\tau=1$} &           \makecell[l]{$-$}                                         &  \makecell{\ding{55}} 
        \\

        \midrule
        \makecell{VR-DIANA\cite{horvath2019stochastic}}  & \makecell[l]{$R=O\left(\frac{{\left(1+\frac{q}{m}\right)^{\frac{1}{2}}\left(n^{2/3}+q\right)}}{\epsilon}\right)$ \\ $\tau=1$}   & \makecell[l]{$-$}   &  \makecell[l]{$R=\tilde{O}\left({\kappa+\frac{\kappa q}{m}+q+n}\right)$ \\ $\tau=1$}          &                                       \makecell[l]{$R=O\left(\frac{{\left(1+\frac{q}{m}\right)\sqrt{n}+\frac{q}{\sqrt{n}}}}{\epsilon}\right)$ \\ $\tau=1$}                                       &   \makecell{\ding{52}} 
         \\
        \midrule
       \makecell{FedCOM (ours)} & \makecell[l]{$\boldsymbol{R=O\left(\frac{1}{\epsilon}\right)}$ \\[3pt] $\boldsymbol{\tau=O\left(\frac{{\frac{q}{m}+1}}{m\epsilon}\right)}$}   & \makecell[l]{$\boldsymbol{R=\tilde{O}\left({\kappa+\frac{\kappa q}{m}}\right)}$ \\[3pt] $\boldsymbol{\tau=O\left(\frac{1}{m\epsilon}\right)}$}               &       \makecell[l]{$\boldsymbol{R=\tilde{O}\left({\kappa+\frac{\kappa q}{m}}\right)}$ \\[3pt] $\boldsymbol{\tau=O\left(\frac{1}{m\epsilon}\right)}$}               &\makecell[l]{$\boldsymbol{R\!=\!\tilde{O}\left(\frac{1+\frac{q}{m}}{\epsilon}\right)}$\\[3pt]
       $\boldsymbol{\tau\!=\!O\left(\frac{1}{m\epsilon^2}\right)}$} &   \makecell{\ding{55}} 
   \\
        \bottomrule
    \end{tabular}
    }
\caption{\textbf{Homogeneous} data distribution with $R$ communication rounds and $\tau$ local updates. F.S. stands for finite-sum assumption, $n = \max_{i\in[m]} n_i$, where $n_i$ is the number of local samples at the $i$th device. $m$ is the total number of devices, and $q$ is the quantization noise. We use $\tilde{O}(.)$ to keep key parameters and to omit $\log(\frac{1}{\epsilon})$ term. }
\label{table:5}
\end{table*}

\begin{table*}[t]
    \centering
    \resizebox{0.9\linewidth}{!}{
    \begin{tabular}{llllll}
        \toprule
                    &  \multicolumn{3}{c}{Objective function} &
        \\ \cmidrule(r){2-5}
        Reference        & Nonconvex      & PL & Strongly Convex                                  & General Convex   & F.S.
        \\
        \midrule
                \makecell{DIANA~\cite{horvath2019stochastic}}  & \makecell[l]{$-$}   &                 \makecell[l]{$-$} & \makecell[l]{$R=\tilde{O}\left({\color{black}\kappa+\frac{\kappa q}{m}+q}\right)$ \\ $\tau=1$} &           \makecell[l]{$-$}                                         &  \makecell{\ding{55}}
        \\

        \midrule
        \makecell{VR-DIANA~\cite{horvath2019stochastic}}  & \makecell[l]{$R=O\left(\frac{{\left(1+\frac{q}{m}\right)^{\frac{1}{2}}\left(n^{2/3}+q\right)}}{\epsilon}\right)$ \\ $\tau=1$}   & \makecell[l]{$-$}   &  \makecell[l]{$R=\tilde{O}\left({\color{black}\kappa+\frac{\kappa q}{m}+q+n}\right)$ \\ $\tau=1$}          &                                       \makecell[l]{$R=O\left(\frac{{\left(1+\frac{q}{m}\right)\sqrt{n}+\frac{q}{\sqrt{n}}}}{\epsilon}\right)$ \\ $\tau=1$}                                       &  \makecell{\ding{52}} 
         \\
        \midrule
       \makecell{FedCOMGATE (ours)} & \makecell[l]{$\boldsymbol{R=O\left(\frac{{q+1}}{\epsilon}\right)}$ \\[3pt] $\boldsymbol{\tau=O\left(\frac{1}{m\epsilon}\right)}$}   & \makecell[l]{$\boldsymbol{R=\tilde{O}\left({\kappa\left(q+1\right)}\right)}$ \\[3pt] $\boldsymbol{\tau=O\left(\frac{1}{m\epsilon}\right)}$}               &       \makecell[l]{$\boldsymbol{R=\tilde{O}\left({\color{black}\kappa\left(q+1\right)}\right)}$ \\[3pt] $\boldsymbol{\tau=O\left(\frac{1}{m\epsilon}\right)}$}                                                &\makecell[l]{$\boldsymbol{R\!=\!\tilde{O}\left(\frac{{1+q}}{\epsilon}\right)}$\\[3pt]
       $\boldsymbol{\tau\!=\!O\left(\frac{1}{m\epsilon^2}\right)}$} &  \makecell{\ding{55}} 
   \\
        \bottomrule
    \end{tabular}
    }
\caption{\textbf{Heterogeneous} data distribution with $R$ communication rounds and $\tau$ local updates. F.S. stands for finite-sum assumption, $n = \max_{i\in[m]} n_i$, where $n_i$ is the number of local samples at the $i$th device. $m$ is the total number of devices, and $q$ is the  quantization noise. We use $\tilde{O}(.)$ to keep key parameters and to omit $\log(\frac{1}{\epsilon})$ term. Note that  our results for PL condition hold for the strongly convex case as the latter is implied by former. }
\label{table:6}
\end{table*}
 We provide a summary of the comparison of our algorithms and algorithms introduced in \cite{horvath2019stochastic} in two tables. We compare the rates in  homogeneous and heterogeneous data distributions separately. 
The following comments are in place:

 In the homogeneous setting, shown in Table \ref{table:5} and in comparison to DIANA and VR-DIANA,  FedCOM improves all the communication rounds in terms of dependency on $q$ (shown in blue). In the heterogeneous setting, shown in Table \ref{table:6}, in comparison to DIANA and VR-DIANA, FedCOMGATE basically improves all the communication rounds in terms of dependency on $q$ (shown in blue) except for the strongly convex (SC) case. For the SC case of heterogeneous setting, we highlight that our results are for the PL, unlike DIANA which is for SC. Thus, we believe that if we derive the results directly for SC we might obtain the same or even better results than DIANA (like homogeneous setting). Comparison of finite-sum and stochastic algorithms does not seem to be fair, but per your request, we provide the full comparison. We believe if we analyze our methods for FS settings the dependency on $n$ would only appear in $\tau$ (not $R$). For deterministic settings, i.e., setting $n=1$ in DIANA and $\sigma^2=0$ in FedCOM and FedCOMGATE, again we observe that the communication bounds for FedCOM and FedCOMGATE are better in terms of dependency on $q$ in homogeneous and heterogeneous settings except for the SC heterogeneous case.

\section{Convergence Analysis}\label{Sec:convergence}
Next, we present the convergence analysis of our proposed methods. First, we state our assumptions. 

\begin{assumption}[Smoothness and Lower Boundedness]\label{Assu:1}
The local objective function $f_j(\cdot)$ of $j$th device is differentiable for $j\in [m]$ and $L$-smooth, i.e., $\|\nabla f_j(\boldsymbol{u})-\nabla f_j(\boldsymbol{v})\|\leq L\|\boldsymbol{u}-\boldsymbol{v}\|,\: \forall \;\boldsymbol{u},\boldsymbol{v}\in\mathbb{R}^d$. Moreover, the optimal value of objective function $f(\cdot)$ is bounded below by ${f^*} = \min_{\boldsymbol{w}} f(\boldsymbol{w})>-\infty$. 
\end{assumption}

\begin{assumption}\label{Assu:09}
The output of the compression operator $Q(\boldsymbol{x})$ is an unbisased estimator of its input $\boldsymbol{x}$, and and its variance grows with the squared of the squared of $\ell_2$-norm
of its argument, i.e., $
    \mathbb{E}[Q(\boldsymbol{x})|\boldsymbol{x}]=\boldsymbol{x}$ and $\mathbb{E}[\|Q(\boldsymbol{x})-\boldsymbol{x}\|^2|\boldsymbol{x}]\leq q\left\|\boldsymbol{x}\right\|^2.
$
\end{assumption}

 Assumptions~\ref{Assu:1}-\ref{Assu:09} are customary in the analysis of methods with compression, and they will all be assumed in all of our results. We should also add that several quantization approaches and sparsification techniques satisfy the condition in Assumption~\ref{Assu:09}. For examples of such compression schemes we refer the reader to \cite{basu2019qsparse,horvath2019stochastic}. We report our results for three different class of loss functions: (i) nonconvex (ii) convex (iii) non-convex \pl~(PL). Indeed, as any $\mu$-strongly convex is $\mu$-PL \cite{karimi2016linear}, our results for the PL case automatically hold for strongly convex functions.

\subsection{Convergence of  \texttt{FedCOM} in the homogeneous data distribution setting}

Now we focus on the homogeneous case in which the stochastic local gradient of each worker is an unbiased estimator of the global gradient.

\begin{assumption}[Bounded Variance]\label{Assu:1.5}
For all $j\in [m]$, we can sample an independent mini-batch $\mathcal{Z}_j$   of size $|\mathcal{Z}_j^{(c,r)}| = b$ and compute an unbiased stochastic gradient  $\tilde{\mathbf{g}}_j = \nabla f_j(\boldsymbol{w}; \mathcal{Z}_j), \mathbb{E}_{\mathcal{Z}_j}[\tilde{\mathbf{g}}_j] = \nabla f(\boldsymbol{w})=\mathbf{g}$. Moreover, their variance is bounded above by a constant $\sigma^2$, i.e., $
\mathbb{E}_{\mathcal{Z}_j}\left[\|\tilde{\mathbf{g}}_j-\mathbf{g}\|^2\right]\leq \sigma^2$.
\end{assumption}

In the following theorem, we state our main theoretical results for \texttt{FedCOM} in the homogeneous setting.

\begin{theorem}\label{thm:homog_case}
 Consider \texttt{FedCOM} in Algorithm~\ref{Alg:one-shot-using data samoples-b}. Suppose that the conditions in Assumptions~\ref{Assu:1}-\ref{Assu:1.5} hold. If the local data distributions of all users are identical (homogeneous setting), then we have  
 \begin{itemize}
     \item \textbf{Nonconvex:}  By choosing stepsizes as $\eta=\frac{1}{L\gamma}\sqrt{\frac{m}{R\tau\left(q+1\right)}}$ and $\gamma\geq m$, the sequence of iterates satisfies  $\frac{1}{R}\sum_{r=0}^{R-1}\left\|\nabla f({\boldsymbol{w}}^{(r)})\right\|_2^2\leq {\epsilon}$ if we set
     $R=O\left(\frac{1}{\epsilon}\right)$ and $ \tau=O\left(\frac{q+1}{{m}\epsilon}\right)$.
     \item \textbf{Strongly convex or PL:}
      By choosing stepsizes as $\eta=\frac{1}{2L\left(\frac{q}{m}+1\right)\tau\gamma}$ and $\gamma\geq m$, we obtain the iterates satisfy $\mathbb{E}\Big[f({\boldsymbol{w}}^{(R)})-f({\boldsymbol{w}}^{(*)})\Big]\leq \epsilon$ if  we set
     $R=O\left(\left(\frac{q}{m}+1\right)\kappa\log\left(\frac{1}{\epsilon}\right)\right)$ and $ \tau=O\left(\frac{q+1}{m\left(\frac{q}{m}+1\right)\epsilon}\right)$.
     \item \textbf{Convex:} By choosing stepsizes as $\eta=\frac{1}{2L\left(\frac{q}{m}+1\right)\tau\gamma}$ and $\gamma\geq m$, we obtain that the iterates satisfy $ \mathbb{E}\Big[f({\boldsymbol{w}}^{(R)})-f({\boldsymbol{w}}^{(*)})\Big]\leq \epsilon$ if we set
     $R=O\left(\frac{L\left(1+\frac{q}{m}\right)}{\epsilon}\log\left(\frac{1}{\epsilon}\right)\right)$ and $ \tau=O\left(\frac{(q+1)^2}{m\left(\frac{q}{m}+1\right)^2\epsilon^2}\right)$.
 \end{itemize}
\end{theorem}

Theorem~\ref{thm:homog_case} characterizes the number of required local updates $\tau$ and communication rounds $R$ to achieve an $\epsilon$-first-order stationary point for the nonconvex setting and an $\epsilon$-suboptimal solution for convex and strongly convex settings, when we are in a homogeneous case. A few important observations follow. First, in all three results the dependency of $\tau$ and $R$ on the variance of compression scheme $q$ is scaled down by a factor of $1/m$. Hence, by cooperative learning the users are able to lower the effect of the noise induced by the compression scheme. Second, in all three cases, the number of local updates $\tau$ required for achieving a specific accuracy is proportional to $1/m$. In the homogeneous setting, this result is expected since we have $m$ machines and the number of samples used per local update is $m$ times of the case that only a single machine runs local SGD. As a result, the overall number of required local updates scales inversely by the number of machines $m$. Third, in all three cases, the dependency of communication rounds $R$ on the required accuracy $\epsilon$ matches the number of required updates for solving that problem in centralized deterministic settings. For instance, in a centralized nonconvex setting, to achieve a point that satisfies $\|\nabla f(\boldsymbol{w})\|^2\leq \epsilon$ we need ${O}(1/\epsilon)$ gradient updates for deterministic case and ${O}(1/\epsilon^2)$ SGD updates for the stochastic case. It is interesting that running $\tau ={O}(1/\epsilon)$ local updates controls the noise of stochastic gradients and the number of communication rounds $R ={O}(1/\epsilon)$ stays same as the centralized deterministic case. Similar observations hold for convex (upto a log factor) and strongly convex cases.
\begin{table}[t]
	\centering
	\resizebox{0.7\linewidth}{!}{
		\begin{tabular}{llll}
			\toprule
			&  \multicolumn{3}{c}{Objective function}
			\\\cmidrule(r){2-4}
			Reference        & Nonconvex      & PL/Strongly Convex                                  & General Convex 
			\\
			\midrule
			\makecell{Local-SGD\cite{haddadpour2019local}}  & \makecell{$-$} & \makecell[l]{$R=O\left(\left(\frac{1}{\epsilon}\right)^{\frac{1}{3}}\right)$ \\ $\tau=O\left(\frac{1}{m\epsilon}\right)$}   & \makecell{$-$}                                                                            
			\\
			\midrule
			 \makecell{Local-SGD\cite{khaled2020tighter}}  & \makecell{$-$} & \makecell[l]{$R={O\left(m\kappa\log\left(\frac{1}{\epsilon}\right)\right)}$ \\ $\tau={O\left(\frac{1}{m^2\epsilon}\right)}$}   & \makecell[l]{$R={O\left(\frac{m}{\epsilon}\right)}$ \\[2pt] $\tau={O\left(\frac{1}{m^2\epsilon}\right)}$}                  
			\\
			\midrule
			\makecell{Local-SGD\cite{wang2018cooperative}}  & \makecell[l]{$R=O\left(\frac{m}{\epsilon}\right)$ \\[2pt] $\tau=O\left(\frac{1}{m^2\epsilon}\right)$}   & \makecell{$-$} & \makecell{$-$}            
			\\
			\midrule
			\makecell{\textbf{ Theorem~\ref{thm:homog_case}}} & \makecell[l]{$\boldsymbol{R=O\left(\frac{1}{\epsilon}\right)}$ \\[3pt] $\boldsymbol{\tau=O\left(\frac{1}{m\epsilon }\right)}$}   & \makecell[l]{$\boldsymbol{R=O\left(\kappa\log\left(\frac{1}{\epsilon}\right)\right)}$ \\[3pt] $\boldsymbol{\tau=O\left(\frac{1}{m\epsilon}\right)}$}               & \makecell[l]{$\boldsymbol{R=O\left(\frac{1}{\epsilon}\log\left(\frac{1}{\epsilon}\right)\right)}$\\[3pt]
				$\boldsymbol{\tau=O\left(\frac{1}{m\epsilon^2}\right)}$}            
			\\
			\bottomrule
		\end{tabular}
	}
	\caption{Comparison of \texttt{FedCOM} with results that use periodic averaging but do not utilize compression, i.e., $q=0$, in the homogeneous setting.}
	\label{table:3}
	\vspace{-4mm}
\end{table}

\begin{remark}
While the bound obtained in Theorem~\ref{thm:homog_case} for general non-convex objectives  indicates that  achieving a convergence rate of $\epsilon$ requires $R=O\left(\frac{q+1}{\epsilon}\right)$ communication rounds with $\tau=O\left(\frac{1}{m\epsilon}\right)$ local updates, in Remark~\ref{rmk:cnd-lr} in Appendix~\ref{sec:app:sgd:undrr-pl}, we  show that the same rate can be achieved with $R=O\left(\frac{1}{\epsilon}\right)$ and $\tau=O\left(\frac{q+1}{m\epsilon}\right)$. Hence, the noise of quantiziation can be compensated with higher number of local steps $\tau$.
\end{remark}

\begin{remark}
The results for \texttt{FedCOM} improve the complexity bounds for other federated leaning methods with compression (in the homogeneous setting) that are proposed in  \cite{reisizadeh2019fedpaq} and \cite{basu2019qsparse}. Check Table 1 for more details.  
\end{remark}

\begin{remark}
To show the tightness of our result for \texttt{FedCOM}, we also compare its results with schemes without compression, $q=0$, developed for homogeneous settings, shown in Table~\ref{table:3}. As we observe, the number of required communication rounds for \texttt{FedCOM} without compression in convex, strongly convex, and nonconvex settings are smaller than the best-known rates for each setting by a factor of $\frac{1}{m}$. 
\end{remark}

\subsection{Convergence of  \texttt{FedCOMGATE} in the data heterogeneous setting}

Next, we report our results for \texttt{FedCOMGATE} in the heterogeneous setting. We consider a less strict assumption compared to Assumption~\ref{Assu:1.5}, that  the stochastic gradient of each user is an unbiased estimator of its local gradient with bounded variance.

\begin{assumption}[Bounded Variance]\label{Assu:2}
For all $j\in [m]$, we can sample an independent mini-batch $\mathcal{Z}_j$   of size $|\mathcal{Z}_j| = b$ and compute an unbiased stochastic gradient $\tilde{\mathbf{g}}_j = \nabla f_j(\boldsymbol{w}; \mathcal{Z}_j), \mathbb{E}_{\xi}[\tilde{\mathbf{g}}_j] = \nabla f_{j}(\boldsymbol{w})={\mathbf{g}}_j$. Moreover, the variance of local stochastic gradients is bounded above by a constant $\sigma^2$, i.e., $
\mathbb{E}_{\xi}\left[\|\tilde{\mathbf{g}}_j-{\mathbf{g}}_j\|^2\right]\leq \sigma^2$.
\end{assumption}

\begin{assumption}\label{assum:009}
The compression scheme $Q$ for the heterogeneous data distribution setting satisfies the following condition $
    \mathbb{E}_Q[\|\frac{1}{m}\sum_{j=1}^m Q(\boldsymbol{x}_j)\|^2-\|Q(\frac{1}{m}\sum_{j=1}^m \boldsymbol{x}_j)\|^2]\leq G_q$.

\end{assumption}
 \begin{wrapfigure}{r}{0.41\linewidth}
\vspace{-0.2cm}
 	\centering
		\includegraphics[width=0.4\textwidth]{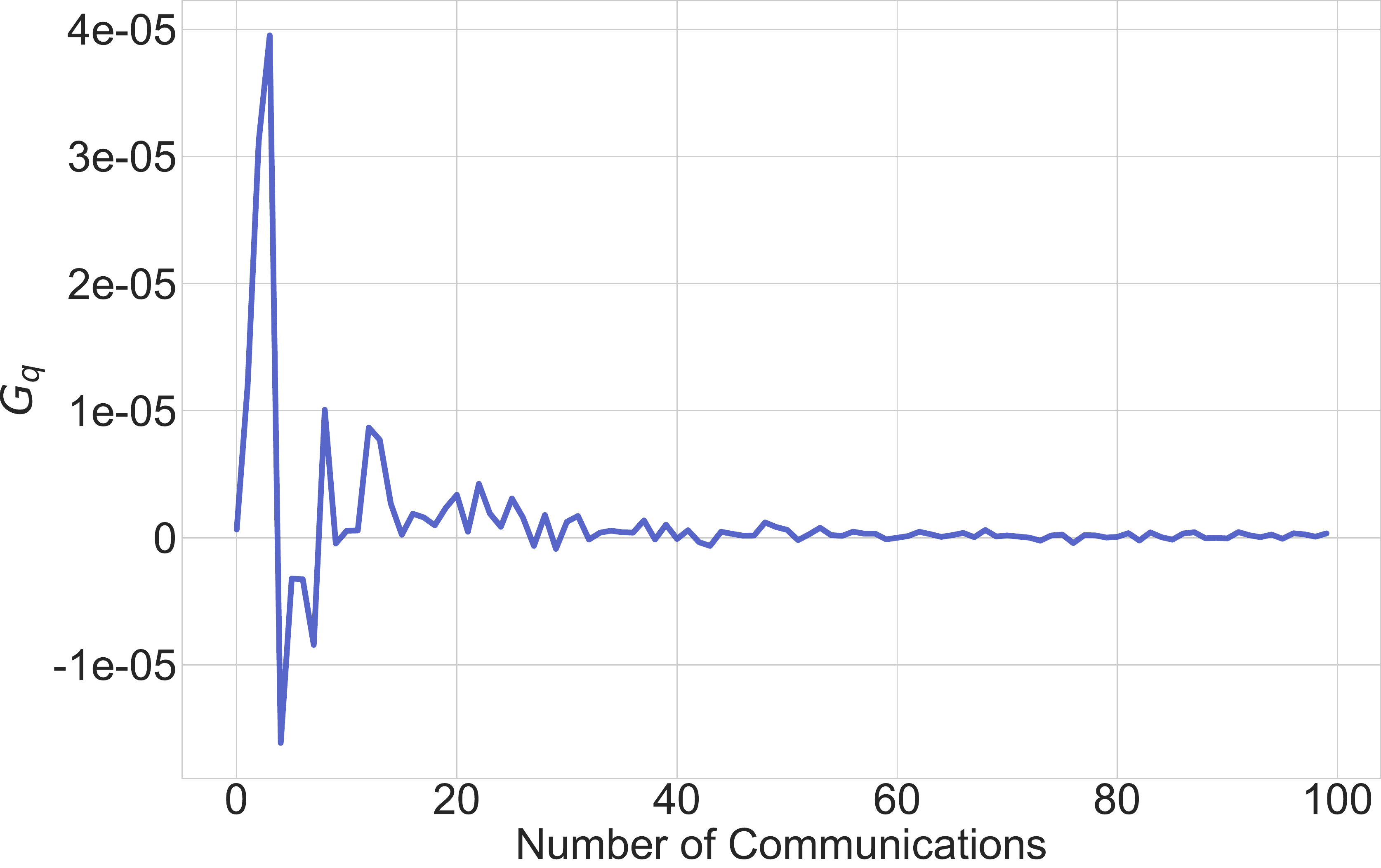}
		\vspace{-1mm}
 	\caption[]{The error of quantization measured by Assumption~\ref{assum:009} for \texttt{FedCOMGATE} with the MNIST dataset applied to a MLP model. We quantized updates $\boldsymbol{\Delta}_j$s from $32$ bits floating-point to $8$ bits integer.}
 	\label{fig:assump6}
 	\vspace{-0.1cm}
 \end{wrapfigure}
The condition in Assumption~\ref{Assu:2} is not strict and only ensures that the local stochastic gradients are unbiased estimators of local gradients with bounded variance.
Regarding Assumption~\ref{assum:009},
for the case of no compression, $Q(\boldsymbol{x})=\boldsymbol{x}$, the compression error becomes naturally $G_q=0$. We highlight that this assumption is only needed in the heterogeneous setting, and since both of the terms in the argument of expectation depend on the quantization, this assumption can be seen as a weaker version of the gradient diversity assumptions in the convergence analysis of heterogeneous settings.
To show how this assumption holds in practice, we run an experiment on the MNIST dataset using \texttt{FedCOMGATE} algorithm with quantizing gradients from $32$ bits floating-point to $8$ bits integer. In Figure~\ref{fig:assump6} we plot  changes in $G_q$ quantity through this experiment. It shows that $G_q$ is decreasing as we proceed with the training, simply because the $\ell_2$-norm of the updated vector is going to zero. Note that the quantity of $G_q$ could be even negative as illustrated in Figure~\ref{fig:assump6}. For more details on the experiment see Section~\ref{sec:exp}.

\begin{table}[t]
    \centering
    \resizebox{0.7\linewidth}{!}{
    \begin{tabular}{llll}
        \toprule
                    &  \multicolumn{3}{c}{Objective function}
        \\ \cmidrule(r){2-4}
        Reference        & Nonconvex      & PL/Strongly Convex                                  & General Convex  
        \\
        \midrule
        \makecell{SCAFFOLD\cite{karimireddy2019scaffold}}  & \makecell[l]{$R=O\left(\frac{1}{\epsilon}\right)$ \\[2pt] $\tau=O\left(\frac{1}{m\epsilon }\right)$}   & \makecell[l]{$R={O\left(\kappa\log\left(\frac{1}{\epsilon}\right)\right)}$ \\[2pt] $\tau={O\left(\frac{1}{m\epsilon}\right)}$}               & \makecell[l]{$R={O\left(\frac{1}{\epsilon}\right)}$ \\[2pt] $\tau={O\left(\frac{1}{m\epsilon}\right)}$}
        \\
        \midrule
         \makecell{Local-SGD\cite{khaled2020tighter}} & \makecell{$-$}  & \makecell{$-$}  & \makecell[l]{$R={O\left(\frac{1}{\epsilon^{1.5}}\right)}$ \\[2pt] $\tau={O\left(\frac{1}{m\epsilon^{0.5}}\right)}$} 
        \\
        \midrule
         \makecell{VRL-SGD\cite{liang2019variance}} & \makecell[l]{$R={O\left(\frac{m}{\epsilon}\right)}$ \\[2pt] $\tau={O\left(\frac{1}{m^2\epsilon}\right)}$} & \makecell{$-$}  & \makecell{$-$}
        \\
        \midrule
       \makecell{\textbf{Theorem~\ref{thm:non-homogen}}} & \makecell[l]{$\boldsymbol{R=O\left(\frac{1}{\epsilon}\right)}$ \\[3pt] $\boldsymbol{\tau=O\left(\frac{1}{m\epsilon }\right)}$}   & \makecell[l]{$\boldsymbol{R=O\left(\kappa\log\left(\frac{1}{\epsilon}\right)\right)}$ \\[3pt] $\boldsymbol{\tau=O\left(\frac{1}{m\epsilon}\right)}$}               & \makecell[l]{$\boldsymbol{R=O\left(\frac{1}{\epsilon}\log\left(\frac{1}{\epsilon}\right)\right)}$\\[3pt]
       $\boldsymbol{\tau=O\left(\frac{1}{m\epsilon^2}\right)}$}    
   \\
        \bottomrule
    \end{tabular}
    }
\caption{Comparison of \texttt{FedCOMGATE} with results that use periodic averaging but do not utilize compression, i.e., $q=0$, in the heterogeneous  setting. While SCAFFOLD~\cite{karimireddy2019scaffold} requires to communicate $2$ vectors to the server in the uplink, other algorithms only communicate $1$ vector.}
\label{table:4}
\vspace{-4mm}
\end{table}

Next, we present our main theoretical results for the \texttt{FedCOMGATE} method in the heterogeneous setting.

\begin{theorem}\label{thm:non-homogen}
 Consider \texttt{FedCOMGATE} in Algorithm~\ref{Alg:VRFLDL}. If Assumptions~\ref{Assu:1}, \ref{Assu:09}, \ref{Assu:2}  and \ref{assum:009} hold, then even for the case the local data distribution of users are different  (heterogeneous setting) we have
 \begin{itemize}
     \item \textbf{Non-convex:} By choosing stepsizes as $\eta=\frac{1}{L\gamma}\sqrt{\frac{m}{R\tau\left(q+1\right)}}$ and $\gamma\geq m$, we obtain that the iterates satsify  $\frac{1}{R}\sum_{r=0}^{R-1}\left\|\nabla f({\boldsymbol{w}}^{(r)})\right\|_2^2\leq \epsilon$ if we set
     $R=O\left(\frac{q+1}{\epsilon}\right)$ and $ \tau=O\left(\frac{1}{m\epsilon}\right)$.
     \item \textbf{Strongly convex or PL:}
      By choosing stepsizes as $\eta=\frac{1}{2L\left(\frac{q}{m}+1\right)\tau\gamma}$ and ${\gamma\geq \sqrt{m\tau}}$, we obtain that the iterates satisfy $\mathbb{E}\Big[f({\boldsymbol{w}}^{(R)})-f({\boldsymbol{w}}^{(*)})\Big]\leq \epsilon$ if we set
      $R=O\left(\left(q+1\right)\kappa\log\left(\frac{1}{\epsilon}\right)\right)$ and $ \tau=O\left(\frac{1}{m\epsilon}\right)$.
     \item \textbf{Convex:}  By choosing stepsizes as $\eta=\frac{1}{2L\left(q+1\right)\tau\gamma}$ and ${\gamma\geq \sqrt{m\tau}}$, we obtain that the iterates satisfy $\mathbb{E}\Big[f({\boldsymbol{w}}^{(R)})-f({\boldsymbol{w}}^{(*)})\Big]\leq \epsilon$ if we set
     $R=O\left(\frac{L\left(1+q\right)}{\epsilon}\log\left(\frac{1}{\epsilon}\right)\right)$ and $ \tau=O\left(\frac{1}{m\epsilon^2}\right)$.
 \end{itemize}
 
\end{theorem}
The implications of Theorem~\ref{thm:non-homogen} are similar to the ones for Theorem~\ref{thm:homog_case}. Yet, unlike the homogeneous setting, the compression variance $q$ does not scale down by a factor of $1/m$. We emphasize that similar to the homogeneous case, in all three cases, the dependency of $R$ on $\epsilon$ matches the number of required update for solving the problem in centralized fashion.

\begin{remark}
\textcolor{blue} To show the tightness of our result for \texttt{FedCOMGATE}, we also compare its complexity bounds with other schemes without compression, $q=0$, for heterogeneous settings, summarized in Table~\ref{table:4}. As it can be observed in all settings, the bound for \texttt{FedCOMGATE} without compression (\texttt{FedGATE}) matches the best-known complexity bounds for these settings (upto a log factor).
\end{remark}
\begin{remark}
We highlight that since we do not need to communicate control variate in uplink, the communication cost of our algorithm is half of the corresponding cost in SCAFFOLD. Yet, for downlink communication, similar to SCAFFOLD our communication cost is doubled compared to FedAvg due to gradient tracking. However, we emphasize that in general communication cost of broadcasting a message is much cheaper than uplink communication.

\end{remark}

\section{Experiments}\label{sec:exp}
In this section, we empirically validate the performance of proposed algorithms. We compare our methods with  FedAvg~\cite{mcmahan2016communication}, its quantized version, FedPAQ~\cite{reisizadeh2019fedpaq}, and SCAFFOLD~\cite{karimireddy2019scaffold} for heterogeneous federated learning. In addition, we present a variant of our algorithm without compression dubbed as \texttt{FedGATE}. The details of \texttt{FedGATE} is described in Algorithm~\ref{Alg:LGT} in Appendix~\ref{app:variant}. Also, a variant of our algorithm with client sampling is presented in Algorithm~\ref{Alg:CLGTS} in Appendix~\ref{app:variant}. In addition to what is presented here, in Appendix~\ref{app:add_exp}, we explore the effects of client sampling, local computation, and sparsification on the convergence of our proposed algorithms.

\paragraph{Setup.} We implement our algorithms on the \texttt{Distributed} library of \texttt{PyTorch}~\cite{paszke2019pytorch}, using Message Passing Interface (\texttt{MPI}), in order to simulate the real-world collaborative learning scenarios such as the one in federated learning. We run the experiments on a HPC cluster with $3$ \texttt{Intel Xeon E5-2695} \texttt{CPU}s, each of which with $28$ processes. 
For this experiment we use four main datasets: \texttt{MNIST}~\cite{mnist},  \texttt{CIFAR10}~\cite{krizhevsky2009learning}, \texttt{Fashion \texttt{MNIST}}~\cite{xiao2017/online} and \texttt{EMNIST}~\cite{caldas2018leaf}. 
For each experiment, we have $100$ devices communicating with the server. Each experiment runs for $100$ rounds of communication between clients and the server, and we report the global model loss on the training data averaged over all clients and the test accuracy over the global model. For \texttt{MNIST} and \texttt{Fashion \texttt{MNIST}} we use an MLP model with two hidden layers, each with $200$ neurons with ReLU activations. For the \texttt{CIFAR10} dataset, we use the same MLP model, each layer with $500$ neurons. For the learning rate, we use a decreasing scheme similar to what is suggested in~\cite{bottou2012stochastic}, where after each iteration the learning rate decreases $1\%$. Then, each experiment's initial learning rate is tuned to achieve the best performance.

\begin{figure}[!t]
\centering
    \begin{minipage}{.64\textwidth}
        \centering
	\subfigure[Homogeneous]{
		\centering
		\includegraphics[width=0.47\textwidth]{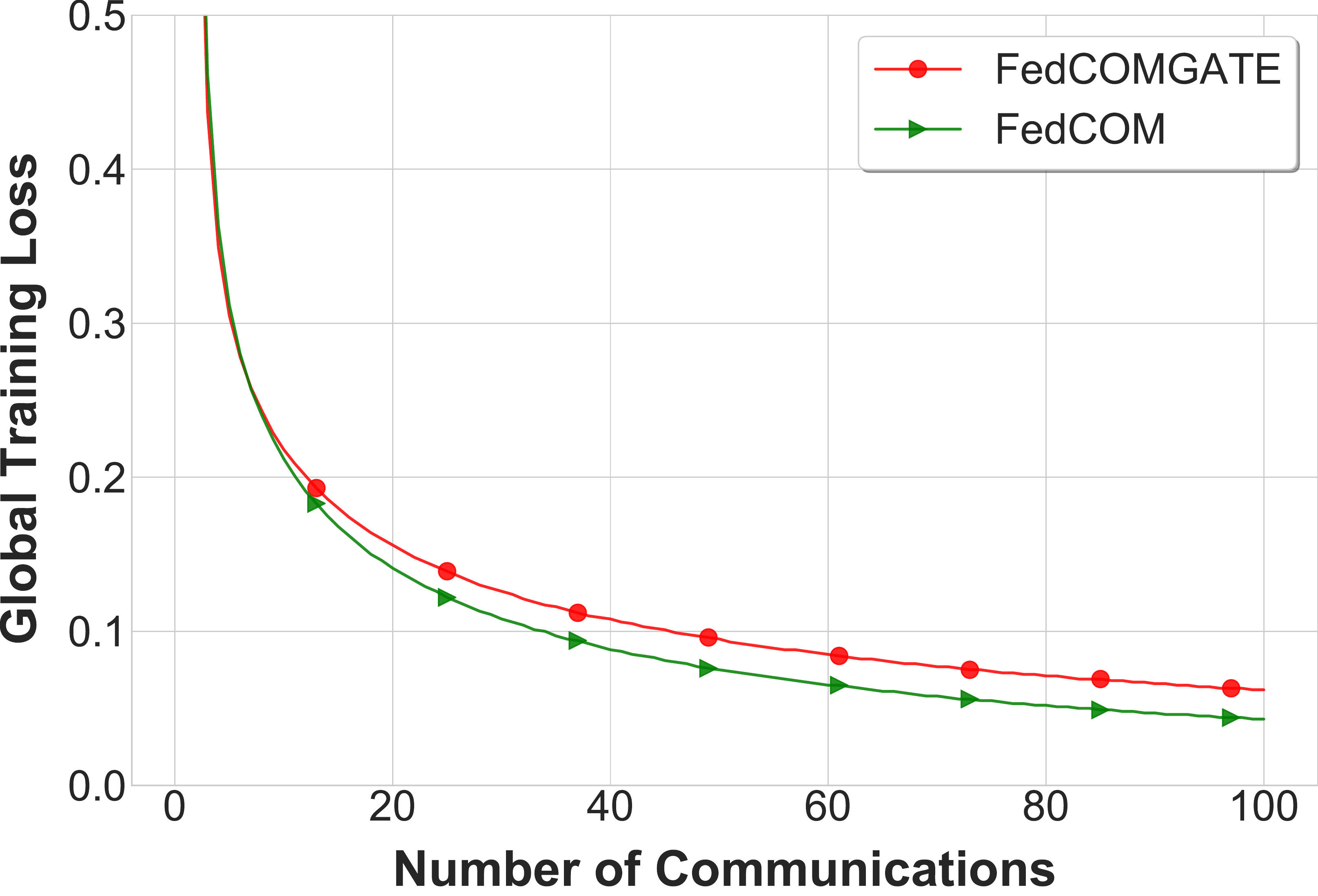}
		\label{fig:comp_mnist_iid}
		}
		\subfigure[Heterogeneous]{
			\centering 
			\includegraphics[width=0.47\textwidth]{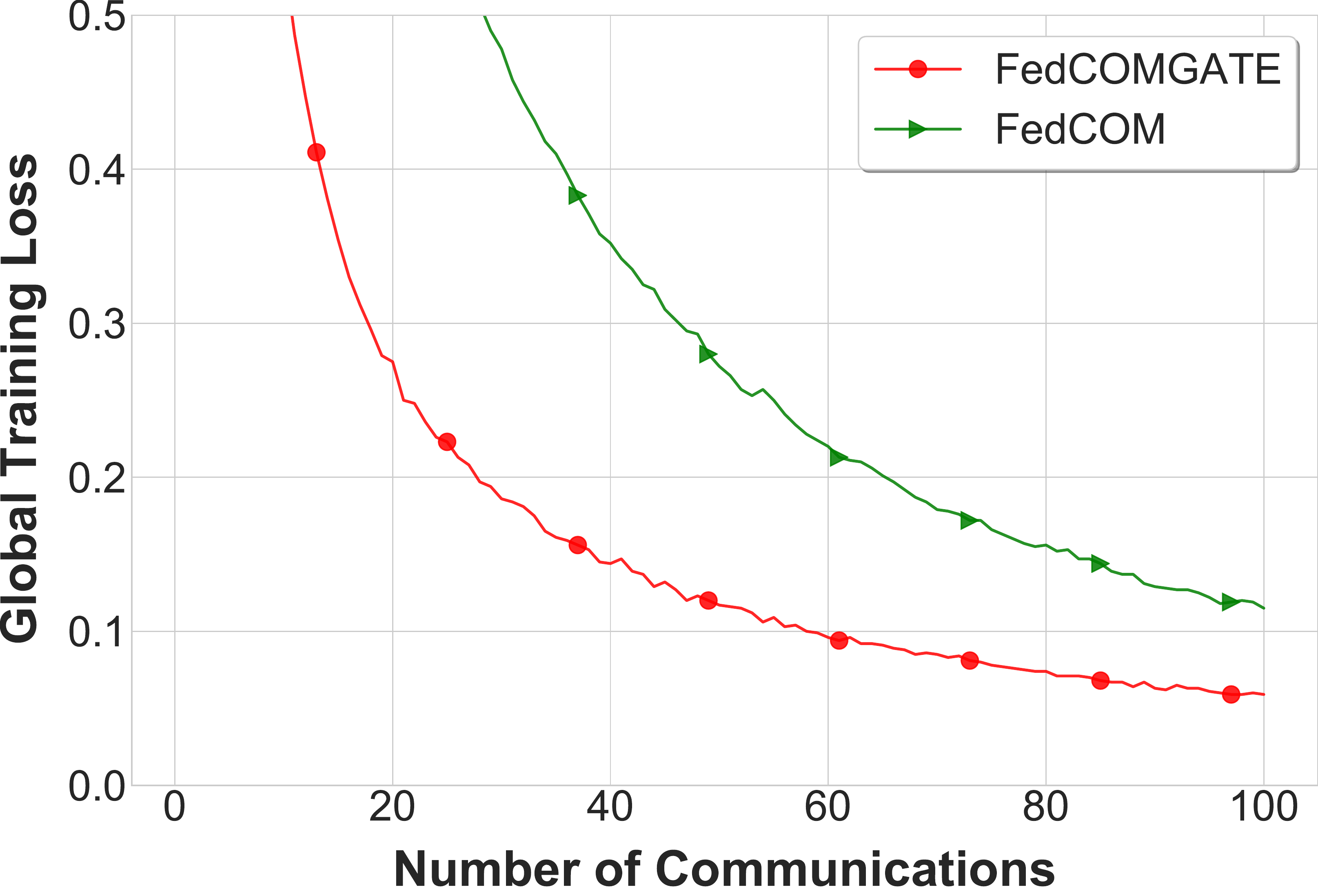}
            \label{fig:comp_mnist_noniid}
			}
			\vspace{-3mm}
	\caption[]{Comparing Algorithm~\ref{Alg:one-shot-using data samoples-b} and Algorithm~\ref{Alg:VRFLDL} for homogeneous and heterogeneous data distributions of the \texttt{MNIST} dataset. In heterogeneous distribution, \texttt{FedCOM} suffers from a residual error.}
	\label{fig:compalgs_mnist}
    \end{minipage}
    \hspace{0.2cm}
    \begin{minipage}{0.31\textwidth}
        \centering
        \includegraphics[width=\textwidth]{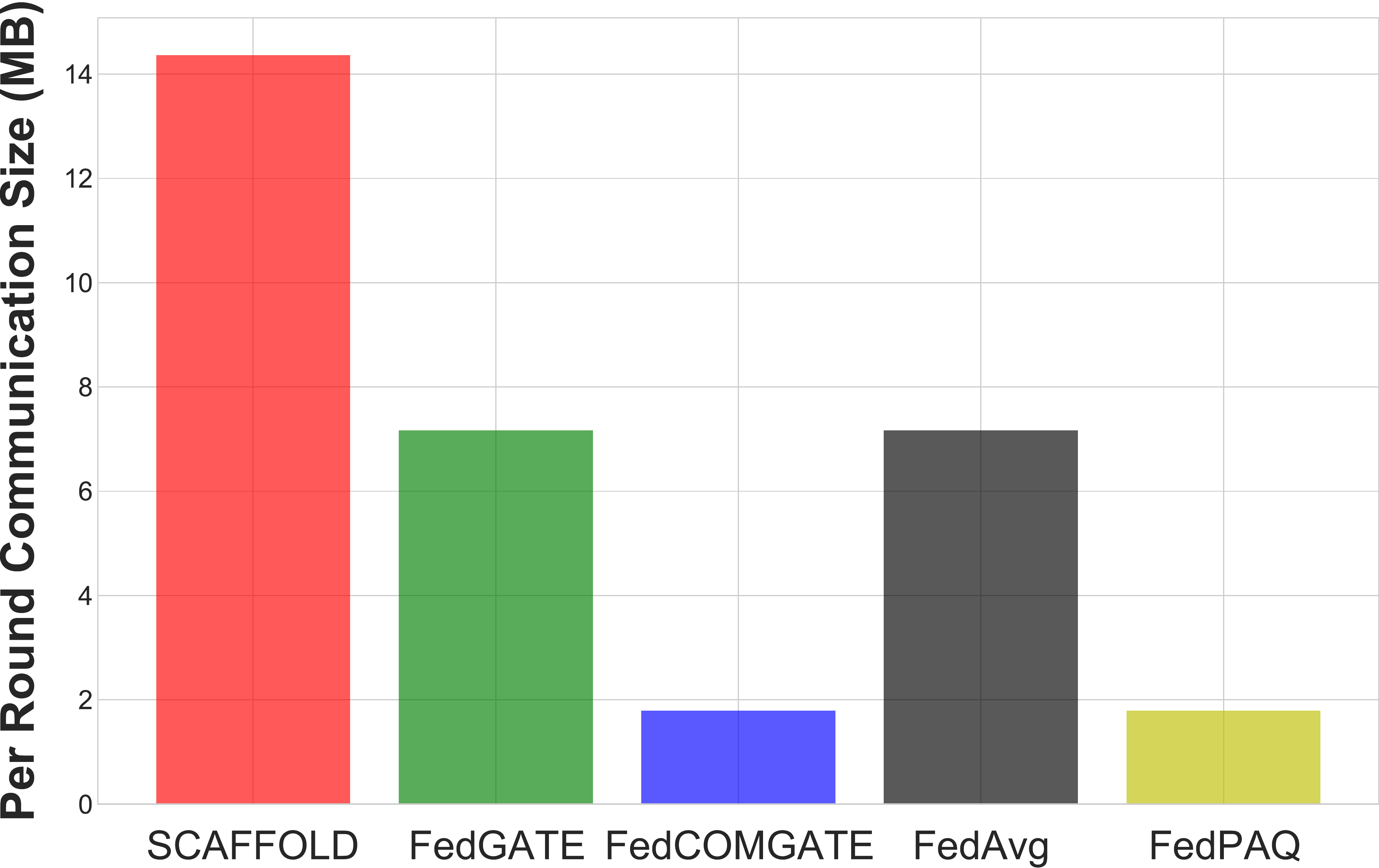}
        	\caption[]{Communication cost at each round for the \texttt{CIFAR10} dataset with a $2$-layer MLP.}
        	\label{fig:comp_mnist_3cpu_l1_time}
     \end{minipage}
\end{figure}

\paragraph{Homogeneous data distribution.}
The \texttt{FedCOM} algorithm is best suited for the homogeneous case, while in the heterogeneous setting it suffers from a residual error. That is why we use gradient tracking in the \texttt{FedCOMGATE} algorithm. This error can be seen in Figure~\ref{fig:compalgs_mnist} for \texttt{MNIST} data, where in Figure~\ref{fig:comp_mnist_iid} data is distributed homogeneously among devices, and in Figure~\ref{fig:comp_mnist_noniid} each device has access to only 2 classes in the dataset. The results indicate that we need gradient tracking in \texttt{FedCOMGATE} to deal with heterogeneity.

\begin{figure*}[t]
		\centering
		\subfigure{
		\centering
		\includegraphics[width=0.31\textwidth]{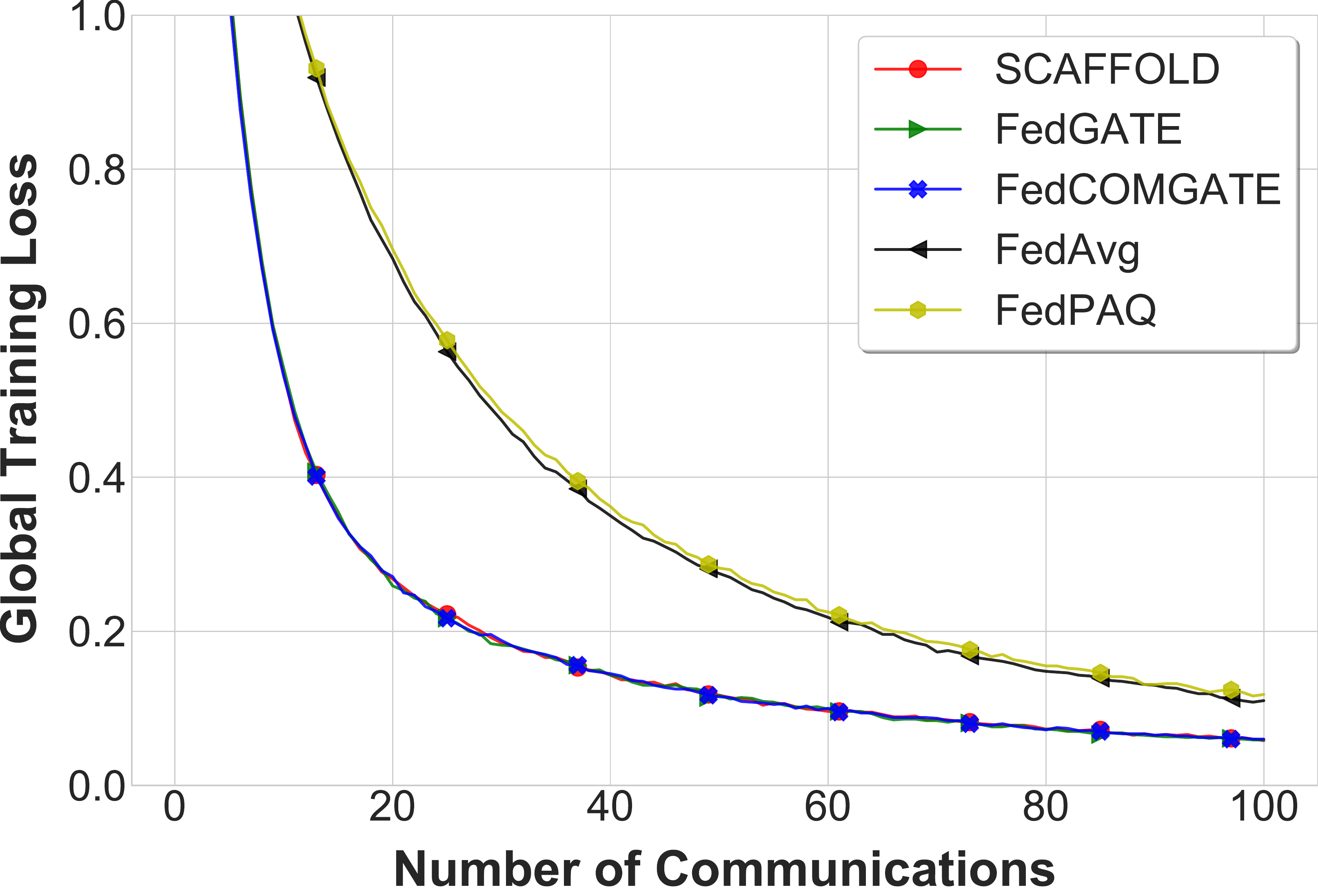}
		\label{fig:comp_mnist_sim_l10_comm_round}
		}
		\hfill
		\subfigure{
			\centering 
			\includegraphics[width=0.31\textwidth]{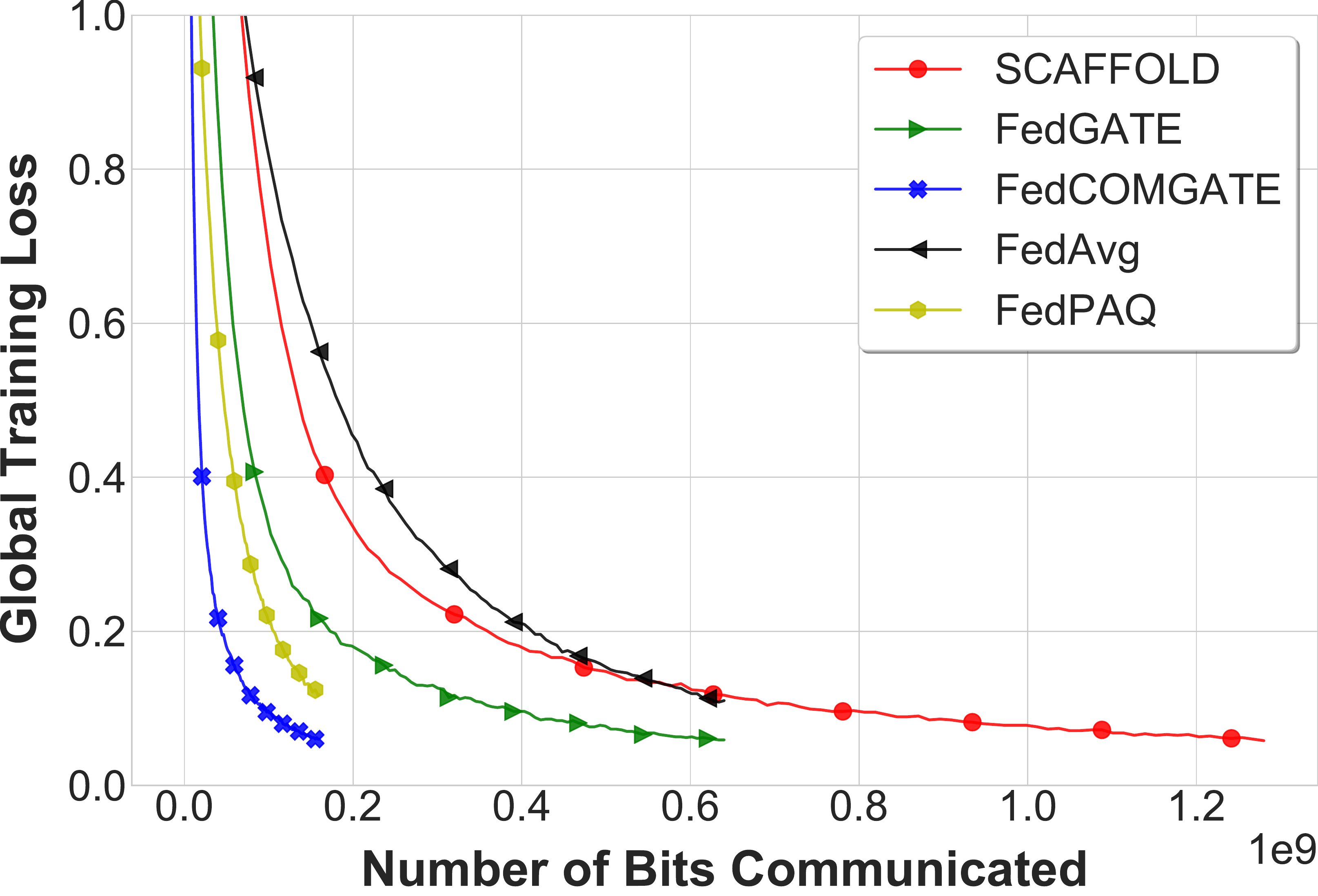}
			\label{fig:comp_mnist_sim_l10_comm_size}
			}
			\hfill
		\subfigure{
			\centering 
			\includegraphics[width=0.31\textwidth]{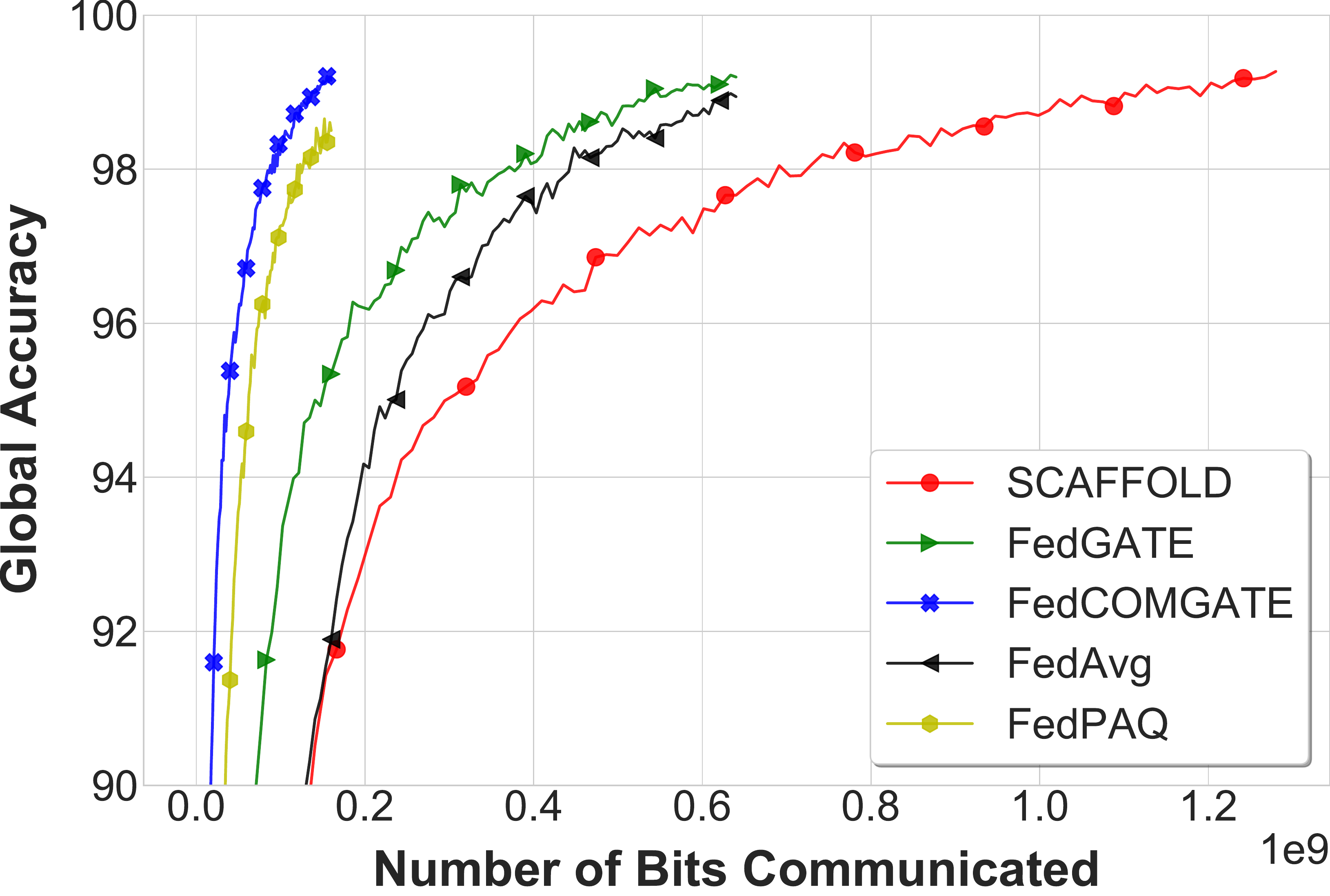}
			\label{comp_mnist_sim_l10_comm_size_acc}
			}
		
		\setcounter{subfigure}{0}
		\subfigure[]{
		\centering
		\includegraphics[width=0.31\textwidth]{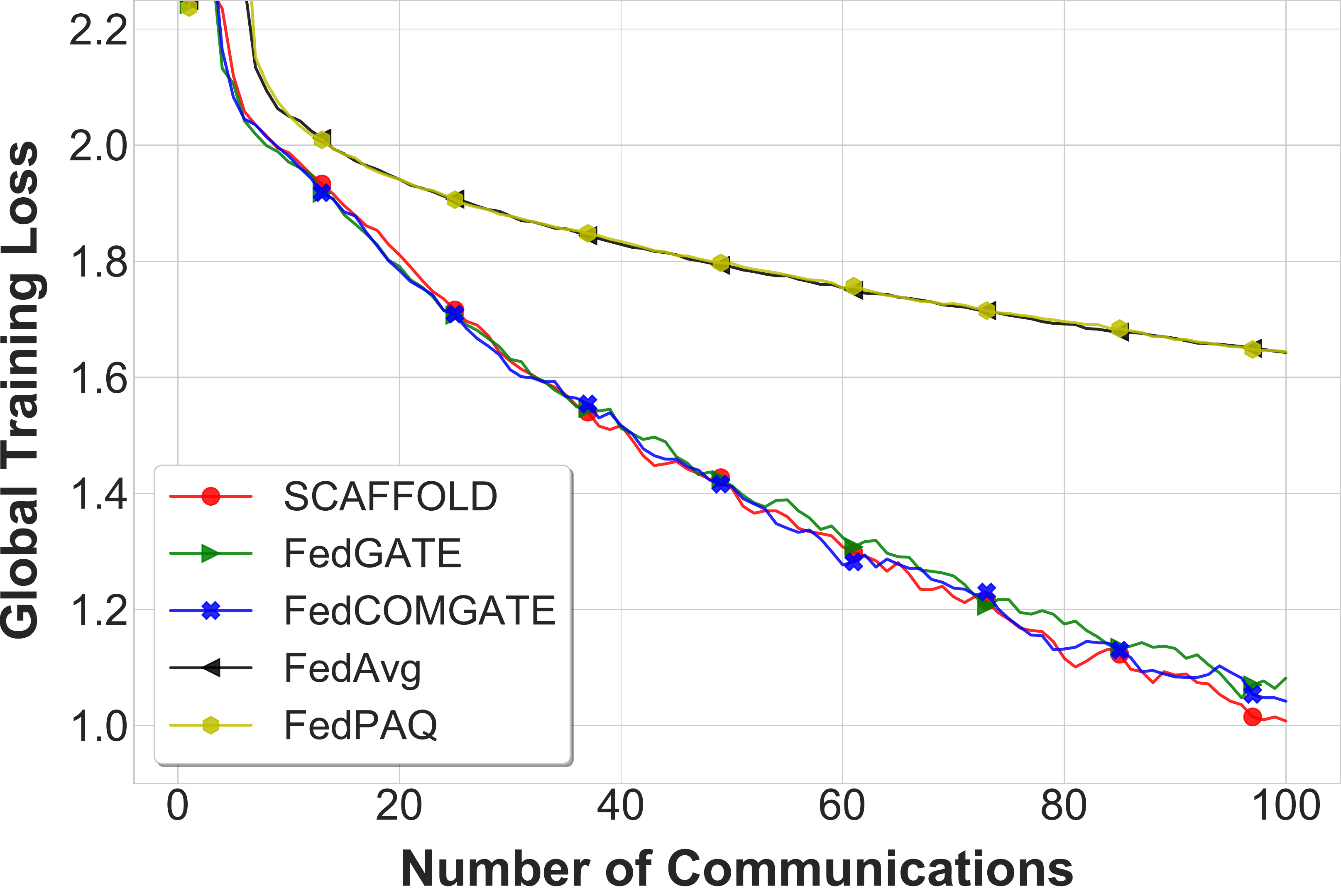}
		\label{fig:comp_cifar10_sim_l10_comm_round}
		}
		\hfill
		\subfigure[]{
			\centering 
			\includegraphics[width=0.31\textwidth]{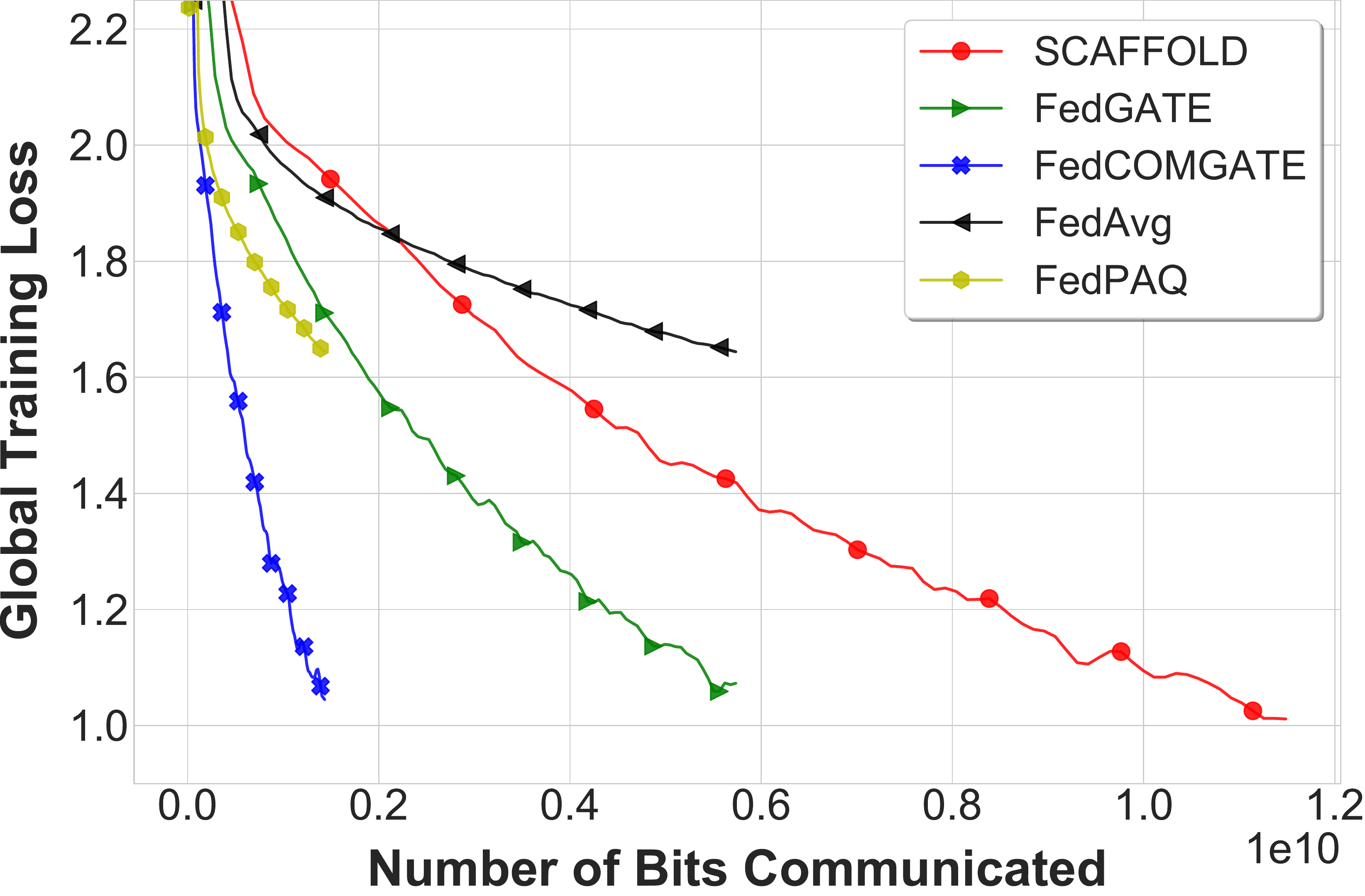}
			\label{fig:comp_cifar10_sim_l10_comm_size}
			}
			\hfill
		\subfigure[]{
			\centering 
			\includegraphics[width=0.31\textwidth]{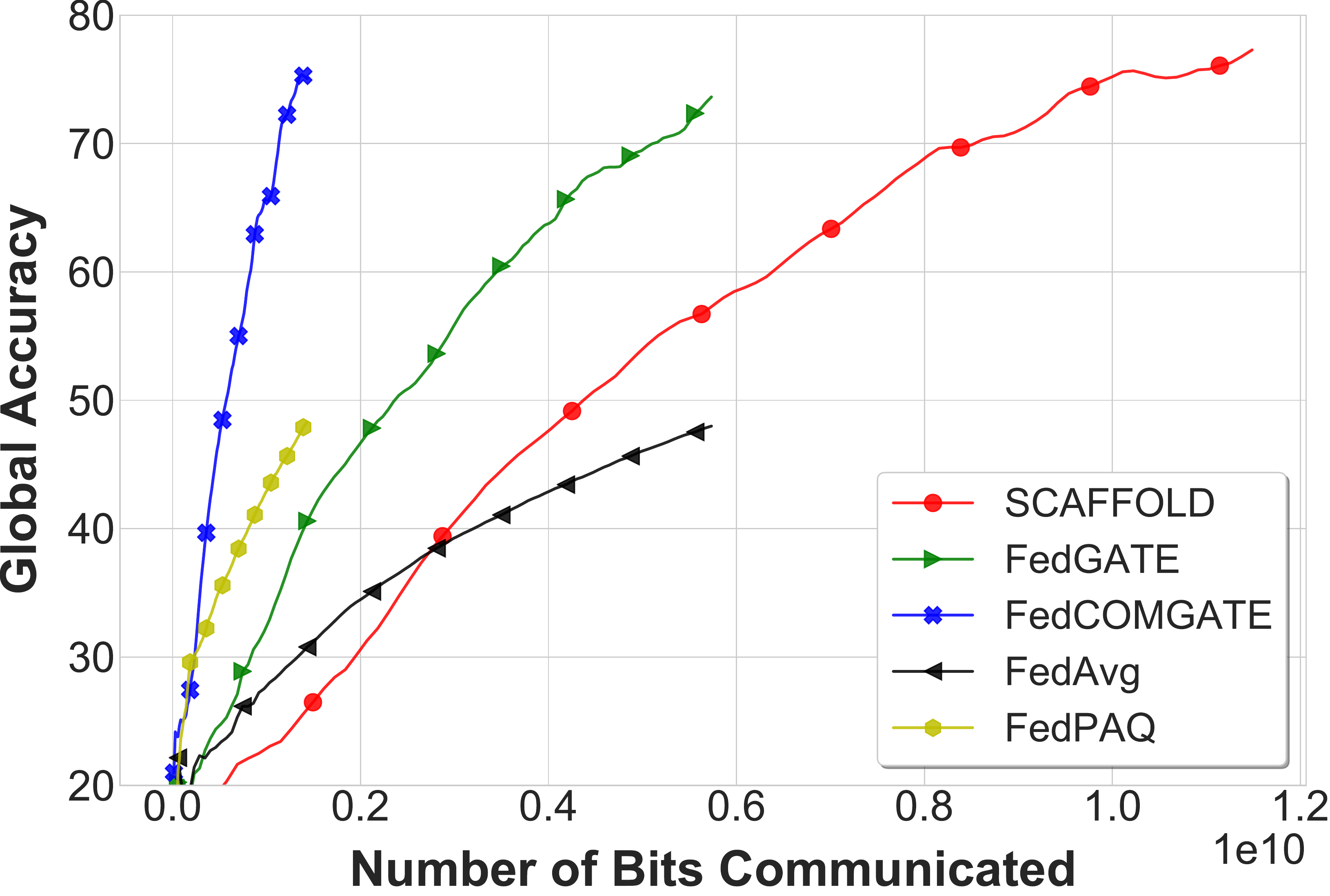}
			\label{fig:comp_cifar10_sim_l10_comm_size_acc}
			}
	\caption[]{Comparing \texttt{FedCOMGATE} and \texttt{FedGATE} with FedAvg~\cite{mcmahan2016communication}, FedPAQ~\cite{reisizadeh2019fedpaq}, and SCAFFOLD~\cite{karimireddy2019scaffold} on the \texttt{MNIST} (first row) and the \texttt{CIFAR10} (second row) datasets. Both \texttt{FedCOMGATE} and \texttt{FedGATE} outperform other algorithms in terms of  communication size between clients and the server and convergence rate.}
	\label{comp_sim_l10}
	\vspace{-4mm}
\end{figure*}

\paragraph{Heterogeneous data distribution.}
To generate heterogeneous data that resembles a real federated learning setup, we will follow a similar approach as in~\cite{mcmahan2016communication}. In this regard, we will distribute the data among clients in a way that each client only has data from two classes, which is highly heterogeneous.
The idea behind \texttt{FedCOMGATE} is similar to the one in SCAFFOLD, except in \texttt{FedCOMGATE} we only have one control variable that gets updated using normal updates in FedAvg. In contrast, SCAFFOLD has two control variables and requires to update the global model and server control variable at each round. Hence, each client in SCAFFOLD communicates at least twice the size as \texttt{FedCOMGATE} with the server at each round, when we do not use any compression. With compression, say $4\times$ quantization, we can substantially reduce the communication cost, say $8\times$, with respect to SCAFFOLD, while preserving the same convergence rate. To compare their communication cost, in Figure~\ref{fig:comp_mnist_3cpu_l1_time} we show the size of variables each client in each algorithm communicates with the server (for the uplink only, since the broadcasting or downlink time is negligible compared to the gathering) for the \texttt{CIFAR10} dataset with an MLP model that has $2$ hidden layers, each with $500$ neurons. For \texttt{FedCOMGATE} and FedPAQ we quantize the updates from $32$ bits floating-point to $8$ bits integer.

To show the effect of this communication size on the real-time convergence of each algorithm, we run each of them on the \texttt{MNIST} and the \texttt{CIFAR10} datasets with MLP models as described before. The data is distributed heterogeneously among clients, where each one has access to only $2$ classes. Figure~\ref{fig:comp_cifar10_sim_l10_comm_round} shows the global model loss on training data on each communication round. FedAvg and FedPAQ are very close to each other on eacc, whereas \texttt{FedCOMGATE}, its normal version without compression \texttt{FedGATE}, and SCAFFOLD are performing similarly based on communication rounds. Figure~\ref{fig:comp_cifar10_sim_l10_comm_size} shows this loss based on the average number of bits communicated between each client and the server during the uplink. Also, Figure~\ref{fig:comp_cifar10_sim_l10_comm_size_acc} shows the test accuracy based on this number of communicated bits. Both figures clearly demonstrate the effectiveness of proposed algorithms. Especially, the \texttt{FedCOMGATE} algorithm superbly outperforms other algorithms where the model size is relatively large.

\begin{figure*}[t!]
		\centering
		\subfigure{
		\centering
		\includegraphics[width=0.425\textwidth]{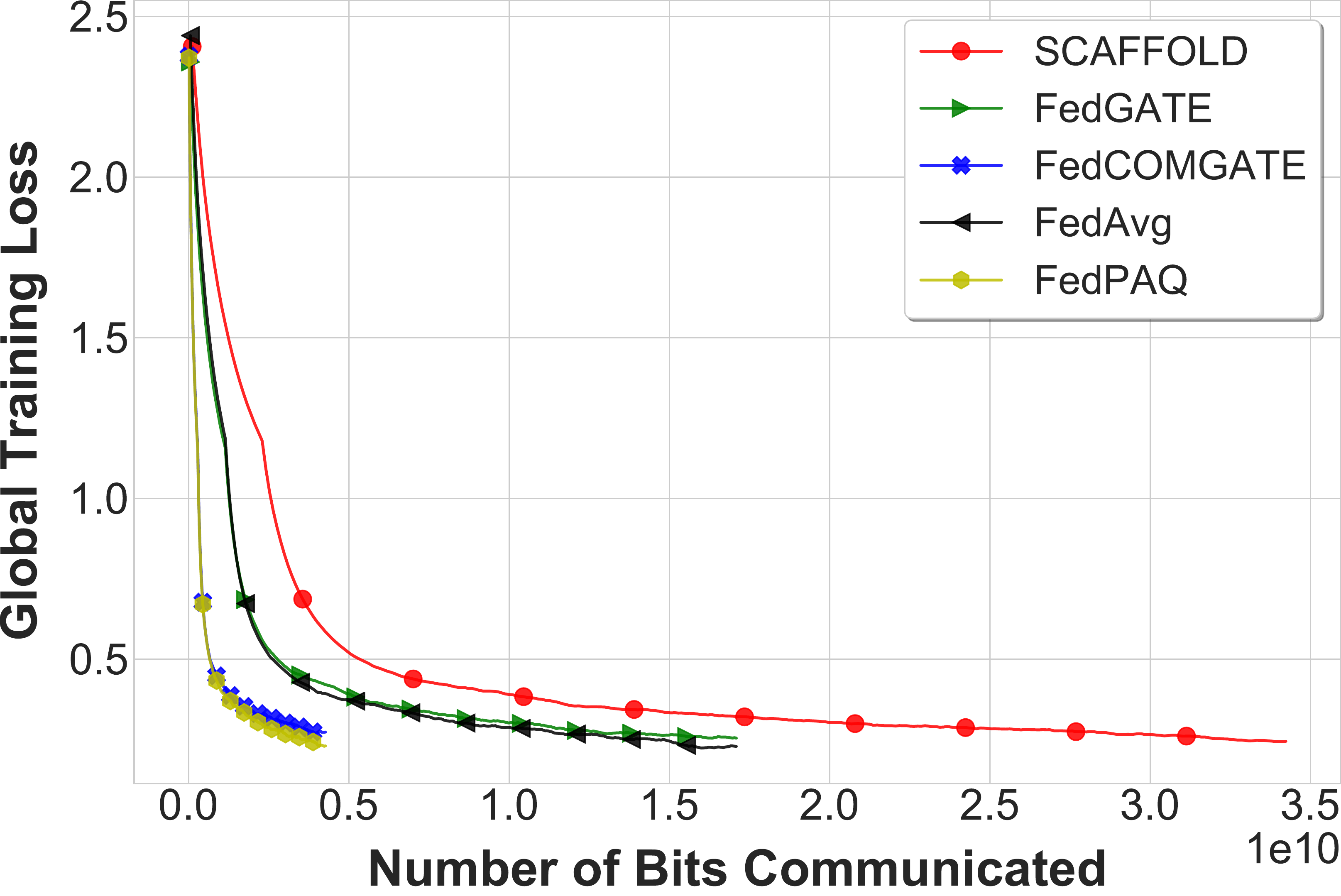}
		\label{fig:emnist1000-loss-commsize}
		}
		\hfill
		\subfigure{
			\centering 
			\includegraphics[width=0.425\textwidth]{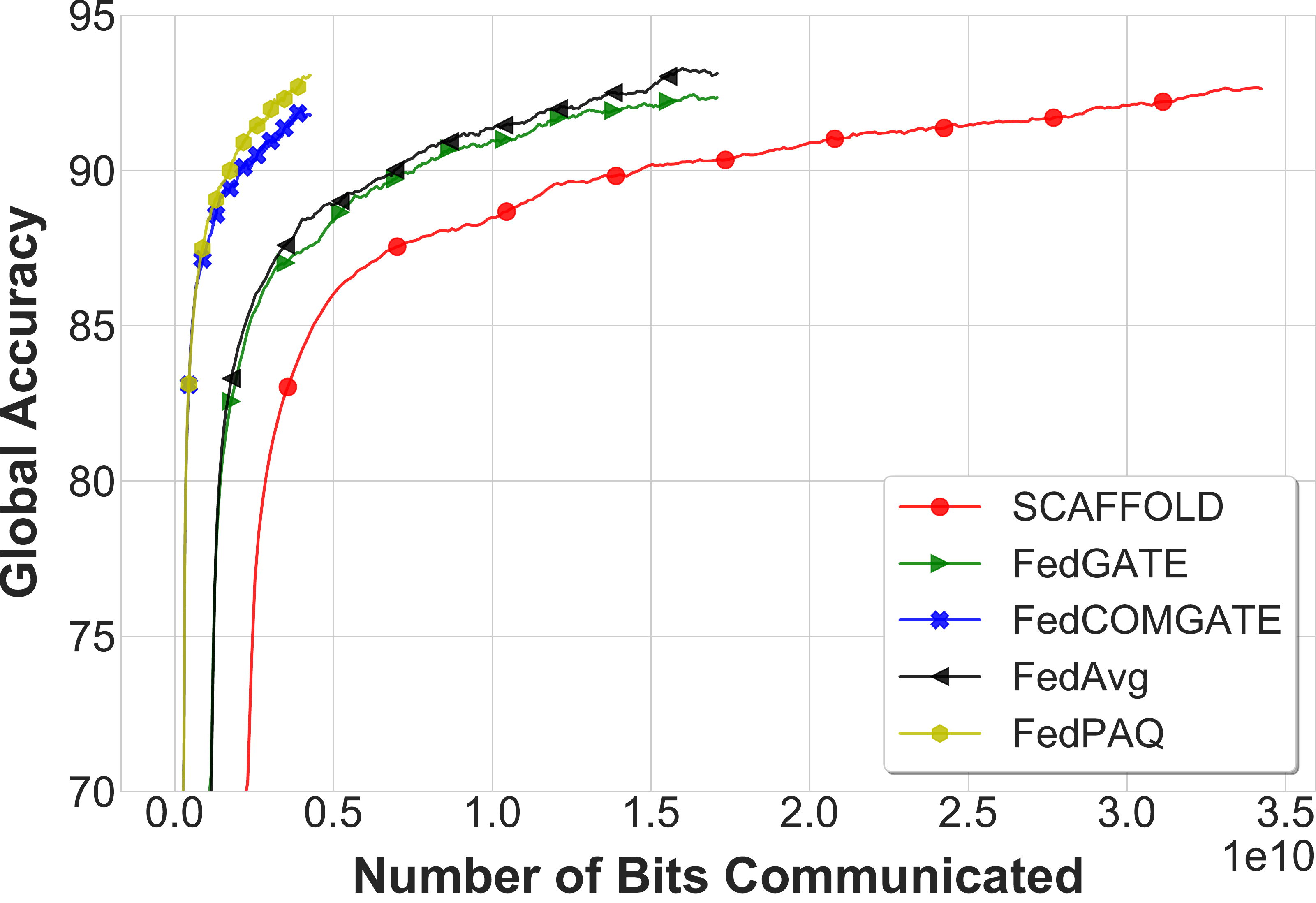}
			\label{fig:emnist1000-acc-commsize}
			}
			\vspace{-2mm}
	\caption[]{Comparing the performance of different algorithms on the \texttt{EMNIST} dataset, using $1000$ clients' data on a 2-layer MLP model. \texttt{FedCOMGATE} and FedPAQ have the fastest convergence in time.}
	\label{emnist1000}
\end{figure*}

\begin{figure*}[t!]
		\centering
		\subfigure[\texttt{EMNIST} dataset]{
		\centering
		\includegraphics[width=0.35\textwidth]{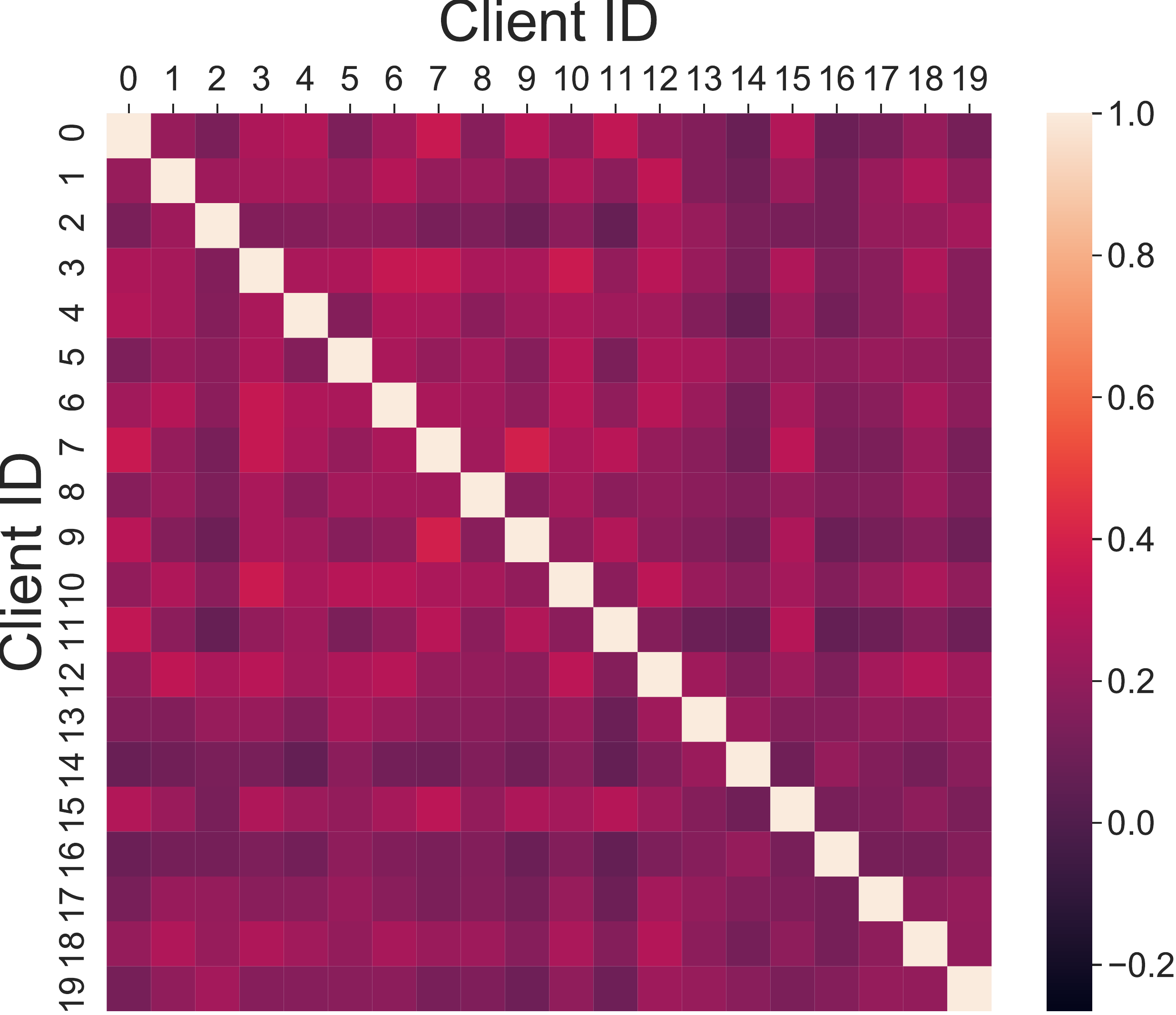}
		\label{fig:emnist_heatmap}
		}
		\hspace{0.2cm}
		\subfigure[\texttt{MNIST} dataset]{
			\centering 
			\includegraphics[width=0.35\textwidth]{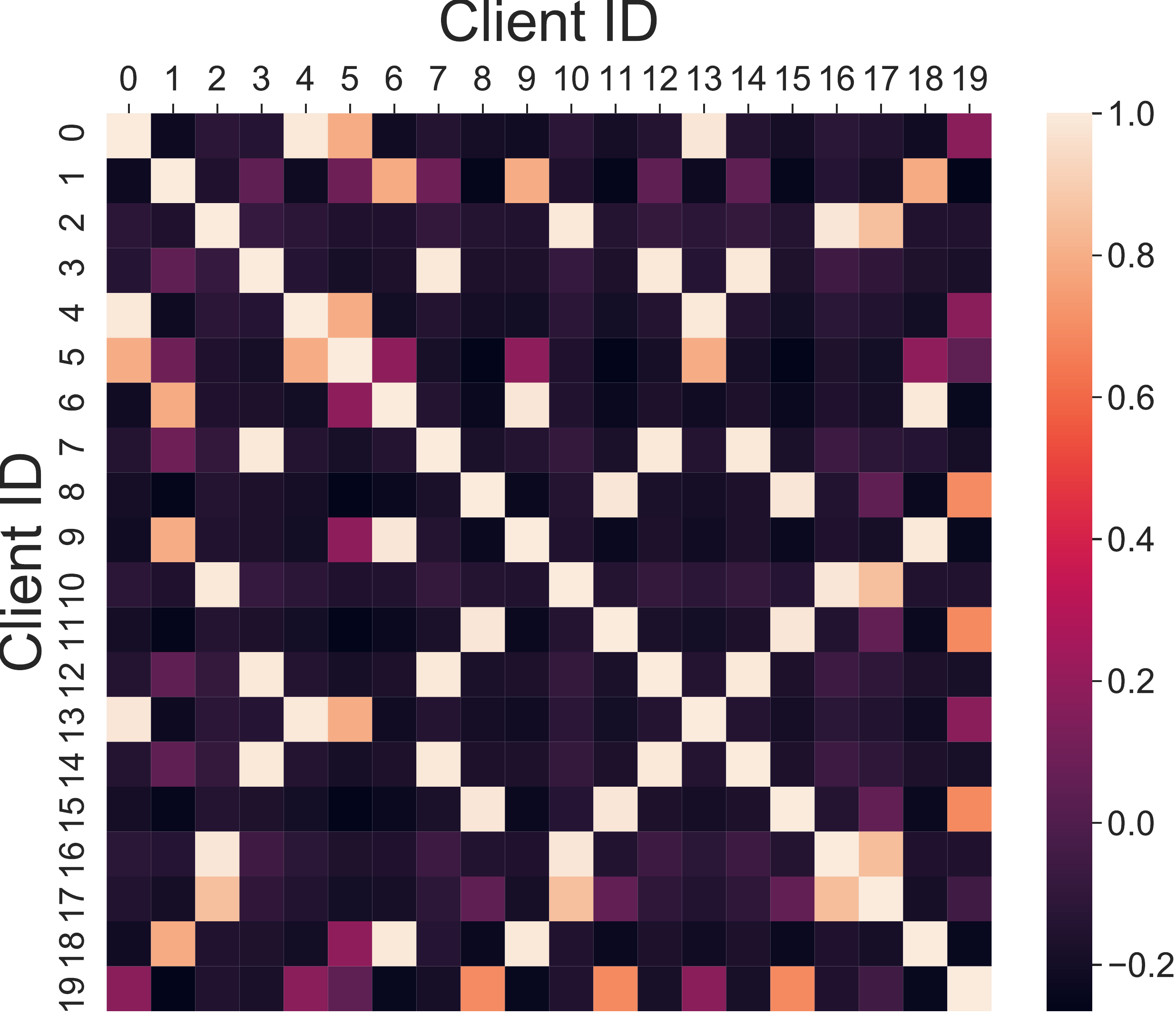}
			\label{fig:mnist_heatmap}
			}
			\vspace{-5mm}
	\caption[]{Cosine similarity between full-gradients of different clients on the same model on \texttt{EMNIST} dataset and the heterogeneous \texttt{MNIST} (with $2$ classes per client) dataset. In the \texttt{EMNIST} dataset each client has a homogeneous correlation with other clients, while the \texttt{MNIST} dataset is highly heterogeneous.}
	\label{fig:heatmap}
\end{figure*}

\paragraph{\texttt{EMNIST} dataset} In addition to the results for the \texttt{MNIST} and \texttt{CIFAR10} datasets we present the results of applying different algorithms on \texttt{EMNIST}~\cite{caldas2018leaf} dataset. This dataset, similar to the \texttt{MNIST} dataset, contains images of characters in $28\times 28$ size. The difference here is that the dataset is separated based on the author of images, hence, the distribution each image is coming from is different for different nodes. In this experiment, we use data from $1000$ authors in the \texttt{EMNIST} dataset, and set the sampling ratio $k=0.1$. Also, we tune the learning rate to the fixed value of $0.01$ for all the algorithms. The model, similar to the \texttt{MNIST} case, is a 2-layer MLP with 200 neurons for each hidden layer and ReLU activations. Figure~\ref{emnist1000} shows the results of this experiment for the training loss and testing accuracy based on the size of communication. It can be inferred that \texttt{FedCOMGATE} and FedPAQ both have the fastest convergence based on the communication size, and accordingly, wall-clock time. The reason that the final convergence rate is the same for all algorithms is that similar to Figure~\ref{fig:comp_mnist_iid}, this dataset is close to the homogeneous setting. To show that, and to compare it to the heterogeneous datatset we created using the \texttt{MNIST} dataset, we run a test on $20$ different clients of this dataset and the heterogeneous \texttt{MNIST} dataset ($2$ classes data per client). We give all the clients the same model and perform a full batch gradient computation over that model. Then, we compute the cosine similarity of this gradients using:
\begin{equation}
    d_{ij} = \frac{\mathbf{g}_i^\top \mathbf{g}_j}{\left\lVert \mathbf{g}_i \right\rVert\cdot \left\lVert \mathbf{g}_j \right\rVert}.
\end{equation}
Figure~\ref{fig:heatmap} shows the heatmap of these correlations among clients for two datasets. As it can be seen, in the \texttt{EMNIST} dataset, each client's data homogeneously correlates with all other clients'. However, in the \texttt{MNIST} dataset (with $2$ classes data per client), each client has high correlation with at most $4$ clients and not correlated or has a negative correlation with other clients' data. This shows that the level of heterogeneity in the \texttt{EMNIST} dataset is much lower than that of in the \texttt{MNIST} dataset, and hence, the result in the Figure~\ref{emnist1000} are in line with our theoretical findings for the gradient tracking technique.

\section{Conclusion}
\label{sec:concl&fud}
In this paper we introduced a set of algorithms for federated learning which lower the communication overhead by periodic averaging and exchanging compressed signals. We considered two separate settings: (i) homogeneous setting in which all the probability distributions and loss functions are identical; and (ii) heterogeneous setting wherein the users' distributions and loss functions could be different. For both cases, we showed that our proposed methods both theoretically and numerically require less communication rounds between server and users compared to state-of-the-art federated algorithms that use compression.

\section*{Acknowledgment}
The authors would like to thank Amirhossein Reisizadeh for his comments on the first draft of the paper. We also gratefully acknowledge the generous support of NVIDIA for providing GPUs for our research. This work has been done using the Extreme Science and Engineering Discovery Environment (XSEDE) resources, which is supported by National Science Foundation under grant number
ASC200045.

\newpage

\appendix
\begin{center}
{\bf{\LARGE Appendix}}
\end{center}

The outline of our supplementary material follows. In Section~\ref{sec:RW}, we first elaborate further on related studies in the literature. In Section~\ref{app:variant}, we propose variations of Algorithm~\ref{Alg:VRFLDL} used in the experimental setup. Then, we present the proofs of our main theoretical results presented in the main body of the paper. In Section~\ref{sec:app:sgd:undrr-pl}, we present the convergence properties of our \texttt{FedCOM} method presented in Algorithm~\ref{Alg:one-shot-using data samoples-b} for the \textbf{homogeneous} setting. In Section~\ref{app_het_sec}, we present the convergence properties of our \texttt{FedCOMGATE} method presented in Algorithm~\ref{Alg:VRFLDL} for the \textbf{heterogeneous} setting. In Section~\ref{sec:appendix:omitted}, we present the proof of some of our intermediate lemmas. 
\appendixwithtoc

\newpage
\section{Additional Related Work}\label{sec:RW}
In this section, we summarize and discuss additional related work. We separate the related work into two broad categories below. 

\paragraph{Local computation with periodic communication.} An elegant idea to reduce the number of communications in vanilla synchronous SGD is to perform averaging periodically instead of averaging models in all clients at every iteration~\cite{zhang2016parallel}, also known as local SGD. The seminal work of~\cite{stich2018local} was among the first to analyze the convergence of local SGD in the homogeneous setting and demonstrated that the number of communication rounds can be significantly reduced for smooth and strongly convex objectives while achieving linear speedup. This result is further improved in follow up studies~\cite{haddadpour2019local,khaled2020tighter,wang2018cooperative,stich2019error,yu2019parallel,haddadpour2019trading}. In~\cite{wang2018cooperative}, the error-runtime trade-off of local SGD is analyzed and it has been shown that it can also alleviate the synchronization delay caused by slow workers. From a practical viewpoint, few recent efforts explored adaptive communication strategies to communicate more frequently early in the process~\cite{haddadpour2019local,lin2018don,scholkopf2002learning}.

The analysis of local SGD in the heterogeneous setting, also known as federated averaging (FedAvg)~\cite{mcmahan2016communication}, has seen a resurgence of
interest very recently. While it is still an active research area~\cite{woodworth2020local},  a few number of recent studies  made efforts to understand  the convergence of local SGD in a heterogeneous setting~\cite{zhou2018convergence,haddadpour2019convergence,karimireddy2019scaffold,koloskova2020unified,li2019convergence,khaled2020tighter}. Also, the personalization of local models for a better generalization in a heterogeneous setting is of great importance from both theoretical and practical point of view~\cite{smith2017federated,deng2020adaptive,fallah2020personalized,mansour2020three,lin2020collaborative}.

\paragraph{Distributed optimization with compressed communication.}Another parallel direction of research has focused on reducing the size of communication by compressing the communicated messages. In quantization based methods, e.g.~\cite{seide20141,lin2018deep,stich2019error,basu2019qsparse}, a quantization operator is applied before transmitting the gradient to server. A gradient acceleration approach with compression is proposed in~\cite{li2020acceleration}. In heterogeneous data distribution,~\cite{jin2020stochastic} proposed the use of sign based SGD algorithms and~\cite{reisizadeh2019fedpaq} employed a quantization scheme in FedAvg with provable guarantees. In sparsification based methods, the idea is to transmit a smaller gradient vector by keeping only very few coordinates of local stochastic gradients, e.g., most significant entries~\cite{aji2017sparse,lin2017deep}. For these methods, theoretical guarantees have been provided in a few recent efforts~\cite{alistarh2018convergence,tang2018communication,wu2018error,stich2018sparsified}. Note that,  most of these studies rely on an error compensation technique as we employ in our experiments. We note that sketching methods are also employed to reduce the number of communication in~\cite{ivkin2019communication}.

The aforementioned studies mostly fall into the centralized distribution optimization. Recently a few attempts are made to explore the compression schema in a decentralized setting where each device shares compressed messages with direct neighbors over the underlying communication network~~\cite{reisizadeh2018quantized,reisizadeh2019robust,singh2020squarm,koloskova2019decentralized}. Another interesting direction for the purpose of reducing the communication complexity is to exploit the sparsity of communication network as explored in~\cite{koloskova2020unified,wang2019matcha,haddadpour2019convergence}.

Finally, more thorough related works that study federated learning from different perspectives can be found in \cite{kairouz2019advances} and \cite{li2020federated}.


\section{Further Experimental Studies and Results}\label{app:aer}
In this section, we present additional experimental results, as well as details, that will further showcase the efficacy of the proposed algorithms in the paper. First, we should elaborate on the algorithms we used in Section~\ref{sec:exp}, but due to lack of space we did not describe in the main body. In addition, we introduce a  version of Algorithm~\ref{Alg:VRFLDL} without compression, and its version with sampling of clients.

\subsection{Variations of Algorithm~\ref{Alg:VRFLDL}}\label{app:variant}
In this section we describe the details of  variants of Algorithm~\ref{Alg:VRFLDL} that are used in experiments. 

\begin{algorithm2e}[t!]
\DontPrintSemicolon
\caption{\texttt{FedGATE}($R, \tau, \eta, \gamma$) Federated Averaging with Local Gradient Tracking}
\SetNoFillComment
\label{Alg:LGT}
\SetKwFor{ForPar}{for}{do in parallel}{end forpar}
\textbf{Inputs:} Number of communication rounds $R$, number of local updates $\tau$, learning rates $\gamma$ and $\eta$, initial  global model $\boldsymbol{w}^{(0)}$, initial gradient tracking $\mathbf{\delta}_j^{(0)}=\boldsymbol{0}, \; \forall j\in[m]$\\[5pt]

\For{$r=0, \ldots, R-1$}{
    \ForPar{each client $j\in[m]$}{
         Set $\boldsymbol{w}_j^{(0,r)}={\boldsymbol{w}}^{(r)}$\\ 
        \For{ $c=0,\ldots,\tau-1$}{
             Set $\tilde{\boldsymbol{d}}^{(c,r)}_{j}=\tilde{\mathbf{g}}_{j}^{(c,r)}-\delta_j^{(r)}$ where $\tilde{\mathbf{g}}_{j}^{(c,r)}\triangleq\nabla f_j (\boldsymbol{w}^{(c,r)}_j;\mathcal{Z}_j^{(c,r)})$\\
             $\boldsymbol{w}^{(c+1,r)}_{j}=\boldsymbol{w}^{(c,r)}_j-\eta~ \tilde{\boldsymbol{d}}^{(c,r)}_{j}$
        }
        Device \textbf{sends} $ \boldsymbol{u}_j^{(r)} = {\boldsymbol{w}^{(r)}-\boldsymbol{w}^{(\tau,r)}_j}$  back to the server and gets $\boldsymbol{u}^{(r)}$\\ 
        Device \textbf{computes} $\bar{\boldsymbol{w}}^{(r)}=\boldsymbol{w}^{(r)}-\boldsymbol{u}^{(r)}$\\
        Device \textbf{updates}  $\delta_j^{(r+1)}=\delta_j^{(r)}+\frac{1}{\eta\tau}\left(\bar{\boldsymbol{w}}^{(r)}-\boldsymbol{w}_j^{(\tau,r)}\right)$ \\
        \algemph{GreenYellow}{0.9}{Device \textbf{updates} server model $\boldsymbol{w}^{(r+1)}=\boldsymbol{w}^{(r)}-\gamma\boldsymbol{u}^{(r)}$ \mycommfont{\tcp*{Option I}}}\\ 
    }
   Server \textbf{computes} $\boldsymbol{u}^{(r)}=\frac{1}{m}\sum_{j=1}^m \boldsymbol{u}_j^{(r)}$ and \textbf{broadcasts} back to clients\\
     \algemph{classicrose}{.94}{Server \textbf{updates} $\boldsymbol{w}^{(r+1)}=\boldsymbol{w}^{(r)}-\gamma\boldsymbol{u}^{(r)}$  \\
    Server \textbf{broadcasts} ${\boldsymbol{w}}^{(r+1)}$ to all devices \mycommfont{\tcp*{Option II}}} 
}
\end{algorithm2e}

\paragraph{Without compression.} In this part, we first elaborate on a variant of Algorithm~\ref{Alg:VRFLDL} without any compression involved, which we call it Federated Averaging with Local Gradient Tracking, \texttt{FedGATE}. Algorithm~\ref{Alg:LGT} describes the steps of \texttt{FedGATE}, which involves a local gradient tracking step. This algorithm is similar to the SCAFFOLD~\cite{karimireddy2019scaffold}, however, the main difference is that we do not use any server control variate. In fact, \texttt{FedGATE}, as well as \texttt{FedCOMGATE}, are implicitly controlling the variance of the server model by controlling its subsidiaries' variances in local models. Therefore, there is no need to have another variable for this purpose, which can help us to greatly reduce the communication size, to half of what SCAFFOLD is using. Hence, even in the simple algorithm of \texttt{FedGATE}, we can gain the same convergence rate as SCAFFOLD, while enjoying the $2\times$ speedup in the communication. Note that, since the communication time of broadcasting from server to clients (or downlink communication) is negligible compared to gathering from clients to the server (or uplink communication), the overall communication complexity of this algorithm is close to FedAvg, and half of the SCAFFOLD, as it is depicted in Figure~\ref{fig:comp_mnist_3cpu_l1_time}. Also, the communication complexity of \texttt{FedCOMGATE} is close to that of FedPAQ~\cite{reisizadeh2019fedpaq}.

The common approach in federated learning without sampling for \texttt{FedGATE} and \texttt{FedCOMGATE} would be similar to \texttt{Option II} in Algorithm~\ref{Alg:LGT}, where the server updates its model and broadcasts it to clients. This approach has one extra downlink step, that is negligible compared to the uplink steps, as it was mentioned before. However, when there is no sampling of the clients, we can avoid this extra downlink by using the \texttt{Option I}, where each local device keeps track of the server model and updates it based on what it gets for updating the gradient tracking variable. In practice, when sampling is not involved, we use \texttt{Option I}. In Section~\ref{sec:exp}, we compare the performance of \texttt{FedGATE} and SCAFFOLD.

\paragraph{User sampling.} One important aspect of federated learning is the sampling of clients since they might not be available all the time. Also, sampling clients can further reduce the per round communication complexity by aggregating information from a subset of clients instead of all clients. Hence, in Algorithm~\ref{Alg:CLGTS}, we incorporate the sampling mechanism into our proposed \texttt{FedCOMGATE} algorithm. Based on this algorithm, at each communication round, the server selects a subset of clients  $\mathcal{S}^{(r)}\subseteq [m]$, and sends the global server model only to selected devices in $\mathcal{S}^{(r)}$. The remaining steps of the algorithm are similar to Algorithm~\ref{Alg:VRFLDL}. In Section~\ref{sec:exp}, we also study the effect of user sampling on the performance of \texttt{FedGATE}, \texttt{FedCOMGATE}, and other state-of-the-art methods for federated learning.

\begin{algorithm2e}[t!]
\DontPrintSemicolon
\SetNoFillComment
\LinesNumbered
\caption{\texttt{FedCOMGATE}($R, \tau, \eta, \gamma, k$), \texttt{FedCOMGATE} algorithm with sampling of clients }\label{Alg:CLGTS}
\SetKwFor{ForPar}{for}{do in parallel}{end forpar}
\textbf{Inputs:} Number of communication rounds $R$, number of local updates $\tau$, learning rates $\gamma$ and $\eta$, initial  global model $\boldsymbol{w}^{(0)}$,  participation ratio of clients $k \in (0,1]$, initial gradient tracking $\mathbf{\delta}_j^{(0)}=\boldsymbol{0}, \; \forall j\in[m]$\\[2pt]
\For{$r=0, \ldots, R-1$}{
    Sever \textbf{selects} a subset of devices $\mathcal{S}^{(r)}\subseteq [m]$, with the size $\floor{km}$\\
    Server \textbf{broadcasts} ${w}^{(r)}$ to the selected devices $ j \in \mathcal{S}^{(r)}$\\[3pt]
    \ForPar{each client $j\in\mathcal{S}^{(r)}$}{
        Set $\boldsymbol{w}_j^{(0,r)}={\boldsymbol{w}}^{(r)}$ \\
        \For{ $c=0,\ldots,\tau-1$}{
            Set $\tilde{\boldsymbol{d}}^{(c,r)}_{j,q}=\tilde{\mathbf{g}}_{j}^{(c,r)}-\delta_j^{(r)}$ where $\tilde{\mathbf{g}}_{j}^{(c,r)}\triangleq\nabla f_j (\boldsymbol{w}^{(c,r)}_j;\mathcal{Z}_j^{(c,r)})$\\
             $\boldsymbol{w}^{(c+1,r)}_{j}=\boldsymbol{w}^{(c,r)}_j-\eta~ \tilde{\boldsymbol{d}}^{(c,r)}_{j,q}$
        }
        Device \textbf{sends} $  \boldsymbol{\Delta}_{j,q}^{(r)} = Q(({\boldsymbol{w}^{(r)}-\boldsymbol{w}^{(\tau,r)}_j})/\eta)$  back to the server and gets $\boldsymbol{\Delta}_{q}^{(r)}$\\
        Device \textbf{updates}  $\delta_j^{(r+1)}=\delta_j^{(r)}+\frac{1}{\tau}( \boldsymbol{\Delta}_{j,q}^{(r)}-\boldsymbol{\Delta}_{q}^{(r)})$
    }
Server  \textbf{computes} $\boldsymbol{\Delta}_q^{(r)}=\frac{1}{m}\sum_{j=1}^m \boldsymbol{\Delta}_{j,q}^{(r)}$ and \textbf{broadcasts} back to devices $ j \in \mathcal{S}^{(r)}$\\
    Server \textbf{computes} $\boldsymbol{w}^{(r+1)}=\boldsymbol{w}^{(r)}-\eta\gamma\boldsymbol{\Delta}_q^{(r)}$ \\
}
\end{algorithm2e}

\subsection{Additional Experiments}\label{app:add_exp}
\begin{figure*}[t!]
		\centering
		\subfigure[MNIST Dataset]{
		\centering
		\includegraphics[width=0.48\textwidth]{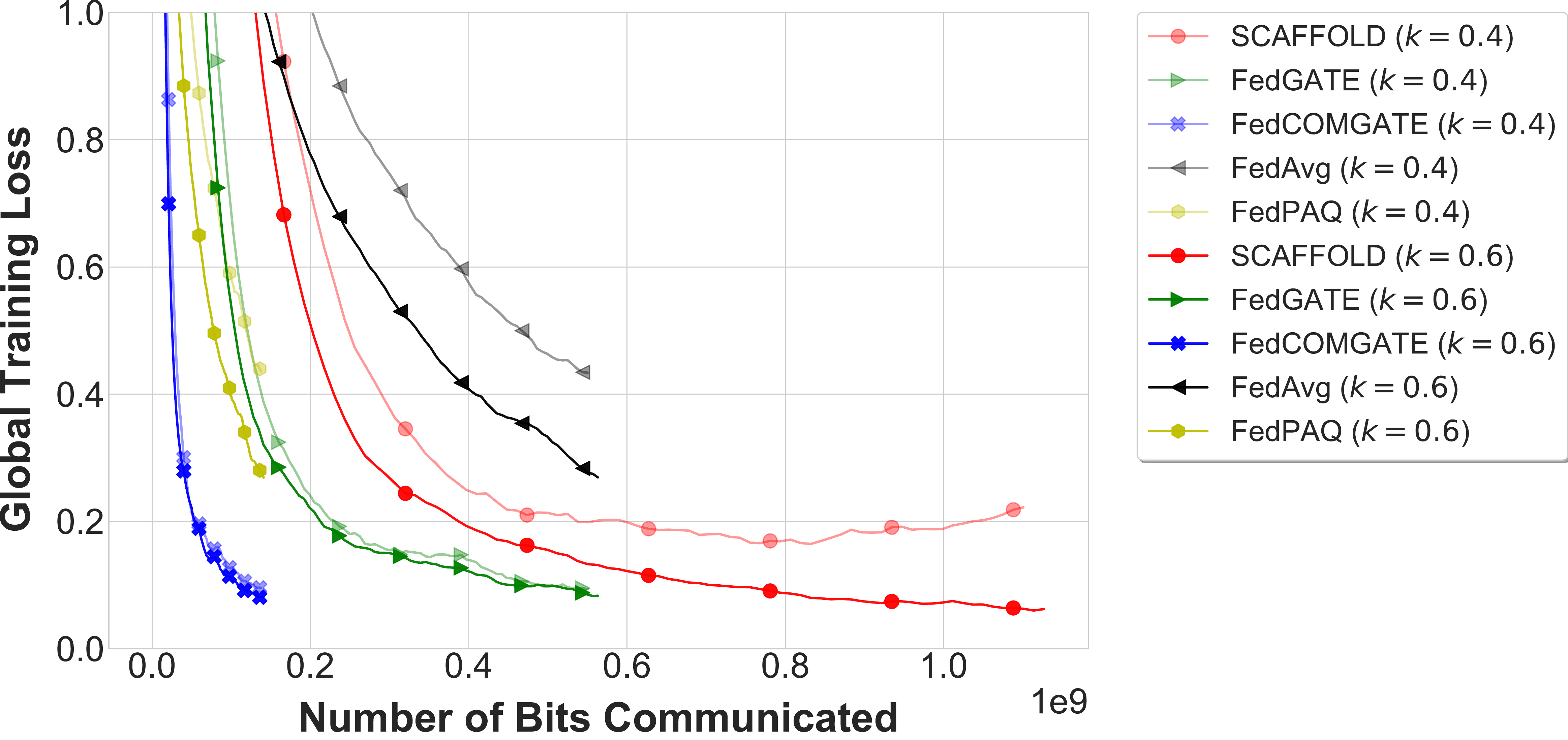}
		\label{fig:sample_comp_fashion_sim_l10_comm_time}
		}
		\hfill
		\subfigure[CIFAR10 Dataset]{
			\centering 
			\includegraphics[width=0.48\textwidth]{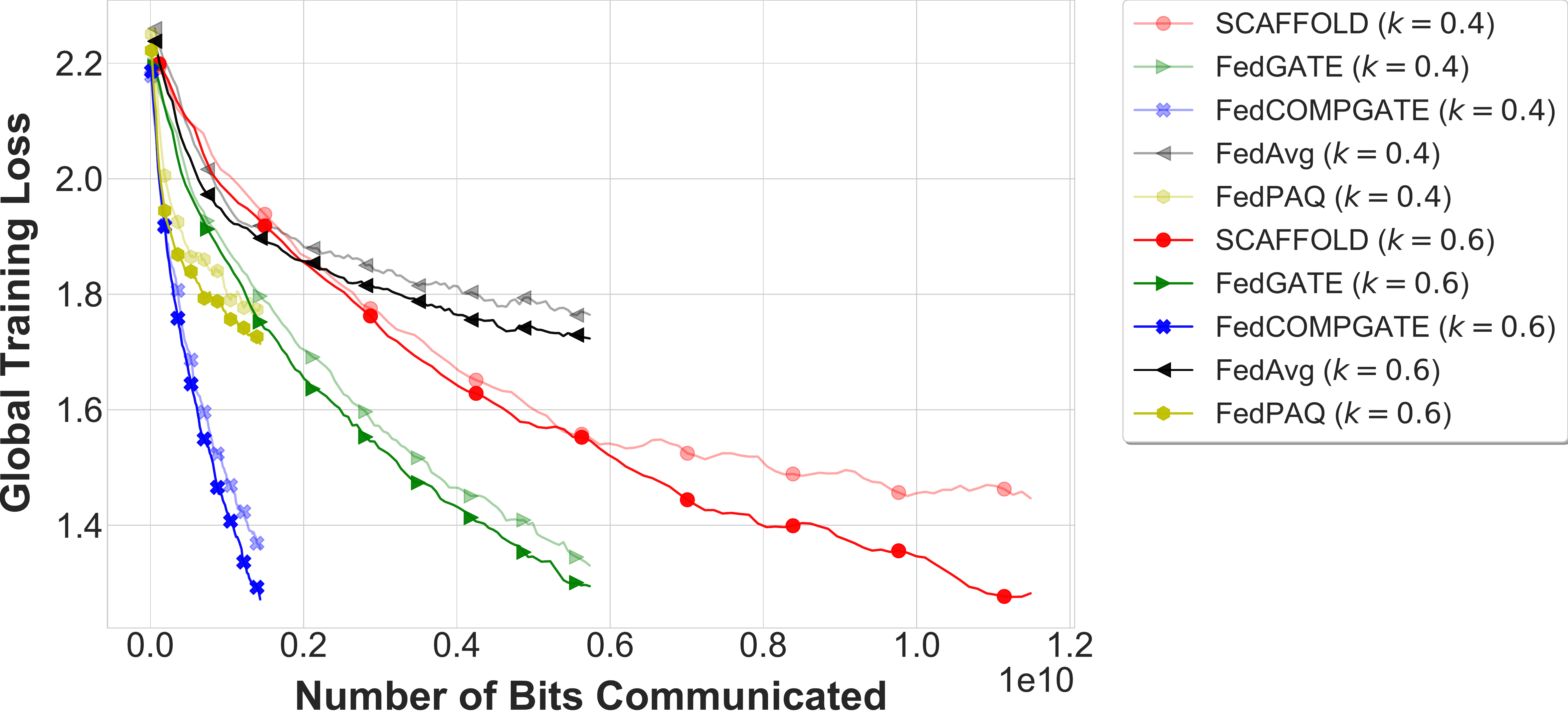}
			\label{fig:sample_comp_cifar10_sim_l10_comm_time}
			}
			\vspace{-2mm}
	\caption[]{Comparing the effect of sampling on different algorithms. We use two datasets: the \texttt{MNIST} and the \texttt{CIFAR10} datasets. We use an MLP with 2 layers for all the datasets, with $200$ neurons per layer for the \texttt{MNIST}, and $500$ neurons per layer for the \texttt{CIFAR10}. \texttt{FedGATE} and \texttt{FedCOMGATE} seem to be more robust against client sampling.}
	\label{sampl_comp_sim_l10}
\end{figure*}

\paragraph{Sampling clients} In this section we assume that only $k\in (0,1]$ portion of the users in the networks are active and exchange information with the server at each round. Indeed, a lower value of $k$ implies that less nodes are active at each round and therefore the communication overhead is lower. However, it could possibly lead to a slower convergence rate and extra communications rounds to achieve a specific accuracy. We formally study the effect of $k$ on the convergence of \texttt{FedCOMGATE} and its version without compression \texttt{FedGATE} and compare their performance with other federated methods with and without compression in Figure~\ref{sampl_comp_sim_l10}. As it can be inferred, when we decrease the $k$ or the participation rate, generally, the performance of the model degrades with the same number of communication rounds. However, the amount of degradation might vary among different algorithms. As it is depicted in Figure~\ref{sampl_comp_sim_l10}, the proposed \texttt{FedCOMGATE} algorithm and its unquantized version, \texttt{FedGATE}, are quite robust against decreasing the participation rate between clients with respect to other algorithms such as SCAFFOLD and FedPAQ.

{
An important difference between algorithms proposed in this work (such as \texttt{FedCOMGATE}) and SCAFFOLD is in their performance when not all clients are active at each round of communication. Since in SCAFFOLD gradient tracking needs to be performed both locally and globally at each round of communication, its performance could highly depend on the availability of control variate in all devices. This can lead to poor performance for SCAFFOLD when the rate of participation of clients at each round is low (as illustrated in Figure~\ref{sampl_comp_sim_l10}). This could be due to stale local control variate and new global control variate that degrade the performance of the algorithm. On the other hand, in our proposed algorithms, the gradient tracking parameter is only performed locally, and hence, the drop in the performance is much smaller than SCAFFOLD. Therefore, we think this property makes our proposed algorithms suitable for the cross-device scenarios as well as cross-silo ones, whereas SCAFFOLD is more suitable for cross-silo scenarios, and not cross-device ones.
}

\begin{figure*}[t!]
		\centering
		\subfigure[MNIST Dataset]{
		\centering
		\includegraphics[width=0.4\textwidth]{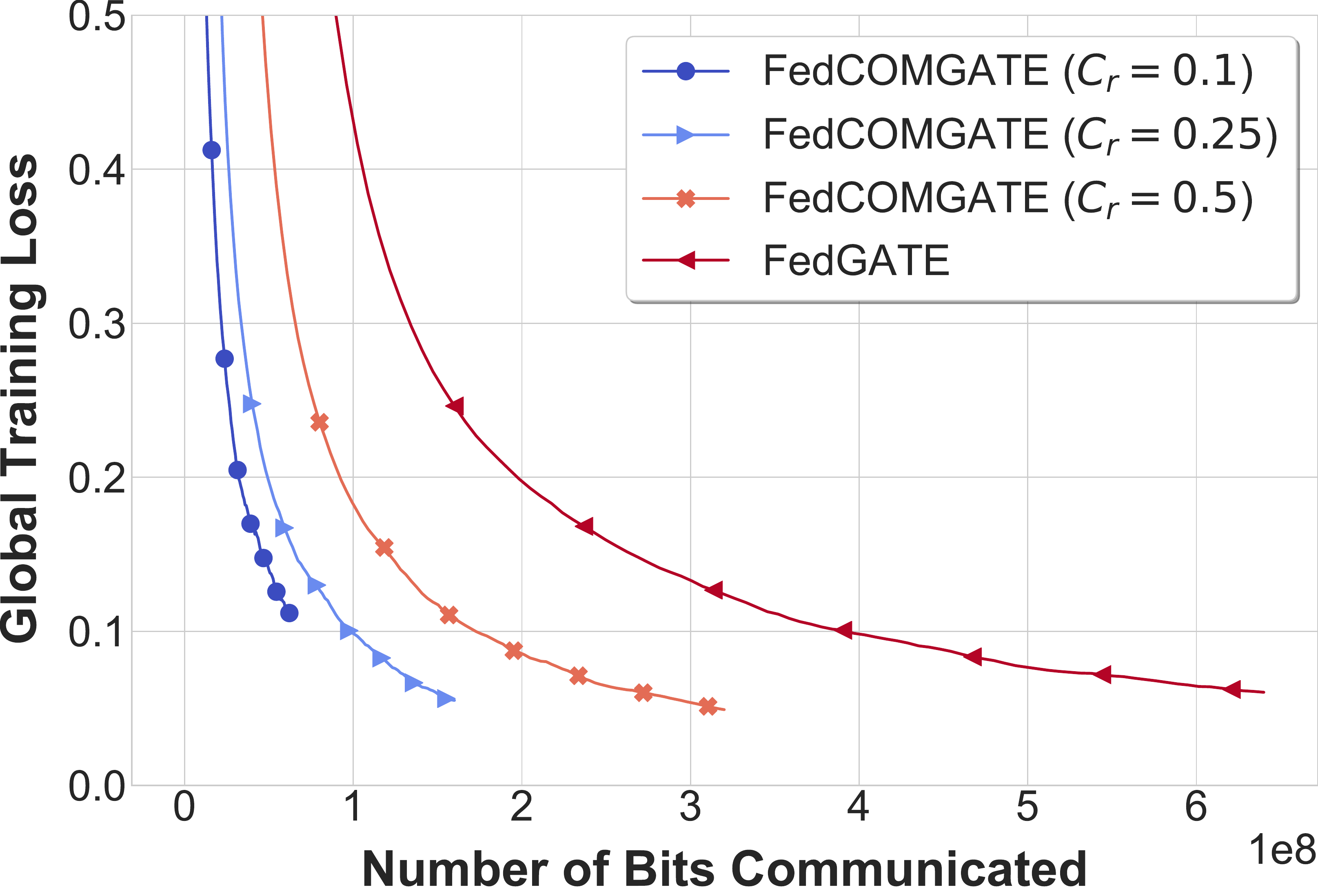}
		\label{fig:sparse_comp_mnist_sim_l10_comm_time}
		}
		\hspace{1cm}
		\subfigure[Fashion MNIST Dataset]{
			\centering 
			\includegraphics[width=0.4\textwidth]{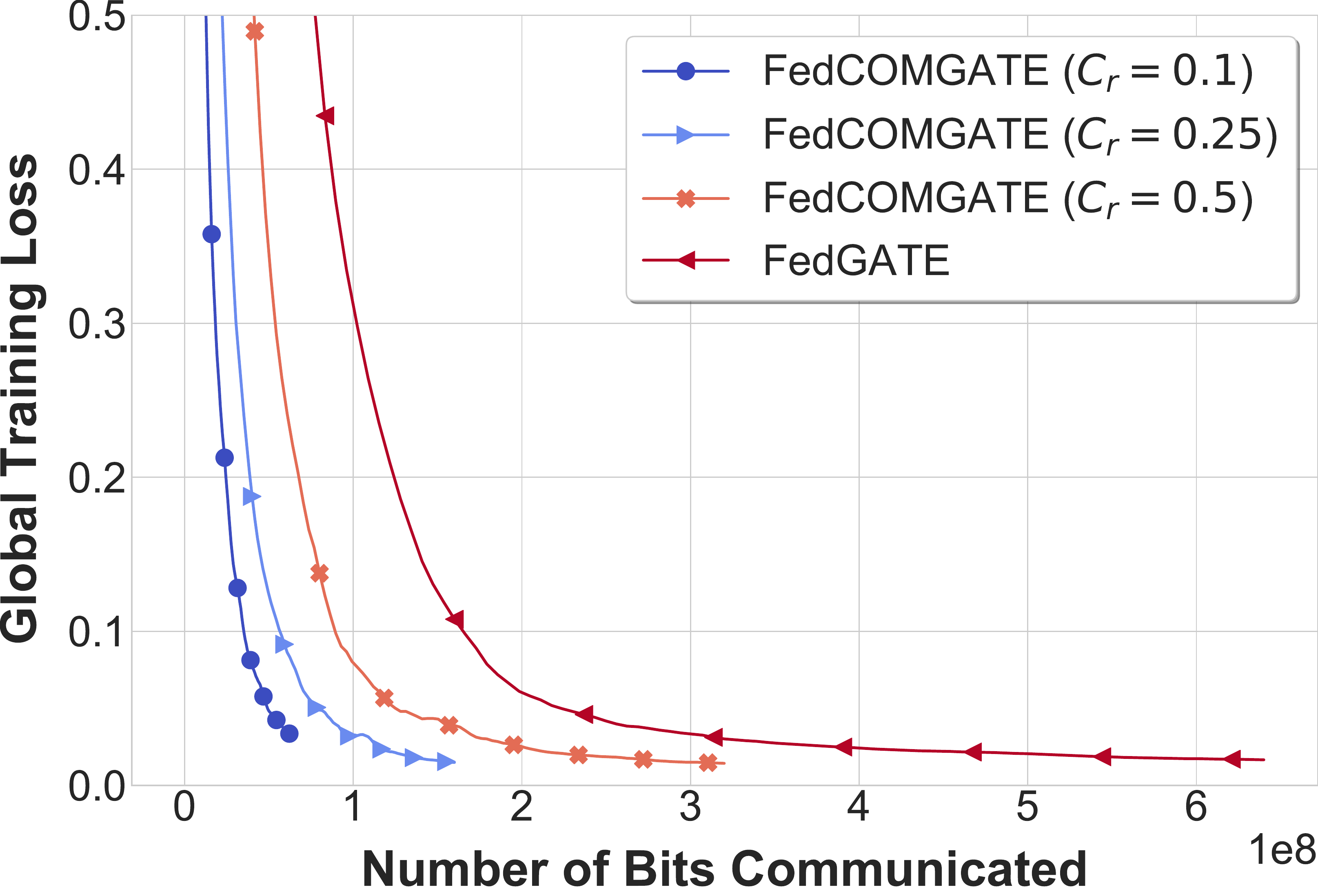}
			\label{fig:sparse_comp_fashion_sim_l10_comm_time}
			}
	\caption[]{The effect of sparsification with memory on the \texttt{FedCOMGATE} algorithm used for the training of the \texttt{MNIST} and the \texttt{Fashion MNIST} datasets. We can achieve almost similar results as the algorithm without compression (\texttt{FedGATE}) with some compression rates. Decreasing the size of communication will speed up the training, in the cost of increasing a residual error as it is evident for the case with $C_r = 0.1$.}
	\label{fig:sparse}
\end{figure*}

\paragraph{Compression via sparsification.} Another approach to compress the gradient updates is sparsification. This method has been vastly used in distributed training of machine learning models~\cite{stich2018sparsified,aji2017sparse,wangni2018gradient}. Using a simple sparsification by choosing random elements or $top_k$ elements, some information will be lost in aggregating gradients, and consequently, the quality of the model will be degraded. To overcome this problem, an elegant idea is proposed in~\cite{stich2018local} to use memory for tracking the history of entries and avoid the accumulation of compression errors. Similarly, we will employ a memory of aggregating gradients in order to compensate for the loss of information from sparsification. This is in addition to the local gradient tracking we incorporated in \texttt{FedCOMGATE}, however, despite the server control variate in SCAFFOLD, this memory is updated locally and is not required to be communicated to the server. We denote the memory in each client $j$ at round $r$ with $\boldsymbol{\nu}_{j}^{(r)}$. Thus, in Algorithm~\ref{Alg:VRFLDL}, we first need to compress the gradients added by the memory, using the $top_k$ operator as:
\begin{equation}
    \boldsymbol{\Delta}_{j,s}^{(r)} = top_k\left\{\left(\boldsymbol{w}^{(r)}-\boldsymbol{w}^{(\tau,r)}_j\right) + \boldsymbol{\nu}_{j}^{(r)}\right\}
\end{equation}
Then, we will send this to the server for aggregation, where the server decompresses them, takes the average, and sends  $\boldsymbol{\Delta}^{(r)}$ back to the clients. Each client updates its gradient tracking parameter as in Algorithm~\ref{Alg:VRFLDL}. Also, in this case, we need to update the memory parameter as:
\begin{equation}
    \boldsymbol{\nu}_{j}^{(r+1)} = \boldsymbol{\nu}_{j}^{(r)} + \frac{1}{m} \left(\boldsymbol{w}^{(r)}-\boldsymbol{w}^{(\tau,r)}_j\right) - \boldsymbol{\Delta}^{(r)},
\end{equation}
where it keeps track of what was not captured by the aggregation using the sparsified gradients. Note that, unlike the quantized \texttt{FedCOMGATE}, in this approach, we cannot compress the downlink gradient broadcasting. However, since the cost of broadcasting is much lower than the uplink communication, this is negligible, especially in lower compression rates compared to quantized \texttt{FedCOMGATE}.

To show how the \texttt{FedCOMGATE} using sparsification with memory works in practice we will apply it to \texttt{MNIST} and \texttt{Fashion MNIST} datasets. Both of them are applied to an MLP model with two hidden layers, each with $200$ neurons. For this experiment, we use the compression ratio parameter of $C_r$, which is the ratio between the size of communication in the compressed and without compression versions. Figure~\ref{fig:sparse} shows the result of this algorithm by changing the compression rate. As it was observed by~\cite{stich2018sparsified}, in some compression rates we can have similar or slightly better results than the without compression distributed SGD solution (here \texttt{FedGATE}), due to the use of memory. However, to gain more from the speedup and decreasing the compression rate, we will incur a residual error, as it can be seen in the results for the compression rate of $0.1$.

\begin{figure}[t!]
		\centering
		\subfigure[Effect of quantization noise $q$]{
		\centering
		\includegraphics[width=0.4\textwidth]{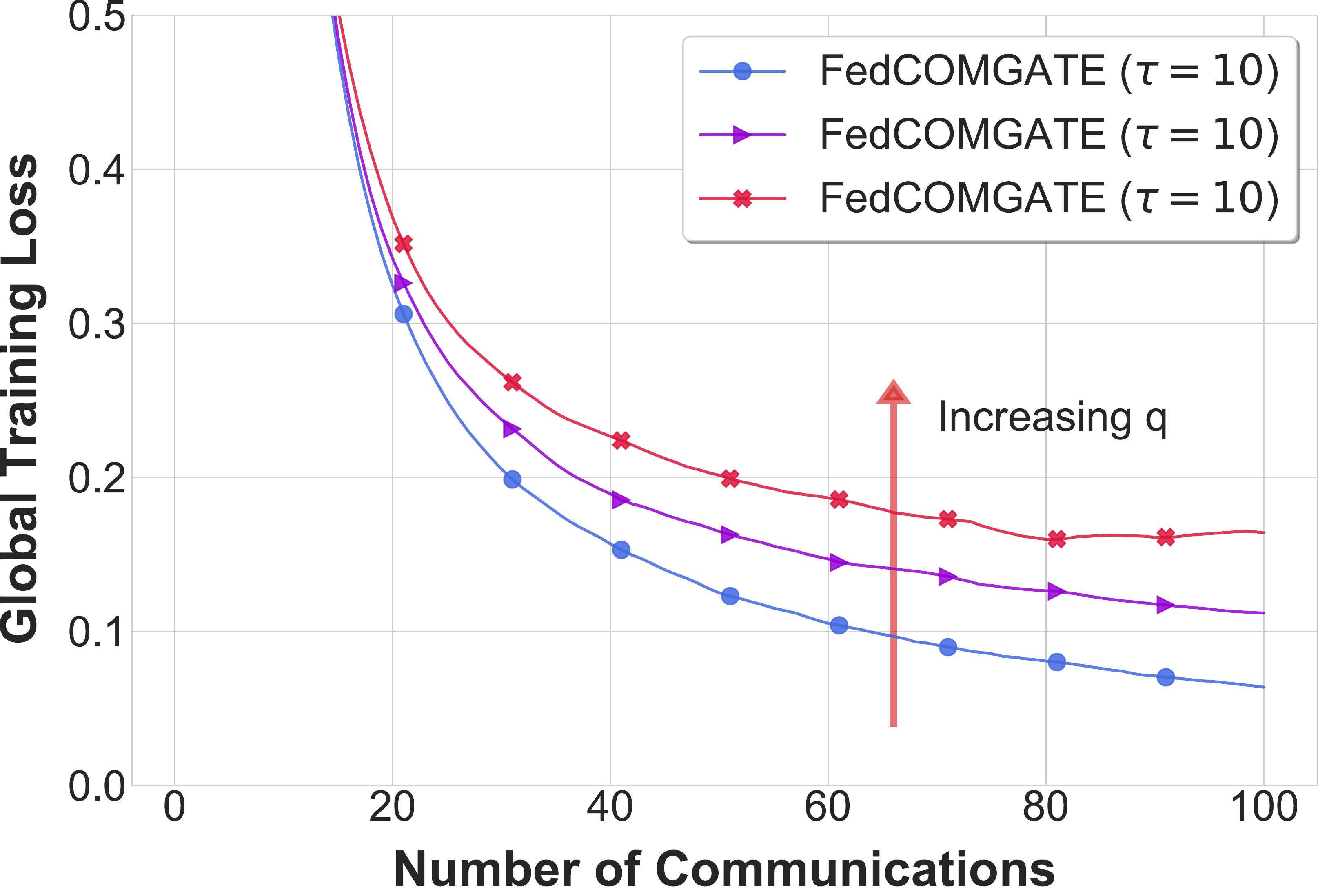}
		\label{fig:q}
		}
		\hspace{1cm}
		\subfigure[Effect of local computation $\tau$]{
			\centering 
			\includegraphics[width=0.4\textwidth]{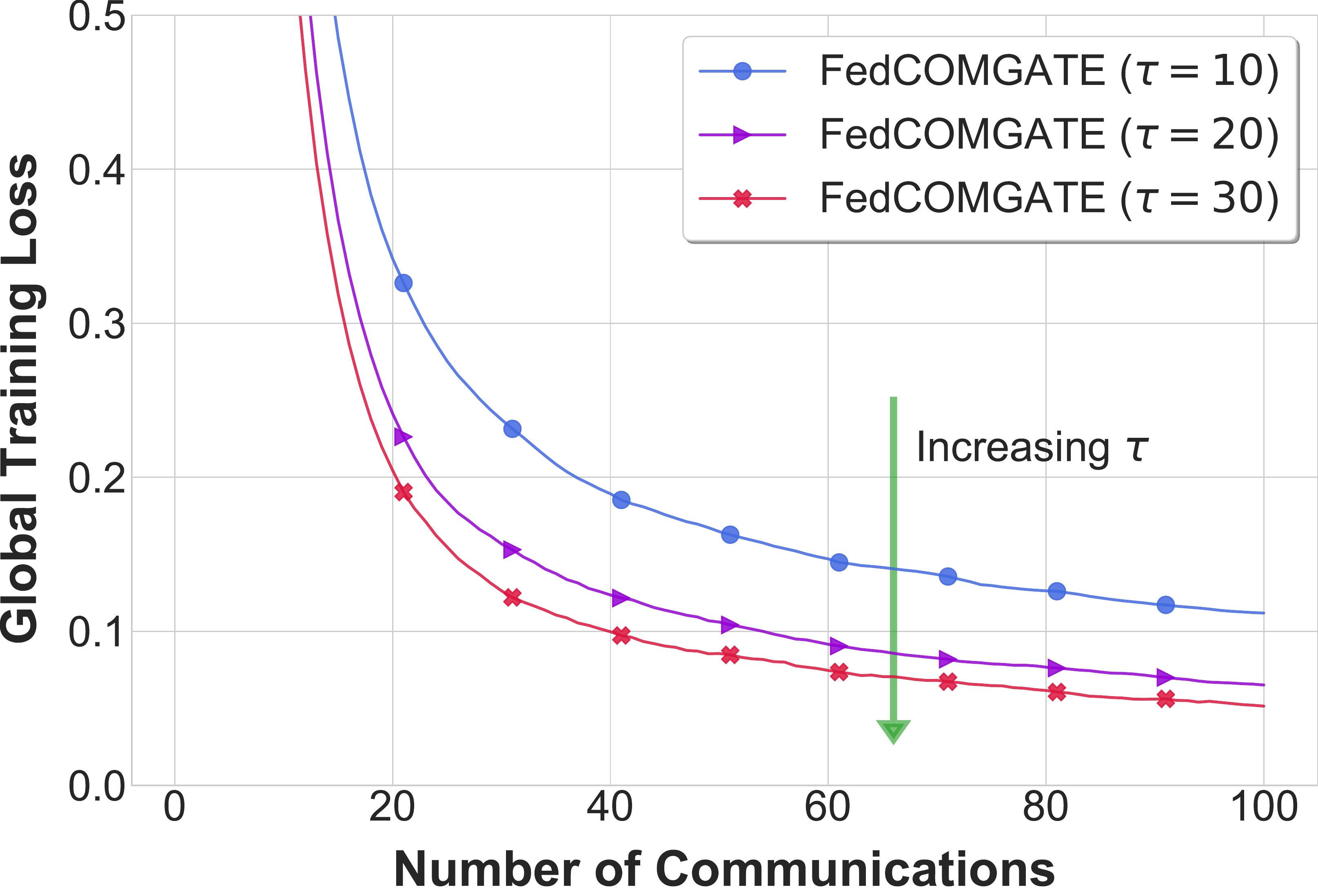}
			\label{fig:tau}
			}
	\caption[]{Investigating the effects of quantization noise and local computations on the convergence rate. We run the experiments on the \texttt{MNIST} dataset with a similar MLP model as before. In (a), we increase the noise of quantization by increasing the range of noise added to the zero-point of the quantizer operator. Increasing $q$ can degrade the convergence rate of the model. On the other hand, in (b), with the same level of quantization noise, we can increase the number of local computations $\tau$ to diminish the effects of quantization.}
	\label{fig:q-tau}
\end{figure}

\paragraph{Effect of local computations.} Finally, we will show the effect of noise in quantization, characterized as $q$ in the paper, on the convergence rate, and how to address it. As it can be inferred from our theoretical analysis, increasing the noise of quantization would degrade the convergence rate of the model. This pattern can be seen in Figure~\ref{fig:q} for the \texttt{MNIST} dataset, where we add noise to quantized arrays by adding a random integer to the zero-point of the quantization operator. By increasing the range of this noise, we can see that the convergence is getting worse with the same number of local computations. On the other hand, based on our analysis, we know that increasing the number of local computations will compensate for the quantization noise, which helps us to achieve the same results with lower communication rounds. This pattern is depicted in Figure~\ref{fig:tau}, where we keep the quantization noise constant and increase the number of local computations.

\newpage
\section{Some Definitions and Notation}\label{app_def_not}

Before stating our proofs we first formally define \pl\ and strongly convex functions.
\begin{assumption}[\pl]\label{assum:pl}
A function $f(\boldsymbol{w})$ satisfies the \pl~ condition with constant $\mu$ if $\frac{1}{2}\|\nabla f(\boldsymbol{w})\|_2^2\geq \mu\big(f(\boldsymbol{w})-f(\boldsymbol{w}^*)\big),\: \forall \boldsymbol{w}\in\mathbb{R}^d $ with $\boldsymbol{w}^*$ is an optimal solution.
\end{assumption}

\begin{assumption}[$\mu$-strong convexity]
A function $f$ is $\mu$-strongly convex if it satisfies
$f(\boldsymbol{u})\geq f(\boldsymbol{v})+\left\langle\nabla{f}(\boldsymbol{v}),\boldsymbol{u}-\boldsymbol{v}\right\rangle+\frac{\mu}{2}\left\|\boldsymbol{u}-\boldsymbol{v}\right\|^2$, for all $\boldsymbol{u},\boldsymbol{v}\in \mathbb{R}^d$.
\end{assumption}

We also introduce some notation for  the clarity in presentation of proofs. Recall that   we use  $\mathbf{g}_i ={\nabla{f}_i(\boldsymbol{w})\triangleq \nabla{f}_i(\boldsymbol{w};\mathcal{S}_i)}$ and $\tilde{\mathbf{g}}_i \triangleq \nabla{f}(\boldsymbol{w};\mathcal{Z}_i)$ for ${1\leq i\leq m}$ to denote the full gradient and stochastic gradient at $i$th data shard, respectively, where $\mathcal{Z}_i \subseteq \mathcal{S}_i$ is a uniformly sampled mini-bath. The corresponding quantities evaluated at $i$th machine's   local solution at  $t$th iteration of optimization $\boldsymbol{w}_i^{(t)}$  are denoted by $\mathbf{g}_i^{(t)}$ and $\tilde{\mathbf{g}}_i^{(t)}$, where we  abuse the notation and use $t = r\tau + c$ to denote the $c$th local update at $r$th round, i.e. $(c, r)$. We also define the following notations   
\begin{equation*}
\begin{aligned}
\boldsymbol{w}^{(t)} &= \{\boldsymbol{w}^{(t)}_1,  \ldots, \boldsymbol{w}^{(t)}_m\},\\
{\xi}^{(t)}&=\{{\xi}^{(t)}_1, \ldots, {\xi}^{(t)}_m\},
\end{aligned}
\end{equation*}
to denote the set of local solutions  and sampled mini-batches at iteration $t$ at different machines, respectively. Finally, we use notation $\mathbb{E}[\cdot]$ to denote the conditional expectation $\mathbb{E}_{{\xi}^{(t)}|{\boldsymbol{w}}^{(t)}}
[\cdot]$. 

\newpage
\section{Results for the Homogeneous Setting}
\label{sec:app:sgd:undrr-pl}
In this section, we study the convergence properties of our  \texttt{FedCOM} method presented in Algorithm~\ref{Alg:one-shot-using data samoples-b}. Before stating the proofs for \texttt{FedCOM} in the homogeneous setting, we first mention the following intermediate lemmas.

\begin{lemma}\label{lemma:tasbih1-iid}
Under Assumptions~\ref{Assu:09} and \ref{Assu:1.5}, we have the following bound: 
\begin{align}
\mathbb{E}_{{Q,\xi^{(r)}}}\Big[\|\tilde{\mathbf{g}}_Q^{(r)}\|^2\Big]&=\mathbb{E}_{{\xi}^{(r)}}\mathbb{E}_{{Q}}\Big[\|\tilde{\mathbf{g}}_Q^{(r)}\|^2\Big]\leq \tau(q+1)\frac{1}{m}\sum_{j=1}^m\sum_{c=0}^{\tau-1}\|\mathbf{g}_j^{(c,r)}\|^2+\left(q+1\right)\frac{\tau\sigma^2}{m}  \label{eq:lemma1}
\end{align}
\end{lemma}

\begin{proof}
\begin{align}
&\mathbb{E}_{{\xi^{(r)}|\boldsymbol{w}^{(r)}}}\mathbb{E}_{{Q}}\Big[\|\frac{1}{m}\sum_{j=1}^m Q\left(\sum_{c=0}^{\tau-1}\tilde{\mathbf{g}}^{(c,r)}_j\right)\|^2\Big]\nonumber\\
&=\mathbb{E}_{{\xi}^{(r)}}\left[\mathbb{E}_{{Q}}\Big[\|\frac{1}{m}\sum_{j=1}^m\underbrace{Q\left(\overbrace{\sum_{c=0}^{\tau-1}\tilde{\mathbf{g}}^{(c,r)}_j}^{\tilde{\mathbf{g}}_j^{(r)}}\right)}_{\tilde{\mathbf{g}}_{Qj}^{(r)}}\|^2\Big]\right]\nonumber\\
&\stackrel{\text{\ding{192}}}{=}\mathbb{E}_{{\xi}^{(r)}}\left[\mathbb{E}_{{Q}}\left[\left[\|\frac{1}{m}\sum_{j=1}^m\tilde{\mathbf{g}}_{Qj}^{(r)}-\frac{1}{m}\sum_{j=1}^m\mathbb{E}_{{Q}}\left[\tilde{\mathbf{g}}_{Qj}^{(r)}\right]\|^2\right]+\|\mathbb{E}_{{Q}}\left[\frac{1}{m}\sum_{j=1}^m\tilde{\mathbf{g}}_{Qj}^{(r)}\right]\|^2\right]\right]\nonumber\\
&\stackrel{\text{\ding{193}}}{=}\mathbb{E}_{{\xi}^{(r)}}\left[\mathbb{E}_{{Q}}\left[\frac{1}{m^2}\sum_{j=1}^m\left[\left\|\tilde{\mathbf{g}}_{Qj}^{(r)}-\tilde{\mathbf{g}}^{(r)}_j\right\|^2\right]\right]+\left\|\frac{1}{m}\sum_{j=1}^m\tilde{\mathbf{g}}_j^{(r)}\right\|^2\right]\nonumber\\
&\stackrel{\text{\ding{194}}}{\leq}\mathbb{E}_{{\xi}^{(r)}}\left[\sum_{j=1}^m\frac{q}{m^2}\left\|\tilde{\mathbf{g}}_j^{(r)}\right\|^2+\left\|\frac{1}{m}\sum_{j=1}^m\tilde{\mathbf{g}}_j^{(r)}\right\|^2\right]\nonumber\\
&=\left[\sum_{j=1}^m\frac{q}{m^2}\left[\text{Var}\left(\tilde{\mathbf{g}}_j^{(r)}\right)+\left\|{\mathbf{g}}_j^{(r)}\right\|^2\right]+\left[\text{Var}\left(\frac{1}{m}\sum_{j=1}^m\tilde{\mathbf{g}}_j^{(r)}\right)+\left\|\frac{1}{m}\sum_{j=1}^m{\mathbf{g}}_j^{(r)}\right\|^2\right]\right]\nonumber\\
&=\sum_{j=1}^m\frac{q}{m^2}\left[\text{Var}\left(\tilde{\mathbf{g}}_j^{(r)}\right)+\left\|{\mathbf{g}}_j^{(r)}\right\|^2\right]+\left[\frac{1}{m^2}\sum_{j=1}^m\text{Var}\left(\tilde{\mathbf{g}}_j^{(r)}\right)+\left\|\frac{1}{m}\sum_{j=1}^m{\mathbf{g}}_j^{(r)}\right\|^2\right]\nonumber\\
&\leq \sum_{j=1}^m\frac{q}{m^2}\left[\text{Var}\left(\tilde{\mathbf{g}}_j^{(r)}\right)+\left\|{\mathbf{g}}_j^{(r)}\right\|^2\right]+\left[\frac{1}{m^2}\sum_{j=1}^m\text{Var}\left(\tilde{\mathbf{g}}_j^{(r)}\right)+\frac{1}{m}\sum_{j=1}^m\left\|{\mathbf{g}}_j^{(r)}\right\|^2\right]\label{eq:lemma111} 
\end{align}
where \text{\ding{192}} holds due to $\mathbb{E}\left[\left\|\mathbf{x}\right\|^2\right]=\text{Var}[\mathbf{x}]+\left\|\mathbb{E}[\mathbf{x}]\right\|^2$, \text{\ding{193}} is due to $\mathbb{E}_{{Q}}\left[\frac{1}{m}\sum_{j=1}^m\tilde{\mathbf{g}}_{Qj}^{(r)}\right]=\frac{1}{m}\sum_{j=1}^m\tilde{\mathbf{g}}_{j}^{(r)}$ and \text{\ding{194}} follows from Assumption~\ref{Assu:09}.

Next we show that from Assumptions~\ref{Assu:2}, we have 
\begin{align}\label{eq:100000}
    \mathbb{E}_{\xi^{(r)}}\left[\Big[\|{\tilde{\mathbf{g}}_j^{(r)}}-{\mathbf{g}_j^{(r)}}\|^2\Big]\right]\leq \tau \sigma^2
\end{align}
To do so, note that 
\begin{align}
    \mathbb{E}_{\xi^{(r)}}\left[\left\|{\tilde{\mathbf{g}}_j^{(r)}}-{\mathbf{g}_j^{(r)}}\right\|^2\right]&\stackrel{\text{\ding{192}}}{=}\mathbb{E}_{\xi^{(r)}}\left[\left\|\sum_{c=0}^{\tau-1}\left[\tilde{\mathbf{g}}_j^{(c,r)}-\mathbf{g}_j^{(c,r)}\right]\right\|^2\right]\nonumber\\
    &{=}\text{Var}\left(\sum_{c=0}^{\tau-1}\tilde{\mathbf{g}}_j^{(c,r)}\right)\nonumber\\
    &\stackrel{\text{\ding{193}}}{=}\sum_{c=0}^{\tau-1}\text{Var}\left(\tilde{\mathbf{g}}_j^{(c,r)}\right)\nonumber\\
    &{=}\sum_{c=0}^{\tau-1}\mathbb{E}\left[\left\|\tilde{\mathbf{g}}_j^{(c,r)}-\mathbf{g}_j^{(c,r)}\right\|^2\right]\nonumber\\
    &\stackrel{\text{\ding{194}}}{\leq}\tau\sigma^2\label{eq:var_b_mid}
    \end{align}
where in \text{\ding{192}} we use the definition of ${\tilde{\mathbf{g}}}_j^{(r)}$ and ${{\mathbf{g}}}_j^{(r)}$, in \text{\ding{193}} we use the fact that mini-batches are chosen in i.i.d. manner at each local machine, and \text{\ding{194}} immediately follows from Assumptions~\ref{Assu:1.5}.

Replacing $\mathbb{E}_{\xi^{(r)}}\left[\|{\tilde{\mathbf{g}}_j^{(r)}}-{\mathbf{g}_j^{(r)}}\|^2\right]$ in \eqref{eq:lemma111} by its upper bound in \eqref{eq:100000} implies that 
\begin{align}
\mathbb{E}_{{\xi^{(r)}|\boldsymbol{w}^{(r)}}}\mathbb{E}_{{Q}}\Big[\|\frac{1}{m}\sum_{j=1}^m Q\left(\sum_{c=0}^{\tau-1}\tilde{\mathbf{g}}^{(c,r)}_j\right)\|^2\Big]
\leq\sum_{j=1}^m\frac{q}{m^2}\left[\tau\sigma^2+\left\|{\mathbf{g}}_j^{(r)}\right\|^2\right]+\left[\frac{1}{m^2}\sum_{j=1}^m\tau\sigma^2+\frac{1}{m}\sum_{j=1}^m\left\|{\mathbf{g}}_j^{(r)}\right\|^2\right]\label{eq:lemma112}
\end{align}

Further note that we have 
\begin{align}
\left\|{\mathbf{g}}_j^{(r)}\right\|^2&=\|\sum_{c=0}^{\tau-1}\mathbf{g}_j^{(c,r)}\|^2\stackrel{}{\leq} \tau\sum_{c=0}^{\tau-1}\|\mathbf{g}_j^{(c,r)}\|^2\label{eq:mid-bounding-absg}
\end{align} 
where the last inequality is due to $\left\|\sum_{j=1}^n\mathbf{a}_i\right\|^2\leq n\sum_{j=1}^n\left\|\mathbf{a}_i\right\|^2$, which together with \eqref{eq:lemma112} leads to the following bound:
\begin{align}
    \mathbb{E}_{{\xi^{(r)}|\boldsymbol{w}^{(r)}}}\mathbb{E}_{{Q}}\Big[\|\frac{1}{m}\sum_{j=1}^m Q\left(\sum_{c=0}^{\tau-1}\tilde{\mathbf{g}}^{(c,r)}_j\right)\|^2\Big]\leq\tau(\frac{q}{m}+1)\frac{1}{m}\sum_{j=1}^m\sum_{c=0}^{\tau-1}\|\mathbf{g}_j^{(c,r)}\|^2+\left(q+1\right)\frac{\tau\sigma^2}{m},
\end{align}
and the proof is complete.
\end{proof}

\begin{lemma}\label{lemma:cross-inner-bound-unbiased}
  Under Assumption~\ref{Assu:1}, and according to the \texttt{FedCOM} algorithm the expected inner product between stochastic gradient and full batch gradient can be bounded with:
\begin{align}
    - \mathbb{E}\left[\left\langle\nabla f({\boldsymbol{w}}^{(r)}),{{\tilde{\mathbf{g}}}^{(r)}}\right\rangle\right]&\leq \frac{1}{2}\eta\frac{1}{m}\sum_{j=1}^m\sum_{c=0}^{\tau-1}\left[-\|\nabla f({\boldsymbol{w}}^{(r)})\|_2^2-\|\nabla{f}(\boldsymbol{w}_j^{(c,r)})\|_2^2+L^2\|{\boldsymbol{w}}^{(r)}-\boldsymbol{w}_j^{(c,r)}\|_2^2\right]\label{eq:lemma3-thm2}
\end{align}

\end{lemma}
\begin{proof}
We have:
\begin{align}
    &-\mathbb{E}_{\{{\xi}^{(t)}_{1}, \ldots, {\xi}^{(t)}_{m}|{\boldsymbol{w}}^{(t)}_{1},\ldots,  {\boldsymbol{w}}^{(t)}_{m}\}} \mathbb{E}_Q\left[ \big\langle\nabla f({\boldsymbol{w}}^{(r)}),\tilde{\mathbf{g}}_Q^{(r)}\big\rangle\right]\nonumber\\
    &=-\mathbb{E}_{\{{\xi}^{(t)}_{1}, \ldots, {\xi}^{(t)}_{m}|{\boldsymbol{w}}^{(t)}_{1},\ldots,  {\boldsymbol{w}}^{(t)}_{m}\}}\left[\left\langle \nabla f({\boldsymbol{w}}^{(r)}),\eta\frac{1}{m}\sum_{j=1}^m\sum_{c=0}^{\tau-1}\tilde{\mathbf{g}}_j^{(c,r)}\right\rangle\right]\nonumber\\
    &=-\left\langle \nabla f({\boldsymbol{w}}^{(r)}),\eta\frac{1}{m}\sum_{j=1}^m\sum_{c=0}^{\tau-1}\mathbb{E}\left[\tilde{\mathbf{g}}_j^{(c,r)}\right]\right\rangle\nonumber\\
        &=-\eta\sum_{c=0}^{\tau-1}\frac{1}{m}\sum_{j=1}^m\left\langle \nabla f({\boldsymbol{w}}^{(r)}),{\mathbf{g}}_j^{(c,r)}\right\rangle\nonumber\\ 
     &\stackrel{\text{\ding{192}}}{=}\frac{1}{2}\eta\sum_{c=0}^{\tau-1}\frac{1}{m}\sum_{j=1}^m\left[-\|\nabla f({\boldsymbol{w}}^{(r)})\|_2^2-\|{{\nabla{f}}}(\boldsymbol{w}_j^{(c,r)})\|_2^2+\|\nabla f({\boldsymbol{w}}^{(r)})-\nabla{f}(\boldsymbol{w}_j^{(c,r)})\|_2^2\right]\nonumber\\
    &\stackrel{\text{\ding{193}}}{\leq}\frac{1}{2}\eta\sum_{c=0}^{\tau-1}\frac{1}{m}\sum_{j=1}^m\left[-\|\nabla f({\boldsymbol{w}}^{(r)})\|_2^2-\|\nabla{f}(\boldsymbol{w}_j^{(c,r)})\|_2^2+L^2\|{\boldsymbol{w}}^{(r)}-\boldsymbol{w}_j^{(c,r)}\|_2^2\right]
   \label{eq:bounding-cross-no-redundancy}
\end{align}

where \ding{192} is due to $2\langle \mathbf{a},\mathbf{b}\rangle=\|\mathbf{a}\|^2+\|\mathbf{b}\|^2-\|\mathbf{a}-\mathbf{b}\|^2$, and \ding{193} follows from Assumption \ref{Assu:1}.
\end{proof}

The following lemma bounds the distance of local solutions from global solution at $r$th communication round.
\begin{lemma}\label{lemma:dif-under-pl-sgd-iid}
Under Assumptions~\ref{Assu:1.5} we have:
\begin{align}
      \mathbb{E}\left[\|{\boldsymbol{w}}^{(r)}-\boldsymbol{w}_j^{(c,r)}\|_2^2\right]&\leq\eta^2\tau\sum_{c=0}^{\tau-1}\left\|{\mathbf{g}}_j^{(c,r)}\right\|_2^2+\eta^2\tau\sigma^2
\end{align}

\end{lemma}

\begin{proof}
Note that
\begin{align}
 \mathbb{E}\left[\left\|{\boldsymbol{w}}^{(r)}-\boldsymbol{w}_j^{(c,r)}\right\|_2^2\right]&=\mathbb{E}\left[\left\|{\boldsymbol{w}}^{(r)}-\left({\boldsymbol{w}}^{(r)}-\eta\sum_{k=0}^{c}\tilde{\mathbf{g}}_j^{(k,r)}\right)\right\|_2^2\right]\nonumber\\
 &=\mathbb{E}\left[\left\|\eta\sum_{k=0}^{c}\tilde{\mathbf{g}}_j^{(k,r)}\right\|_2^2\right]\nonumber\\
 &\stackrel{\text{\ding{192}}}{=}\mathbb{E}\left[\left\|\eta\sum_{k=0}^{c}\left(\tilde{\mathbf{g}}_j^{(k,r)}-{\mathbf{g}}_j^{(k,r)}\right)\right\|_2^2\right]+\left[\left\|\eta\sum_{k=0}^{c}{\mathbf{g}}_j^{(k,r)}\right\|_2^2\right]\nonumber\\
 &\stackrel{\text{\ding{193}}}{=}\eta^2\sum_{k=0}^{c}\mathbb{E}\left[\left\|\left(\tilde{\mathbf{g}}_j^{(k,r)}-{\mathbf{g}}_j^{(k,r)}\right)\right\|_2^2\right]+\left(c+1\right)\eta^2\sum_{k=0}^{c}\left[\left\|{\mathbf{g}}_j^{(k,r)}\right\|_2^2\right]\nonumber\\
  &{\leq}\eta^2\sum_{k=0}^{\tau-1}\mathbb{E}\left[\left\|\left(\tilde{\mathbf{g}}_j^{(k,r)}-{\mathbf{g}}_j^{(k,r)}\right)\right\|_2^2\right]+\tau\eta^2\sum_{k=0}^{\tau-1}\left[\left\|{\mathbf{g}}_j^{(k,r)}\right\|_2^2\right]\nonumber\\
  &\stackrel{\text{\ding{194}}}{\leq}\eta^2\sum_{k=0}^{\tau-1}\sigma^2+\tau\eta^2\sum_{k=0}^{\tau-1}\left[\left\|{\mathbf{g}}_j^{(k,r)}\right\|_2^2\right]\nonumber\\
 &{=}\eta^2\tau\sigma^2+\eta^2\sum_{k=0}^{\tau-1}\tau\left\|{\mathbf{g}}_j^{(k,r)}\right\|_2^2
\end{align}

where \ding{192} comes from $\mathbb{E}\left[\mathbf{x}^2\right]=\text{Var}\left[\mathbf{x}\right]+\left[\mathbb{E}\left[\mathbf{x}\right]\right]^2$ and \ding{193} holds because $\text{Var}\left(\sum_{j=1}^n\mathbf{x}_j\right)=\sum_{j=1}^n\text{Var}\left(\mathbf{x}_j\right)$ for i.i.d. vectors $\mathbf{x}_i$ (and i.i.d. assumption comes from i.i.d. sampling), and finally \ding{194} follows from Assumption~\ref{Assu:1.5}.
\end{proof}

\subsection{Main result for the non-convex setting}
Now we are ready to present our result for the homogeneous setting. We first state and prove the result for the general nonconvex objectives.  
\begin{theorem}[Non-convex]\label{thm:lsgwd-lr} For \texttt{FedCOM}$(\tau, \eta, \gamma)$, for all $0\leq t\leq R\tau-1$,  under Assumptions \ref{Assu:1} to \ref{Assu:1.5}, if the learning rate satisfies \begin{align}
   1\geq {\tau^2 L^2\eta^2}+\left(\frac{q}{m}+1\right){\eta\gamma L}{\tau}
\label{eq:cnd-thm4.3}
\end{align}
and all local model parameters are initialized at the same point ${\boldsymbol{w}}^{(0)}$, then the average-squared gradient after $\tau$ iterations is bounded as follows:
\begin{align}
        \frac{1}{R}\sum_{r=0}^{R-1}\left\|\nabla f({\boldsymbol{w}}^{(r)})\right\|_2^2\leq \frac{2\left(f(\boldsymbol{w}^{(0)})-f(\boldsymbol{w}^{(*)})\right)}{\eta\gamma\tau R}+\frac{L\eta\gamma{\left(q+1\right)}}{m}\sigma^2+{L^2\eta^2\tau }\sigma^2\label{eq:thm1-result} 
\end{align}
where $\boldsymbol{w}^{(*)}$ is the global optimal solution with  function value $f(\boldsymbol{w}^{(*)})$.
\end{theorem}

\begin{proof}
Before proceeding to the proof of Theorem~\ref{thm:lsgwd-lr}, we would like to highlight that 
\begin{align}
    \boldsymbol{w}^{(r)}- ~{\boldsymbol{w}}_{j}^{(\tau,r)}=\eta\sum_{c=0}^{\tau-1}\tilde{\mathbf{g}}_j^{(c,r)}.\label{eq:decent-smoothe}
\end{align}

From the updating rule of Algorithm~\ref{Alg:one-shot-using data samoples-b} we have

{
\begin{align}
     {\boldsymbol{w}}^{(r+1)}=\boldsymbol{w}^{(r)}-\gamma\eta\left(\frac{1}{m}\sum_{j=1}^mQ\Big(\sum_{c=0,r}^{\tau-1}\tilde{\mathbf{g}}_{j}^{(c,r)}\Big)\right)=\boldsymbol{w}^{(r)}-\gamma\left[\frac{\eta}{m}\sum_{j=1}^mQ\left(\sum_{c=0}^{\tau-1}\tilde{\mathbf{g}}_{j}^{(c,r)}\right)\right]\label{eq:update-rule-dec}
\end{align}
}
In what follows, we use the following notation to denote the stochastic gradient used to update the global model at $r$th communication round $$\tilde{\mathbf{g}}_Q^{(r)}\triangleq\frac{\eta}{m}\sum_{j=1}^{m}{Q}\left(\frac{\boldsymbol{w}^{(r)}- ~{\boldsymbol{w}}_{j}^{(\tau,r)}}{\eta}\right)=\frac{\eta}{m}\sum_{j=1}^{m}{Q}\left(\sum_{c=0}^{\tau-1}\tilde{\mathbf{g}}_j^{(c,r)}\right).$$ 
and notice that $\boldsymbol{w}^{(r)} = \boldsymbol{w}^{(r-1)} - \gamma \tilde{\mathbf{g}}^{(r)}$.

Then using the Assumption~\ref{Assu:09} we have:
\begin{align}
  \mathbb{E}_Q\left[\tilde{\mathbf{g}}_Q^{(r)}\right]=\frac{1}{m}\sum_{j=1}\left[-\eta\mathbb{E}_Q\left[ Q\left(\sum_{c=0}^{\tau-1}\tilde{\mathbf{g}}_j^{(c,r)}\right)\right]\right]=\frac{1}{m}\sum_{j=1}\left[-\eta\left(\sum_{c=0}^{\tau-1}\tilde{\mathbf{g}}_j^{(c,r)}\right)\right]\triangleq \tilde{\mathbf{g}}^{(r)}\label{eq:unbiased_gd1} 
\end{align}


From the $L$-smoothness gradient assumption on global objective, by using  $\tilde{\mathbf{g}}^{(r)}$ in inequality (\ref{eq:decent-smoothe}) we have:
\begin{align}
    f({\boldsymbol{w}}^{(r+1)})-f({\boldsymbol{w}}^{(r)})\leq -\gamma \big\langle\nabla f({\boldsymbol{w}}^{(r)}),\tilde{\mathbf{g}}^{(r)}\big\rangle+\frac{\gamma^2 L}{2}\|\tilde{\mathbf{g}}^{(r)}\|^2\label{eq:Lipschitz-c1}
\end{align}
By taking expectation on both sides of above inequality over sampling, we get:
\begin{align}
    \mathbb{E}\left[\mathbb{E}_Q\Big[f({\boldsymbol{w}}^{(r+1)})-f({\boldsymbol{w}}^{(r)})\Big]\right]&\leq -\gamma\mathbb{E}\left[\mathbb{E}_Q\left[ \big\langle\nabla f({\boldsymbol{w}}^{(r)}),\tilde{\mathbf{g}}_Q^{(r)}\big\rangle\right]\right]+\frac{\gamma^2 L}{2}\mathbb{E}\left[\mathbb{E}_Q\|\tilde{\mathbf{g}}_Q^{(r)}\|^2\right]\nonumber\\
    &\stackrel{(a)}{=}-\gamma\underbrace{\mathbb{E}\left[\left[ \big\langle\nabla f({\boldsymbol{w}}^{(r)}),\tilde{\mathbf{g}}^{(r)}\big\rangle\right]\right]}_{(\mathrm{I})}+\frac{\gamma^2 L}{2}\underbrace{\mathbb{E}\left[\mathbb{E}_Q\Big[\|\tilde{\mathbf{g}}_Q^{(r)}\|^2\Big]\right]}_{\mathrm{(II)}}\label{eq:Lipschitz-c-gd}
\end{align}
We proceed to use Lemma~\ref{lemma:tasbih1-iid}, Lemma~\ref{lemma:cross-inner-bound-unbiased}, and Lemma~\ref{lemma:dif-under-pl-sgd-iid}, to bound  terms $(\mathrm{I})$ and $(\mathrm{II})$ in right hand side of (\ref{eq:Lipschitz-c-gd}), which gives
\begin{align}
     &\mathbb{E}\left[\mathbb{E}_Q\Big[f({\boldsymbol{w}}^{(r+1)})-f({\boldsymbol{w}}^{(r)})\Big]\right]\nonumber\\
     &\leq \gamma\frac{1}{2}\eta\frac{1}{m}\sum_{j=1}^m\sum_{c=0}^{\tau-1}\left[-\left\|\nabla f({\boldsymbol{w}}^{(r)})\right\|_2^2-\left\|\mathbf{g}_j^{(c,r)}\right\|_2^2+L^2\eta^2\sum_{c=0}^{\tau-1}\left[\tau\left\|{\mathbf{g}}_j^{(c,r)}\right\|_2^2+\sigma^2\right]\right]\nonumber\\
     &\quad+\frac{\gamma^2 L(\frac{q}{m}+1)}{2}\left[\frac{\eta^2\tau}{m}\sum_{j=1}^m\sum_{c=0}^{\tau-1}\|\mathbf{g}^{(c,r)}_{j}\|^2\right]+\frac{\gamma^2\eta^2 L(q+1)}{2}\frac{\tau \sigma^2}{m}\nonumber\\
     &\stackrel{\text{\ding{192}}}{\leq}\frac{\gamma\eta}{2m}\sum_{j=1}^m\sum_{c=0}^{\tau-1}\left[-\left\|\nabla f({\boldsymbol{w}}^{(r)})\right\|_2^2-\left\|\mathbf{g}_j^{(c,r)}\right\|_2^2+\tau L^2\eta^2\left[\tau\left\|{\mathbf{g}}_j^{(c,r)}\right\|_2^2+\sigma^2\right]\right]\nonumber\\
     &\quad+\frac{\gamma^2 L(\frac{q}{m}+1)}{2}\left[\frac{\eta^2\tau}{m}\sum_{j=1}^m\sum_{c=0}^{\tau-1}\|\mathbf{g}^{(c,r)}_{j}\|^2\right]+\frac{\gamma^2\eta^2 L(q+1)}{2}\frac{\tau \sigma^2}{m}\nonumber\\
     &=-\eta\gamma\frac{\tau}{2}\left\|\nabla f({\boldsymbol{w}}^{(r)})\right\|_2^2\nonumber\\
     &\quad-\left(1-{\tau L^2\eta^2\tau}-{(\frac{q}{m}+1)\eta\gamma L}{\tau}\right)\frac{\eta\gamma}{2m}\sum_{j=1}^m\sum_{c=0}^{\tau-1}\|\mathbf{g}^{(c,r)}_{j}\|^2+\frac{L\tau\gamma\eta^2 }{2m}\left(mL\tau\eta+\gamma(q+1)\right)\sigma^2\nonumber\\
     &\stackrel{\text{\ding{193}}}{\leq} -\eta\gamma\frac{\tau}{2}\left\|\nabla f({\boldsymbol{w}}^{(r)})\right\|_2^2+\frac{L\tau\gamma\eta^2 }{2m}\left(mL\tau\eta+\gamma(q+1)\right)\sigma^2\label{eq:finalll}
\end{align}
where in \ding{192} we incorporate outer summation $\sum_{c=0}^{\tau-1}$, and  \ding{193} follows from condition 
\begin{align}
   1\geq {\tau L^2\eta^2\tau}+(\frac{q}{m}+1)\eta\gamma L{\tau}. 
\end{align}
Summing up for all $R$ communication rounds and  rearranging the terms gives:
\begin{align}
    \frac{1}{R}\sum_{r=0}^{R-1}\left\|\nabla f({\boldsymbol{w}}^{(r)})\right\|_2^2\leq \frac{2\left(f(\boldsymbol{w}^{(0)})-f(\boldsymbol{w}^{(*)})\right)}{\eta\gamma\tau R}+\frac{L\eta\gamma{(q+1)}}{m}\sigma^2+{L^2\eta^2\tau }\sigma^2
\end{align}
From above inequality, is it easy to see that in order to achieve a linear speed up, we need to have $\eta\gamma=O\left(\frac{\sqrt{m}}{\sqrt{R \tau}}\right)$.
\end{proof}

\begin{corollary}[Linear speed up] 
In Eq.~(\ref{eq:thm1-result}) for the choice of  $\eta\gamma=O\left(\frac{1}{L}\sqrt{\frac{m}{R\tau\left(q+1\right)}}\right)$, and $\gamma\geq m$  the  convergence rate reduces to:
\begin{align}
    \frac{1}{R}\sum_{r=0}^{R-1}\left\|\nabla f({\boldsymbol{w}}^{(r)})\right\|_2^2&\leq O\left(\frac{L\sqrt{\left(q+1\right)}\left(f(\boldsymbol{w}^{(0)})-f(\boldsymbol{w}^{*})\right)}{\sqrt{mR\tau}}+\frac{\left(\sqrt{\left(q+1\right)}\right)\sigma^2}{\sqrt{mR\tau}}+\frac{m\sigma^2}{R\gamma^2}\right).\label{eq:convg-error}
\end{align}
Note that according to Eq.~(\ref{eq:convg-error}), if we pick  a fixed constant value for  $\gamma$, in order to achieve an $\epsilon$-accurate solution, $R=O\left(\frac{1}{\epsilon}\right)$ communication rounds and $\tau=O\left(\frac{q+1}{m\epsilon}\right)$ local updates are necessary. We also highlight  that Eq.~(\ref{eq:convg-error}) also allows us to choose $R=O\left(\frac{q+1}{\epsilon}\right)$ and $\tau=O\left(\frac{1}{m\epsilon}\right)$ to get the  same convergence rate.
\end{corollary}

\begin{remark}\label{rmk:cnd-lr}

Condition in Eq.~(\ref{eq:cnd-thm4.3}) can be rewritten as 
\begin{align}
    \eta&\leq \frac{-\gamma L\tau\left(\frac{q}{m}+1\right)+\sqrt{\gamma^2 \left(L\tau\left(\frac{q}{m}+1\right)\right)^2+4L^2\tau^2}}{2L^2\tau^2}\nonumber\\
    &= \frac{-\gamma L\tau\left(\frac{q}{m}+1\right)+L\tau\sqrt{\left(\frac{q}{m}+1\right)^2\gamma^2 +4}}{2L^2\tau^2}\nonumber\\
    &=\frac{\sqrt{\left(\frac{q}{m}+1\right)^2\gamma^2 +4}-\left(\frac{q}{m}+1\right)\gamma}{2L\tau}\label{eq:lrcnd}
\end{align}

So based on Eq.~(\ref{eq:lrcnd}), if we set $\eta=O\left(\frac{1}{L\gamma}\sqrt{\frac{m}{R\tau\left(q+1\right)}}\right)$, it implies that:
\begin{align}
    R\geq \frac{\tau m}{\left(q+1\right)\gamma^2\left(\sqrt{\left(\frac{q}{m}+1\right)^2\gamma^2+4}-\left(\frac{q}{m}+1\right)\gamma\right)^2}\label{eq:iidexact}
\end{align}
We note that $\gamma^2\left(\sqrt{\left(q+1\right)^2\gamma^2+4}-\left(q+1\right)\gamma\right)^2=\Theta(1)\leq 5 $ therefore even for $\gamma\geq m$ we need to have 
\begin{align}
    R\geq \frac{\tau m}{5\left(q+1\right)}=O\left(\frac{\tau m}{q+1}\right)\label{eq:lrbnd-homog}
\end{align}

\textbf{Therefore, for the choice of $\tau=O\left(\frac{q+1}{m\epsilon}\right)$, due to condition in Eq.~(\ref{eq:lrbnd-homog}), we need to have $R=O\left(\frac{1}{\epsilon}\right)$. Similarly, we can have $R=O\left(\frac{q+1}{\epsilon}\right)$ and $\tau=O\left(\frac{1}{m\epsilon}\right)$.}

\end{remark}

\begin{corollary}[Special case, $\gamma=1$]
By letting $\gamma=1$, $q=0$ the convergence rate in Eq.~(\ref{eq:thm1-result}) reduces to 
\begin{align}
     \frac{1}{R}\sum_{r=0}^{R-1}\left\|\nabla f({\boldsymbol{w}}^{(r)})\right\|_2^2&\leq \frac{2\left(f(\boldsymbol{w}^{(0)})-f(\boldsymbol{w}^{(*)})\right)}{\eta R\tau}+\frac{L\eta }{m}\sigma^2+{L^2\eta^2\tau }\sigma^2
\end{align}
which matches the rate  obtained in~\cite{wang2018cooperative}. In this case the communication complexity and the number of local updates become \begin{align}
    {R}=O\left(\frac{m}{\epsilon}\right), \:\:\: \tau=O\left(\frac{1}{\epsilon}\right).
\end{align}
This simply implies  that in this special case the convergence rate of our algorithm reduces to the  rate obtained in~\cite{wang2018cooperative}, which indicates the tightness of  our analysis.
\end{corollary}

\subsection{Main result for the PL/strongly convex setting}
 
We now turn to stating the convergence rate for the homogeneous setting under PL condition which naturally leads to the same rate for strongly convex functions.
\begin{theorem}[PL or strongly convex]\label{thm:pl-iid}
For \texttt{FedCOM}$(\tau, \eta, \gamma)$, for all $0\leq t\leq R\tau-1$,  under Assumptions \ref{Assu:1} to \ref{Assu:1.5} and \ref{assum:pl},if the learning rate satisfies \begin{align}
   1\geq {\tau^2 L^2\eta^2}+\left(\frac{q}{m}+1\right){\eta\gamma L}{\tau} 
\end{align}
and if the all the models are initialized with $\boldsymbol{w}^{(0)}$ we obtain:
\begin{align}
        \mathbb{E}\Big[f({\boldsymbol{w}}^{(R)})-f({\boldsymbol{w}}^{(*)})\Big]&\leq \left(1-\eta\gamma{\mu\tau}\right)^R\left(f(\boldsymbol{w}^{(0)})-f(\boldsymbol{w}^{(*)})\right)+\frac{1}{{\mu}}\left[\frac{1}{2} L^2\tau\eta^2\sigma^2+\left(1+q\right)\frac{\gamma\eta L\sigma^2}{2m}\right]
\end{align}
\end{theorem}

\begin{proof}
From Eq.~(\ref{eq:finalll}) under condition:
\begin{align}
       1\geq {\tau L^2\eta^2\tau}+{{(\frac{q}{m}+1)}\eta\gamma L}{\tau} 
\end{align}
we obtain:
\begin{align}
         \mathbb{E}\Big[f({\boldsymbol{w}}^{(r+1)})-f({\boldsymbol{w}}^{(r)})\Big]&\leq -\eta\gamma\frac{\tau}{2}\left\|\nabla f({\boldsymbol{w}}^{(r)})\right\|_2^2+\frac{L\tau\gamma\eta^2 }{2m}\left(mL\tau\eta+\gamma(q+1)\right)\sigma^2\nonumber\\
         &\leq -\eta\mu\gamma{\tau} \left(f({\boldsymbol{w}}^{(r)})-f({\boldsymbol{w}}^{(r)})\right)+\frac{L\tau\gamma\eta^2 }{2m}\left(mL\tau\eta+\gamma(q+1)\right)\sigma^2 
\end{align}
which leads to the following bound:
\begin{align}
            \mathbb{E}\Big[f({\boldsymbol{w}}^{(r+1)})-f({\boldsymbol{w}}^{(*)})\Big]&\leq \left(1-\eta\mu\gamma{\tau}\right) \Big[f({\boldsymbol{w}}^{(r)})-f({\boldsymbol{w}}^{(*)})\Big]+\frac{L\tau\gamma\eta^2 }{2m}\left(mL\tau\eta+{(\frac{q}{m}+1)}\gamma\right)\sigma^2
\end{align}
By setting $\Delta=1-\eta\mu\gamma{\tau}$ we obtain  the following bound:
\begin{align}
            &\mathbb{E}\Big[f({\boldsymbol{w}}^{(R)})-f({\boldsymbol{w}}^{(*)})\Big]\nonumber\\
            &\leq \Delta^R \Big[f({\boldsymbol{w}}^{(0)})-f({\boldsymbol{w}}^{(*)})\Big]+\frac{1-\Delta^R}{1-\Delta}\frac{L\tau\gamma\eta^2 }{2m}\left(mL\tau\eta+{(q+1)}\gamma\right)\sigma^2\nonumber\\
            &\leq \Delta^R \Big[f({\boldsymbol{w}}^{(0)})-f({\boldsymbol{w}}^{(*)})\Big]+\frac{1}{1-\Delta}\frac{L\tau\gamma\eta^2 }{2m}\left(mL\tau\eta+{(q+1)}\gamma\right)\sigma^2\nonumber\\
            &={\left(1-\eta\mu\gamma{\tau}\right)}^R \Big[f({\boldsymbol{w}}^{(0)})-f({\boldsymbol{w}}^{(*)})\Big]+\frac{1}{\eta\mu\gamma{\tau}}\frac{L\tau\gamma\eta^2 }{2m}\left(mL\tau\eta+{(q+1)}\gamma\right)\sigma^2
\end{align}
\end{proof}

\begin{corollary}
If we  let $\eta\gamma\mu\tau\leq\frac{1}{2}$, $\eta=\frac{1}{2L\left(\frac{q}{m}+1\right)\tau\gamma }$ and $\kappa=\frac{L}{\mu}$ the convergence error in Theorem~\ref{thm:pl-iid}, with $\gamma\geq m$ results in:
\begin{align}
&\mathbb{E}\Big[f({\boldsymbol{w}}^{(R)})-f({\boldsymbol{w}}^{(*)})\Big]\nonumber\\
&\leq e^{-\eta\gamma{\mu\tau}R}\left(f(\boldsymbol{w}^{(0)})-f(\boldsymbol{w}^{(*)})\right)+\frac{1}{{\mu}}\left[\frac{1}{2} \tau L^2\eta^2\sigma^2+\left(1+q\right)\frac{\gamma\eta L\sigma^2}{2m}\right]\nonumber\\
&\leq e^{-\frac{R}{2\left(\frac{q}{m}+1\right)\kappa}}\left(f(\boldsymbol{w}^{(0)})-f(\boldsymbol{w}^{(*)})\right)+\frac{1}{{\mu}}\left[\frac{1}{2} L^2\frac{\tau\sigma^2}{L^2\left(\frac{q}{m}+1\right)^2\gamma^2\tau^2}+\left(1+q\right)\frac{ L\sigma^2}{2\left(\frac{q}{m}+1\right)L\tau m}\right]\nonumber\\
&=O\left(e^{-\frac{R}{2\left(\frac{q}{m}+1\right)\kappa}}\left(f(\boldsymbol{w}^{(0)})-f(\boldsymbol{w}^{(*)})\right)+\frac{\sigma^2}{\left(\frac{q}{m}+1\right)^2\gamma^2\mu\tau}+\frac{\left(q+1\right)\sigma^2}{\mu\left(\frac{q}{m}+1\right) \tau m}\right)
\nonumber\\
&=O\left(e^{-\frac{R}{2\left(\frac{q}{m}+1\right)\kappa}}\left(f(\boldsymbol{w}^{(0)})-f(\boldsymbol{w}^{(*)})\right)+\frac{\sigma^2}{\gamma^2\mu\tau}+\frac{\left(q+1\right)\sigma^2}{\mu\left(\frac{q}{m}+1\right) \tau m}\right)
\label{eq:pliid}
\end{align}
which indicates  that to achieve an error of $\epsilon$, we need to have $R=O\left(\left(\frac{q}{m}+1\right)\kappa\log\left(\frac{1}{\epsilon}\right)\right)$ and $\tau=\frac{\left(q+1\right)}{\left(\frac{q}{m}+1\right)m\epsilon}$. {Additionally, we note that if $\gamma\rightarrow\infty$, yet $R=O\left(\left(q+1\right)\kappa\log\left(\frac{1}{\epsilon}\right)\right)$ and $\tau=\frac{\left(q+1\right)}{\left(\frac{q}{m}+1\right)m\epsilon}$ will be necessary.}
\end{corollary}

\subsection{Main result for the general convex setting}
\begin{theorem}[Convex]\label{thm:cvx-iid}
 For a general convex function $f(\boldsymbol{w})$ with optimal solution $\boldsymbol{w}^{(*)}$, using  \texttt{FedCOM}$(\tau, \eta, \gamma)$ (Algorithm~\ref{Alg:one-shot-using data samoples-b}) to optimize $\tilde{f}(\boldsymbol{w},\phi)=f(\mathbf{\boldsymbol{w}})+\frac{\phi}{2}\left\|\boldsymbol{w}\right\|^2$,  for all $0\leq t\leq R\tau-1$,  under Assumptions \ref{Assu:1} to \ref{Assu:1.5}, if the learning rate satisfies \begin{align}
   1\geq {\tau^2 L^2\eta^2}+\left(\frac{q}{m}+1\right){\eta\gamma L}{\tau} 
\end{align}
and if the all the models initiate with $\boldsymbol{w}^{(0)}$, with $\phi=\frac{1}{\sqrt{m\tau}}$ and $\eta=\frac{1}{2L\gamma\tau\left(1+\frac{q}{m}\right)}$ we obtain:
\begin{align}
        \mathbb{E}\Big[f({\boldsymbol{w}}^{(R)})-f({\boldsymbol{w}}^{(*)})\Big]&\leq e^{-\frac{ R}{2L\left(1+\frac{q}{m}\right) \sqrt{m\tau}}}\left(f(\boldsymbol{w}^{(0)})-f(\boldsymbol{w}^{(*)})\right)\nonumber\\
        &\qquad +\left[\frac{\sqrt{m}\sigma^2}{8\sqrt{\tau}\gamma^2\left(1+\frac{q}{m}\right)^2} +\frac{\left(1+q\right)\sigma^2}{4\left(1+\frac{q}{m}\right)\sqrt{m\tau}} \right] +\frac{1}{2\sqrt{m\tau}}\left\|\boldsymbol{w}^{(*)}\right\|^2\label{eq:cvx-iid}
\end{align}{{}}
\end{theorem}
We note that above theorem implies that to achieve a convergence error of $\epsilon$ we need to have $R=O\left(L\left(1+q\right)\frac{1}{\epsilon}\log\left(\frac{1}{\epsilon}\right)\right)$ and $\tau=O\left(\frac{(q+1)^2}{m\left(\frac{q}{m}+1\right)^2\epsilon^2}\right)$.

\begin{proof}
Since $\tilde{f}(\boldsymbol{w}^{(r)},\phi)=f(\boldsymbol{w}^{(r)})+\frac{\phi}{2}\left\|\boldsymbol{w}^{(r)}\right\|^2$ is $\phi$-PL, according to Theorem~\ref{thm:pl-iid}, we have:
\begin{align}
   & \tilde{f}(\boldsymbol{w}^{(R)},\phi)-\tilde{f}(\boldsymbol{w}^{(*)},\phi)\nonumber\\
   &={f}(\boldsymbol{w}^{(r)})+\frac{\phi}{2}\left\|\boldsymbol{w}^{(r)}\right\|^2-\left({f}(\boldsymbol{w}^{(*)})+\frac{\phi}{2}\left\|\boldsymbol{w}^{(*)}\right\|^2\right)\nonumber\\
    &\leq \left(1-\eta\gamma{\phi\tau}\right)^R\left(f(\boldsymbol{w}^{(0)})-f(\boldsymbol{w}^{(*)})\right)+\frac{1}{{\phi}}\left[\frac{1}{2} L^2\tau\eta^2\sigma^2+\left(1+q\right)\frac{\gamma\eta L\sigma^2}{2m}\right]\label{eq:mid-cvx}
\end{align}
Next rearranging Eq.~(\ref{eq:mid-cvx}) and replacing $\mu$ with $\phi$ leads to the following error bound:
\begin{align}
  &  {f}(\boldsymbol{w}^{(R)})-f^*\nonumber\\
  &\leq \left(1-\eta\gamma{\phi\tau}\right)^R\left(f(\boldsymbol{w}^{(0)})-f(\boldsymbol{w}^{(*)})\right)+\frac{1}{{\phi}}\left[\frac{1}{2} L^2\tau\eta^2\sigma^2+\left(1+q\right)\frac{\gamma\eta L\sigma^2}{2m}\right] \nonumber\\
  &\qquad +\frac{\phi}{2}\left(\left\|\boldsymbol{w}^*\right\|^2-\left\|\boldsymbol{w}^{(r)}\right\|^2\right)\nonumber\\
    &\leq e^{-\left(\eta\gamma{\phi\tau}\right)R}\left(f(\boldsymbol{w}^{(0)})-f(\boldsymbol{w}^{(*)})\right)+\frac{1}{{\phi}}\left[\frac{1}{2} L^2\tau\eta^2\sigma^2+\left(1+q\right)\frac{\gamma\eta L\sigma^2}{2m}\right] +\frac{\phi}{2}\left\|\boldsymbol{w}^{(*)}\right\|^2 
\end{align}
Next, if we set $\phi=\frac{1}{\sqrt{m\tau}}$ and $\eta=\frac{1}{2\left(1+\frac{q}{m}\right)L\gamma \tau}$, we obtain that
\begin{align}
        &{f}(\boldsymbol{w}^{(R)})-f^*\nonumber\\
        &\leq e^{-\frac{R}{2\left(1+\frac{q}{m}\right)L \sqrt{m\tau}}}\left(f(\boldsymbol{w}^{(0)})-f(\boldsymbol{w}^{(*)})\right)+\sqrt{m\tau}\left[\frac{\sigma^2}{8\tau\gamma^2\left(1+\frac{q}{m}\right)^2} +\frac{\left(1+q\right)\sigma^2}{4\left(1+\frac{q}{m}\right)\tau m}\right] +\frac{1}{2\sqrt{m\tau}}\left\|\boldsymbol{w}^{(*)}\right\|^2 ,
\end{align}
thus the proof is complete. 
\end{proof}

\newpage
\section{Results for the Heterogeneous Setting}\label{app_het_sec}
In this section, we study the convergence properties of  \texttt{FedCOMGATE} method presented in Algorithm~\ref{Alg:VRFLDL}. For this algorithm recall that the update rule can be written as:
\begin{align}
    \boldsymbol{w}^{(r+1)}=\boldsymbol{w}^{(r)}-\eta\gamma\frac{1}{m}\sum_{j=1}^m{Q\left(\sum_{c=0}^{\tau-1}\tilde{\boldsymbol{d}}^{(c,r)}_j\right)}=\boldsymbol{w}^{(r)}-\gamma\frac{1}{m}\sum_{j=1}^m{\eta Q\left(\sum_{c=0}^{\tau-1}\left(\tilde{\mathbf{g}}^{(c,r)}_j-\boldsymbol{\Delta}_j^{(r)}\right)\right)}
\end{align}
Before stating the proofs for \texttt{FedCOMGATE} in the heterogeneous setting, we first mention the following intermediate lemmas.

\begin{lemma}\label{lemm:tasbih000}
Under Assumptions~\ref{Assu:09}, \ref{Assu:2} and \ref{assum:009}, for the updates of \texttt{FedCOMGATE} we have the following bound:
\begin{align}\label{claim_opo}
    &\mathbb{E}\left[\mathbb{E}_Q\left[\left\|\frac{\eta}{m}\sum_{j=1}^m{Q\left(\sum_{c=0}^{\tau-1}\tilde{\mathbf{g}}^{(c,r)}_j-\boldsymbol{\Delta}_j^{(r)}\right)}\right\|^2\right]\right]\nonumber\\
    &\qquad \leq (q+1)\eta^2\tau\frac{\sigma^2}{m}+(q+1)\eta^2\tau\sum_{c=0}^{\tau-1}\left\|\frac{1}{m}\sum_{j=1}^m{\mathbf{g}}_j^{(c,r)}\right\|^2+\eta^2G_q 
\end{align}
\end{lemma}


\begin{proof}
First, note that the expression on the left hand side of \eqref{claim_opo} can be upper bounded by
\begin{align}
\mathbb{E}_{{\xi}}&\mathbb{E}_{{Q}}\Big[\|\frac{1}{m}\sum_{j=1}^m\eta Q\left(\sum_{c=0}^{\tau-1}\tilde{\mathbf{g}}^{(c,r)}_j-\boldsymbol{\Delta}_j^{(r)}\right)\|^2\Big]\nonumber\\
&\stackrel{\text{\ding{192}}}{=} \eta^2\mathbb{E}_{{\xi}}\mathbb{E}_{{Q}}\Big[\|\underbrace{Q\left(\frac{1}{m}\sum_{j=1}^m\sum_{c=0}^{\tau-1}\left(\tilde{\mathbf{g}}^{(c,r)}_j-\boldsymbol{\Delta}_j^{(r)}\right)\right)}_{\tilde{\mathbf{g}}_Q^{(r)}}\|^2+G_q\Big]\nonumber\\
&=\eta^2\mathbb{E}_{{\xi}}\mathbb{E}_{{Q}}\Big[\|\underbrace{Q\left(\overbrace{\frac{1}{m}\sum_{j=1}^m\sum_{c=0}^{\tau-1}\left(\tilde{\mathbf{g}}^{(c,r)}_j\right)}^{\tilde{\mathbf{g}}^{(r)}}\right)}_{\tilde{\mathbf{g}}_Q^{(r)}}\|^2+G_q\Big]\nonumber\\
&{=}\eta^2\mathbb{E}_{{\xi}}\left[\mathbb{E}_{{Q}}\left[\|\tilde{\mathbf{g}}_Q^{(r)}-\mathbb{E}_{{Q}}\left[\tilde{\mathbf{g}}_Q^{(r)}\right]\|^2\right]+\|\mathbb{E}_{{Q}}\left[\tilde{\mathbf{g}}_Q^{(r)}\right]\|^2\right]+\eta^2G_q\nonumber\\
&{=}\eta^2\mathbb{E}_{{\xi}}\left[\mathbb{E}_{{Q}}\left[\left\|\tilde{\mathbf{g}}_Q^{(r)}-\tilde{\mathbf{g}}^{(r)}\right\|^2\right]+\left\|\tilde{\mathbf{g}}^{(r)}\right\|^2\right]+\eta^2G_q\nonumber\\
&{\leq}\eta^2\mathbb{E}_{{\xi}}\left[q\left\|\tilde{\mathbf{g}}^{(r)}\right\|^2+\left\|\tilde{\mathbf{g}}^{(r)}\right\|^2\right]+\eta^2G_q\nonumber\\
&=(q+1)\eta^2\mathbb{E}_{{\xi}}\left[\left\|\tilde{\mathbf{g}}^{(r)}\right\|^2\right]+\eta^2G_q\nonumber\\
&=(q+1)\eta^2\mathbb{E}_{{\xi}}\left[\left\|\tilde{\mathbf{g}}^{(r)}-\mathbb{E}_{{\xi}}\left[\tilde{\mathbf{g}}^{(r)}\right]\right\|^2\right]+(q+1)\eta^2\left\|\mathbb{E}_{\xi}\left[\tilde{\mathbf{g}}^{(r)}\right]\right\|^2+\eta^2G_q\label{eq:lemma19090}
\end{align}
where \text{\ding{192}} comes from Assumption~\ref{Assu:09}.

Moreover, under Assumption \ref{Assu:2}, we can show following variance bound from the averaged stochastic gradient:
\begin{align}\label{eq:eeeee}
    \mathbb{E}\left[\Big[\|{{\tilde{\mathbf{g}}}^{(r)}}-{{{\mathbf{g}}}^{(r)}}\|^2\Big]\right]\leq \frac{\tau\eta^2 \sigma^2}{m}
\end{align}
To prove this claim, note that
\begin{align}
    \mathbb{E}\left[\left\|{{\tilde{\mathbf{g}}}^{(t)}}-{{{\mathbf{g}}}^{(t)}}\right\|^2\right]&\stackrel{\text{\ding{192}}}{=}\mathbb{E}\left[\left\|\frac{1}{m}\sum_{j=1}^m\left[\sum_{c=0}^{\tau-1}\tilde{\mathbf{g}}_j^{(c,r)}-\sum_{c=0}^{\tau-1}\mathbf{g}_j^{(c,r)}\right]\right\|^2\right]\nonumber\\
    &\stackrel{\text{\ding{193}}}{=} \frac{1}{m^2}\sum_{j=1}^m\mathbb{E}\left[\left\|\sum_{c=0}^{\tau-1}\Big[\tilde{\mathbf{g}}_j^{(c,r)}-\mathbf{g}_j^{(c,r)}\Big]\right\|^2\right]\nonumber\\
    &=\frac{1}{m^2}\sum_{j=1}^m\sum_{c=0}^{\tau-1}\mathbb{E}\left[\left\|\tilde{\mathbf{g}}_j^{(c,r)}-\mathbf{g}_j^{(c,r)}\right\|^2\right]\nonumber\\
    &\stackrel{\text{\ding{194}}}{\leq}\frac{1}{m^2}\sum_{j=1}^m\sum_{c=0}^{\tau-1}\sigma^2\label{eq:after-eq}\\
    &=\frac{\tau \sigma^2}{m}\label{eq:var_b_mid1}
    \end{align}
where in \text{\ding{192}} we use the definition of ${\tilde{\mathbf{g}}}^t$ and ${{\mathbf{g}}}^t$, in {\ding{193}} we use the fact that mini-batches are chosen in i.i.d. manner at each local machine, and {\ding{194}} immediately follows from Assumptions~\ref{Assu:2}.

Now replace the upper bound in \eqref{eq:eeeee} into the last expression in \eqref{eq:lemma19090} to obtain
\begin{align}\label{lllsss}
   & \mathbb{E}_{{\xi}}\mathbb{E}_{{Q}}\Big[\|\frac{1}{m}\sum_{j=1}^m\eta Q\left(\sum_{c=0}^{\tau-1}\tilde{\mathbf{g}}^{(c,r)}_j-\boldsymbol{\Delta}_j^{(r)}\right)\|^2\Big]\nonumber\\
   &\leq (q+1)\eta^2\tau\frac{\sigma^2}{m}+(q+1)\eta^2\left\|\mathbb{E}_{\xi}\left[\tilde{\mathbf{g}}^{(r)}\right]\right\|^2+\eta^2G_q
\end{align}
Next, note that i.i.d. data distribution implies $\mathbb{E}[{\tilde{\mathbf{g}}}_j^{(r)}]={{{\mathbf{g}}}}_j^{(r)}$, from which we have
\begin{align}
\left\|\mathbb{E}\left[{\tilde{\mathbf{g}}}^{(r)}\right]\right\|^2
&=\|{{{\mathbf{g}}}}^{(r)}\|^2\nonumber\\
&\leq\left\|\frac{1}{m}\sum_{j=1}\left[\sum_{c=0}^{\tau-1}{g}_j^{(c,r)}\right]\right\|^2\nonumber\\
&\stackrel{\text{\ding{192}}}{\leq}\left\|\frac{1}{m}\sum_{j=1}^m\sum_{c=0}^{\tau-1}\mathbf{g}_j^{(c,r)}\right\|^2\nonumber\\
&\stackrel{\text{\ding{193}}}{\leq}{\tau}\sum_{c=0}^{\tau-1}\left\|\frac{1}{m}\sum_{j=1}^m{g}_j^{(c,r)}\right\|^2\nonumber\\
&=\tau\sum_{c=0}^{\tau-1}\left\|\frac{1}{m}\sum_{j=1}^m\mathbf{g}^{(c,r)}_{j}\right\|^2\label{eq:mid-bounding-absg1}
\end{align}
where \ding{192} follows from convexity of $\|.\|$ and \ding{193} is due to $\left\|\sum_{j=1}^n\mathbf{a}_i\right\|^2\leq n\sum_{j=1}^n\left\|\mathbf{a}_i\right\|^2$. 

Applying this upper bound into \eqref{lllsss} implies that 
\begin{align}
   & \mathbb{E}_{{\xi}}\mathbb{E}_{{Q}}\Big[\|\frac{1}{m}\sum_{j=1}^m\eta Q\left(\sum_{c=0}^{\tau-1}\tilde{\mathbf{g}}^{(c,r)}_j-\boldsymbol{\Delta}_j^{(r)}\right)\|^2\Big]\nonumber\\
   &\leq (q+1)\eta^2\tau\frac{\sigma^2}{m}+(q+1)\eta^2\tau\sum_{c=0}^{\tau-1}\left\|\frac{1}{m}\sum_{j=1}^m{\mathbf{g}}^{(r)}\right\|^2+\eta^2G_q ,
\end{align}
and the proof is complete.
\end{proof}

\begin{lemma}\label{lemma:cross-inner-bound-unbiased_noniid}
  Under Assumptions \ref{Assu:1}, for the updates of \texttt{FedCOMGATE} we can show that the expected inner product between stochastic gradient and full batch gradient can be bounded as
\begin{align}
   & -\eta \mathbb{E}\left[\left\langle\nabla f(\boldsymbol{w}^{(t)}),{{\tilde{\mathbf{g}}}^{(t)}}\right\rangle\right]\nonumber\\
   &\quad\leq \frac{1}{2}\eta\sum_{c=0}^{\tau-1}\left[-\|\nabla f({\boldsymbol{w}}^{(r)})\|_2^2-\|\sum_{j=1}^m\frac{1}{m}\nabla{f}_j(\boldsymbol{w}_j^{(c,r)})\|_2^2+L^2\sum_{j=1}^m\frac{1}{m}\|{\boldsymbol{w}}^{(r)}-\boldsymbol{w}_j^{(c,r)}\|_2^2\right]\label{eq:lemma3-thm2-1}
\end{align}
\end{lemma}

\begin{proof}
This proof is relatively as we state in the following expressions:
\begin{align}
   & -\mathbb{E}_{\{{\xi}^{(t)}_{1}, \ldots, {\xi}^{(t)}_{m}|{\boldsymbol{w}}^{(t)}_{1},\ldots,  {\boldsymbol{w}}^{(t)}_{m}\}} \mathbb{E}_Q\left[ \big\langle\nabla f({\boldsymbol{w}}^{(r)}),\tilde{\mathbf{g}}^{(r)}\big\rangle\right]\nonumber\\
    &=-\mathbb{E}_{\{{\xi}^{(t)}_{1}, \ldots, {\xi}^{(t)}_{m}|{\boldsymbol{w}}^{(t)}_{1},\ldots,  {\boldsymbol{w}}^{(t)}_{m}\}}\left[\left\langle \nabla f({\boldsymbol{w}}^{(r)}),\eta\frac{1}{m}\sum_{j=1}^m\sum_{c=0}^{\tau-1}\tilde{\mathbf{g}}_j^{(c,r)}\right\rangle\right]\nonumber\\
    &=-\left\langle \nabla f({\boldsymbol{w}}^{(r)}),\eta\frac{1}{m}\sum_{j=1}^m\sum_{c=0}^{\tau-1}\mathbb{E}\left[\tilde{\mathbf{g}}_j^{(c,r)}\right]\right\rangle\nonumber\\
        &=-\eta\sum_{c=0}^{\tau-1}\left\langle \nabla f({\boldsymbol{w}}^{(r)}),\frac{1}{m}\sum_{j=1}^m{\mathbf{g}}_j^{(c,r)}\right\rangle\nonumber\\ 
     &\stackrel{\text{\ding{192}}}{=}\frac{1}{2}\eta\sum_{c=0}^{\tau-1}\left[-\|\nabla f({\boldsymbol{w}}^{(r)})\|_2^2-\|\frac{1}{m}\sum_{j=1}^m\nabla{f}_j(\boldsymbol{w}_j^{(c,r)})\|_2^2+\|\nabla f({\boldsymbol{w}}^{(r)})-\frac{1}{m}\sum_{j=1}^m\nabla{f}_j(\boldsymbol{w}_j^{(c,r)})\|_2^2\right]\nonumber\\
    &\stackrel{\text{\ding{193}}}{\leq}\frac{1}{2}\eta\sum_{c=0}^{\tau-1}\frac{1}{m}\left[-\|\nabla f({\boldsymbol{w}}^{(r)})\|_2^2-\|\frac{1}{m}\sum_{j=1}^m\nabla{f}(\boldsymbol{w}_j^{(c,r)})\|_2^2+L^2\frac{1}{m}\sum_{j=1}^m\|{\boldsymbol{w}}^{(r)}-\boldsymbol{w}_j^{(c,r)}\|_2^2\right],
   \label{eq:bounding-cross-no-redundancy1}
\end{align}
where \ding{192} is due to $2\langle \mathbf{a},\mathbf{b}\rangle=\|\mathbf{a}\|^2+\|\mathbf{b}\|^2-\|\mathbf{a}-\mathbf{b}\|^2$, and \ding{193} follows from Assumption \ref{Assu:1}.

\end{proof}

\begin{lemma}\label{lemma:var-red-main}
Under Assumptions~\ref{Assu:09}, \ref{Assu:2} and ~\ref{assum:009}, with $30\eta^2L^2\tau^2\leq 1$ we have:
\begin{align}
    \frac{1}{R}\sum_{r=0}^{R-1}&\sum_{c=0,r}^{\tau-1}\frac{1}{m}\sum_{j=1}^m\mathbb{E}\left\|\boldsymbol{w}_j^{(c,r)}-\boldsymbol{w}^{(r)}\right\|^2\nonumber\\
    &\leq {36\eta^2\tau^2\sigma^2}+ \frac{8\eta^2}{mR}\sum_{j=1}^m\sum_{c=0}^{\tau-1}\left\|\sum_{c=0,r=0}^{\tau-1}\left({\mathbf{g}}_j^{(c,0)}-{\mathbf{g}}^{(0)}\right)\right\|^2 \nonumber\\
&\qquad+10\eta^2(\eta\gamma)^2(q+1)L^2\left[
\frac{\tau^4}{R}\sum_{r=1}^{R-1}\sum_{c=0,r-1}^{\tau-1}\Big[\Big\|\frac{1}{m}\sum_{j=1}^m{\mathbf{g}}_j^{(c,r-1)}\Big\|^2\Big]+
\tau^4\frac{\sigma^2}{m}+{
}\tau^3G_q\right]\nonumber\\
&\qquad+\frac{20\eta^2
\tau^2}{R}\sum_{r=0}^{R-1}\sum_{c=0}^{\tau-1}\Big\|{\mathbf{g}}^{(r)}\Big\|^2.
\end{align}
\end{lemma}
The proof of this intermediate lemma is deferred to Appendix~\ref{sec:appendix:omitted}.

\newpage
\subsection{Main result for the nonconvex setting}
\begin{theorem}[General Non-convex]\label{thm:ncvx-varred}
For \texttt{FedCOMGATE}$(\tau, \eta, \gamma)$, for all $0\leq t\leq R\tau-1$,  under Assumptions \ref{Assu:1}, \ref{Assu:09}, \ref{Assu:2} and \ref{assum:009} and if the learning rate satisfies \begin{align}
    1-10\eta^2(\eta\gamma)^2(q+1)L^4\tau^4-L\eta\gamma\tau(q+1)\geq 0\quad \& \quad 30\eta^2L^2\tau^2\leq 1\label{eq:lr-cnd-varred}
\end{align}
and all local model parameters are initialized at the same point $\bar{\boldsymbol{w}}^{(0)}=\boldsymbol{w}^{(0)}$, we obtain:
\begin{align}
        \frac{1}{R}\sum_{r=0}^{R-1}\|\nabla f({\boldsymbol{w}}^{(r)})\|_2^2
&\leq \frac{2\left(f(\boldsymbol{w}^{(0)})-f(\boldsymbol{w}^{(*)})\right)}{\tau \eta\gamma R}+\frac{(q+1)\gamma L\eta\sigma^2}{m}+{36\eta^2L^2\tau\sigma^2}+{10\eta^2
L^4\tau^3(\eta\gamma)^2(q+1)\frac{\sigma^2}{m}}\nonumber\\
&\quad+{10\eta^2
L^4\tau^2(\eta\gamma)^2(q+1){G_q}} +\frac{32\eta L^2\tau}{m R}\sum_{j=1}^m\left[f_j(\boldsymbol{w}^{(0)}_j)-f_j(\boldsymbol{w}^{(*)}_j)\right]+\frac{16\eta^3L^2\tau^2}{ R}\sigma^2\nonumber\\
&\quad+\frac{32\eta^2L^3\tau^2}{ R}\left(f(\boldsymbol{w}^{(0)})-f(\boldsymbol{w}^{(*)})\right)+\frac{\gamma\eta L}{\tau}G_q\label{eq:thm1-result-varredd}
\end{align}

\end{theorem}

\begin{proof}
Before proceeding to the proof we need to review some properties of our algorithm: 

\begin{itemize}
    \item[1)] $\delta_j^{(0)}=0 $
    \item[2)] $  \boldsymbol{\Delta}_{j,q}^{(r)} = \mathsf{Q}\left(\left({\boldsymbol{w}^{(r)}-\boldsymbol{w}^{(\tau,r)}_j}\right)/\eta\right)$
        \item[3)] $  \boldsymbol{\Delta}_{q}^{(r)} =\frac{1}{m}\sum_{j=1}^m\boldsymbol{\Delta}_{j,q}^{(r)} $
    \item[4)]$\delta_j^{(r)}=\frac{1}{\tau}\sum_{k=0}^{r}\left(\boldsymbol{\Delta}_{q}^{(k)}-\boldsymbol{\Delta}_{j,q}^{(k)}\right)$
    \item[5)]$\frac{1}{m}\sum_{j=1}^m\delta_j^{(r)}=0$.
    \item[6)]We have: $$\boldsymbol{w}^{(r+1)}=\boldsymbol{w}^{(r)}-\gamma\eta\frac{1}{m}\sum_{j=1}^m{Q\left(-\sum_{c=0}^{\tau-1}\tilde{\boldsymbol{d}}^{(c,r)}_{jQ}\right)}=\boldsymbol{w}^{(r)}-\gamma\eta\frac{1}{m}\sum_{j=1}^m{Q\left(\sum_{c=0}^{\tau-1}\left[\tilde{\mathbf{g}}^{(c,r)}_j-\delta_j^{(r)}\right]\right)}$$ which is equivalent to the update rule of the global model of Algorithm~\ref{Alg:VRFLDL}.
    \item[7)] We have:
    \begin{align}
        \delta_j^{(r)}&=\delta_j^{(r-1)}+\frac{1}{\tau }\left(\boldsymbol{\Delta}_{q}^{(k)}-\boldsymbol{\Delta}_{j,q}^{(k)}\right)\nonumber\\
        &=\delta_j^{(r-1)}+\frac{1}{\tau }\left(\frac{1}{m}\sum_{j=1}^mQ\left(-\sum_{c=0}^{\tau-1}\left(\tilde{\mathbf{g}}_{j}^{(c,r-1)}-\delta_j^{(r-1)}\right)\right)+Q\left(\sum_{c=0}^{\tau-1}\left(\tilde{\mathbf{g}}_{j}^{(c,r-1)}-\delta_j^{(r-1)}\right)\right)\right)
    \end{align}

    Therefore, we have 
    \begin{align}
        \mathbb{E}_{Q}\left[\delta_j^{(r)}\right]&=\frac{1}{\tau }\left(-\frac{1}{m}\sum_{j=1}^m\sum_{c=0}^{\tau-1}\tilde{\mathbf{g}}_{j}^{(c,r-1)}+\sum_{c=0}^{\tau-1}\tilde{\mathbf{g}}_{j}^{(c,r-1)}\right)+\frac{1}{\tau}\left(\frac{1}{m}\sum_{j=1}^m \delta_j^{(r-1)}\right)\nonumber\\
        &=\frac{1}{\tau }\left(-\frac{1}{m}\sum_{j=1}^m\sum_{c=0}^{\tau-1}\tilde{\mathbf{g}}_{j}^{(c,r-1)}+\sum_{c=0}^{\tau-1}\tilde{\mathbf{g}}_{j}^{(c,r-1)}\right)
    \end{align}
    \item[8)] From item (7), for $R\geq 1$ we obtain:
    \begin{align}
    \mathbb{E}_{Q}\left[\tilde{\boldsymbol{d}}_{jq}^{(c,r)}\right]=\mathbb{E}_{Q}\left[\tilde{\mathbf{g}}_{j}^{(c,r)}-\delta_j^{(r)}\right]=\tilde{\mathbf{g}}_{j}^{(c,r)}+\frac{1}{\tau }\left(\frac{1}{m}\sum_{j=1}^m\sum_{c=0}^{\tau-1}\tilde{\mathbf{g}}_{j}^{(c,r-1)}-\sum_{c=0}^{\tau-1}\tilde{\mathbf{g}}_{j}^{(c,r-1)}\right)=\tilde{\boldsymbol{d}}_{j}^{(c,r)}\label{eq:ref_for_updrl}
    \end{align}
\end{itemize}

We would like to also highlight that 
\begin{align}
    -\eta Q\left(\frac{\boldsymbol{w}^{(r)}- ~{\boldsymbol{w}}_{j}^{(\tau,r)}}{\eta}\right)=-\eta Q\left(\sum_{c=0}^{\tau-1}\tilde{\boldsymbol{d}}_{jQ}^{(c,r)}\right)\label{eq:decent-smoothe1}
\end{align}

Towards this end,  recalling the notation $$\tilde{\mathbf{g}}_Q^{(r)}\triangleq\frac{1}{m}\sum_{j=1}\left[-\eta Q\left(\sum_{c=0}^{\tau-1}\tilde{\boldsymbol{d}}_{jQ}^{(c,r)}\right)\right],$$
and using the Assumption~\ref{Assu:09} we have:
\begin{align}
  \mathbb{E}_Q\left[\tilde{\mathbf{g}}_Q^{(r)}\right]=\frac{1}{m}\sum_{j=1}\left[-\eta\mathbb{E}_Q\left[ Q\left(\sum_{c=0}^{\tau-1}\tilde{\boldsymbol{d}}_{jq}^{(c,r)}\right)\right]\right]=\frac{1}{m}\sum_{j=1}\left[-\eta\left(\sum_{c=0}^{\tau-1}\tilde{\mathbf{g}}_j^{(c,r)}\right)\right]\triangleq \tilde{\mathbf{g}}^{(r)}\label{eq:unbiased_gd} 
\end{align}
Then following the $L$-smoothness gradient assumption on global objective, by using  $\tilde{\mathbf{g}}^{(r)}$ in inequality (\ref{eq:decent-smoothe1}) we have:
\begin{align}
    f({\boldsymbol{w}}^{(r+1)})-f({\boldsymbol{w}}^{(r)})\leq -\gamma \big\langle\nabla f({\boldsymbol{w}}^{(r)}),\tilde{\mathbf{g}}_Q^{(r)}\big\rangle+\frac{\gamma^2 L}{2}\|\tilde{\mathbf{g}}_Q^{(r)}\|^2\label{eq:Lipschitz-c1-1}
\end{align}

By taking expectation on both sides of above inequality over sampling, we get:
\begin{align}
    \mathbb{E}\left[\mathbb{E}_Q\Big[f({\boldsymbol{w}}^{(r+1)})-f({\boldsymbol{w}}^{(r)})\Big]\right]&\leq -\gamma\mathbb{E}\left[\mathbb{E}_Q\left[ \big\langle\nabla f({\boldsymbol{w}}^{(r)}),\tilde{\mathbf{g}}_Q^{(r)}\big\rangle\right]\right]+\frac{\gamma^2 L}{2}\mathbb{E}\left[\mathbb{E}_Q\|\tilde{\mathbf{g}}_Q^{(r)}\|^2\right]\nonumber\\
    &\stackrel{\text{\ding{192}}}{=}-\gamma\underbrace{\mathbb{E}\left[\left[ \big\langle\nabla f({\boldsymbol{w}}^{(r)}),\tilde{\mathbf{g}}^{(r)}\big\rangle\right]\right]}_{(\mathrm{I})}+\frac{\gamma^2 L}{2}\underbrace{\mathbb{E}\left[\mathbb{E}_Q\Big[\|\tilde{\mathbf{g}}_Q^{(r)}\|^2\Big]\right]}_{\mathrm{(II)}}\label{eq:Lipschitz-c-gd1}
\end{align}
where \text{\ding{192}} follows from Eq.~(\ref{eq:unbiased_gd}). 
Next, by plugging back the results in Lemma~\ref{lemm:tasbih000}, Lemma~\ref{lemma:cross-inner-bound-unbiased_noniid}, and Lemma~\ref{lemma:var-red-main} we obtain  
\begin{align}
     &\mathbb{E}\Big[f({\boldsymbol{w}}^{(r+1)})-f({\boldsymbol{w}}^{(r)})\Big]\nonumber\\
     &\leq \frac{1}{2}\gamma\eta\sum_{c=0,r}^{\tau-1}\Big[-\|\nabla f({\boldsymbol{w}}^{(r)})\|_2^2-\|\sum_{j=1}^m\frac{1}{m}\nabla{f}_j(\boldsymbol{w}_j^{(c,r)})\|_2^2+\frac{L^2}{m}\sum_{j=1}^m\left[\mathbb{E}\left\|\boldsymbol{w}_j^{(c,r)}-\boldsymbol{w}^{(r)}\right\|^2\right]\Big]\nonumber\\
     &+\frac{\gamma^2 L}{2}\left[(q+1)\eta^2\tau\frac{\sigma^2}{m}+(q+1)\eta^2\tau\sum_{c=0}^{\tau-1}\left\|\frac{1}{m}\sum_{j=1}^m{\mathbf{g}}_j^{(c,r)}\right\|^2+\eta^2G_q\right]\nonumber\\
\label{eq:final1}
\end{align}
which leads to

\begin{align}
     &\frac{1}{R}\sum_{r=0}^{R-1}\mathbb{E}\Big[f({\boldsymbol{w}}^{(r+1)})-f({\boldsymbol{w}}^{(r)})\Big]\nonumber\\
     &\leq -\frac{\gamma\eta}{2}\frac{\tau}{R}\sum_{r=0}^{R-1}\|\nabla f({\boldsymbol{w}}^{(r)})\|_2^2-\frac{\gamma\eta}{2}\frac{1}{R}\sum_{r=0}^{R-1}\sum_{c=0,r}^{\tau-1}\|\sum_{j=1}^m\frac{1}{m}\nabla{f}_j(\boldsymbol{w}_j^{(c,r)})\|_2^2\nonumber\\
     &\quad+\frac{\gamma\eta}{2}\frac{L^2}{R}\sum_{r=0}^{R-1}\sum_{c=0,r}^{\tau-1}\frac{1}{m}\sum_{j=1}^m\mathbb{E}\left\|\boldsymbol{w}_j^{(c,r)}-\boldsymbol{w}^{(r)}\right\|^2\nonumber\\
     &\quad+\frac{\gamma^2 L\eta^2\tau\sigma^2(q+1)}{2m}+\frac{\gamma^2\eta^2L}{2}G_q+\frac{(q+1)\gamma^2 L\eta^2\tau}{2R}\sum_{r=0}^{R-1}\sum_{c=0,r}^{\tau-1}\|\frac{1}{m}\sum_{j=1}^m{g}_j^{(c,r)}\|^2\nonumber\\
       &\stackrel{\text{\ding{192}}}{\leq} -\frac{\gamma\eta}{2}\frac{\tau}{R}\sum_{r=0}^{R-1}\|\nabla f({\boldsymbol{w}}^{(r)})\|_2^2-\frac{\gamma\eta}{2}\frac{1}{R}\sum_{r=0}^{R-1}\sum_{c=0,r}^{\tau-1}\|\sum_{j=1}^m\frac{1}{m}\nabla{f}_j(\boldsymbol{w}_j^{(c,r)})\|_2^2\nonumber\\
     &\quad+\frac{\gamma\eta}{2}\frac{L^2}{R}{36\eta^2\tau^2\sigma^2}+ \frac{\gamma\eta}{2}\frac{8L^2\eta^2}{mR}\sum_{j=1}^m\sum_{c=0}^{\tau-1}\left\|\sum_{c=0,r=0}^{\tau-1}\left({\mathbf{g}}_j^{(c,0)}-{\mathbf{g}}^{(0)}\right)\right\|^2 \nonumber\\
&\quad+\frac{L^2\gamma\eta}{2}10\eta^2(\eta\gamma)^2(q+1)L^2\left[
\frac{\tau^4}{R}\sum_{r=0}^{R-1}\sum_{c=0,r=0}^{\tau-1}\Big[\Big\|\frac{1}{m}\sum_{j=1}^m{\mathbf{g}}_j^{(c,r)}\Big\|^2\Big]+
\tau^4\frac{\sigma^2}{m}+{
}\tau^3G_q\right]\nonumber\\
&\quad+\frac{L^2\gamma\eta}{2}\frac{20\eta^2
\tau^2}{R}\sum_{r=0}^{R-1}\sum_{c=0}^{\tau-1}\Big\|{\mathbf{g}}^{(r)}\Big\|^2\nonumber\\
     &\quad+\frac{\gamma^2\eta^2L}{2}G_q+\frac{(q+1)\gamma^2 L\eta^2\tau\sigma^2}{2m}+\frac{(q+1)\gamma^2 L\eta^2\tau}{2R}\sum_{r=0}^{R-1}\sum_{c=0,r}^{\tau-1}\|\frac{1}{m}\sum_{j=1}^m{g}_j^{(c,r)}\|^2\nonumber\\
     &{=} -\left(\frac{\gamma\eta}{2}\frac{\tau}{R}-\frac{L^2\gamma\eta}{2}\frac{20\eta^2
\tau^3}{R}\right)\sum_{r=0}^{R-1}\|\mathbf{g}^{(r)}\|_2^2\nonumber\\
&\quad-\frac{\gamma\eta}{2}\left(1-L^210\eta^2(\eta\gamma)^2(q+1)L^2\tau^4-L(q+1)\eta\gamma\tau\right)\frac{1}{R}\sum_{r=0}^{R-1}\sum_{c=0,r}^{\tau-1}\|\sum_{j=1}^m\frac{1}{m}\nabla{f}_j(\boldsymbol{w}_j^{(c,r)})\|_2^2\nonumber\\
     &\quad+\frac{\gamma\eta}{2}\frac{L^2}{R}{36\eta^2\tau^2\sigma^2}+ \frac{\gamma\eta}{2}\frac{8L^2\eta^2}{mR}\sum_{j=1}^m\sum_{c=0}^{\tau-1}\left\|\sum_{c=0,r=0}^{\tau-1}\left({\mathbf{g}}_j^{(c,0)}-{\mathbf{g}}^{(0)}\right)\right\|^2\nonumber\\
&\quad+\frac{L^2\gamma\eta}{2}10\eta^2(\eta\gamma)^2(q+1)L^2\left[\tau^4\frac{\sigma^2}{m}+{
}\tau^3G_q\right]+\frac{(q+1)\gamma^2 L\eta^2\tau\sigma^2}{2m}+\frac{\gamma^2\eta^2L}{2}G_q\nonumber\\
&\stackrel{\text{\ding{193}}}{\leq} -\quad\frac{\gamma\eta}{2}\frac{\tau}{R}\left(1-{L^2}20\eta^2\tau^2\right)\sum_{r=0}^{R-1}\|\mathbf{g}^{(r)}\|_2^2\nonumber\\
     &\quad+\frac{\gamma\eta}{2}\frac{L^2}{R}{36\eta^2\tau^2\sigma^2}+ \frac{\gamma\eta}{2}\frac{8L^2\eta^2}{mR}\sum_{j=1}^m\sum_{c=0}^{\tau-1}\left\|\sum_{c=0,r=0}^{\tau-1}\left({\mathbf{g}}_j^{(c,0)}-{\mathbf{g}}^{(0)}\right)\right\|^2 \nonumber\\
&\quad+\frac{L^2\gamma\eta}{2}10\eta^2(\eta\gamma)^2(q+1)L^2\left[\tau^4\frac{\sigma^2}{m}+{
}\tau^3G_q\right]+\frac{(q+1)\gamma^2 L\eta^2\tau\sigma^2}{2m}+\frac{\gamma^2\eta^2L}{2}G_q \label{eq:final}
\end{align}
where \text{\ding{192}} comes from Lemma~\ref{lemma:var-red-main} and \text{\ding{193}} follows by imposing the following condition:
\begin{align}
    1-10\eta^2(\eta\gamma)^2(q+1)L^4\tau^4-(q+1)L\eta\gamma\tau\geq 0.
\end{align}
Rearranging Eq.~(\ref{eq:final}) we obtain:
\begin{align}
    &\left(1-{20\eta^2L^2
\tau^2}\right)\frac{1}{R}\sum_{r=0}^{R-1}\|\nabla f({\boldsymbol{w}}^{(r)})\|_2^2\nonumber\\
&\quad\leq \frac{2\left(f(\boldsymbol{w}^{(0)})-f(\boldsymbol{w}^*)\right)}{\tau \eta\gamma R}+\frac{(q+1)\gamma L\eta\sigma^2}{m}+{36\eta^2L^2\tau\sigma^2}+{10\eta^2
L^4\tau^3(\eta\gamma)^2(q+1)\frac{\sigma^2}{m}}\nonumber\\
&\quad+{10\eta^2
L^4\tau^2(\eta\gamma)^2(q+1){G_q}} +\underbrace{\frac{8\eta^2L^2}{m\tau R}\sum_{j=1}^m\sum_{c=0}^{\tau-1}\left\|\sum_{c=0}^{\tau-1}\left({\mathbf{g}}_j^{(c,0)}-{\mathbf{g}}^{(0)}\right)\right\|^2}_{(\mathrm{IV})}+\frac{\gamma\eta L}{\tau}G_q,
\end{align}
and the claim follows.

{ The final step is to simplify the term $(\mathrm{IV})$. To this purpose, first notice that 
\begin{align}
    \frac{8\eta^2L^2}{m\tau R}&\sum_{j=1}^m\sum_{c=0}^{\tau-1}\left\|\sum_{c=0,r=0}^{\tau-1}\left({\mathbf{g}}_j^{(c,0)}-{\mathbf{g}}^{(0)}\right)\right\|^2\nonumber\\
    &=\frac{8\eta^2L^2\tau}{m\tau R}\sum_{j=1}^m\left\|\sum_{c=0}^{\tau-1}\left({\mathbf{g}}_j^{(c,0)}-{\mathbf{g}}^{(0)}\right)\right\|^2\nonumber\\
    &\stackrel{\text{\ding{192}}}{\leq}\frac{8\eta^2L^2\tau}{m R}\sum_{j=1}^m\sum_{c=0}^{\tau-1}\left\|{\mathbf{g}}_j^{(c,0)}-{\mathbf{g}}^{(0)}\right\|^2\nonumber\\
    &\leq \frac{16\eta^2L^2\tau}{m R}\sum_{j=1}^m\sum_{c=0}^{\tau-1}\left[\left\|{\mathbf{g}}_j^{(c,0)}\right\|^2+\left\|{\mathbf{g}}^{(0)}\right\|^2\right]\nonumber\\
    &=\frac{16\eta^2L^2\tau^2}{m R}\sum_{j=1}^m\left[\frac{1}{\tau}\sum_{c=0}^{\tau-1}\left\|{\mathbf{g}}_j^{(c,0)}\right\|^2\right]+\frac{16\eta^2L^2\tau}{m R}\sum_{j=1}^m\sum_{c=0}^{\tau-1}\left\|{\mathbf{g}}^{(0)}\right\|^2\nonumber\\
    &\stackrel{\text{\ding{193}}}{\leq}\frac{16\eta^2L^2\tau^2}{m R}\sum_{j=1}^m\left[\frac{2\left[f_j(\boldsymbol{w}^{(0)}_j)-f_j(\boldsymbol{w}^{(*)}_j)\right]}{\eta\tau}+\eta\sigma^2\right]+\frac{16\eta^2L^2\tau^2}{ R}\left\|{\mathbf{g}}^{(0)}\right\|^2\nonumber\\
    &=\frac{32\eta L^2\tau}{m R}\sum_{j=1}^m\left[f_j(\boldsymbol{w}^{(0)}_j)-f_j(\boldsymbol{w}^{(*)}_j)\right]+\frac{16\eta^3L^2\tau^2}{ R}\sigma^2+\frac{16\eta^2L^2\tau^2}{ R}\left\|{\mathbf{g}}^{(0)}\right\|^2\nonumber\\
    &\stackrel{\text{\ding{194}}}{\leq} \frac{32\eta L^2\tau}{m R}\sum_{j=1}^m\left[f_j(\boldsymbol{w}^{(0)}_j)-f_j(\boldsymbol{w}^{(*)}_j)\right]+\frac{16\eta^3L^2\tau^2}{ R}\sigma^2+\frac{32\eta^2L^3\tau^2}{ R}\left(f(\boldsymbol{w}^{(0)})-f(\boldsymbol{w}^{(*)})\right)
    \label{eq:bnding_init_diff}
\end{align}
where \text{\ding{192}} comes from $\|\sum_{i=1}^n\mathbf{a}_i\|^2\leq n\sum_{i=1}^n\|\mathbf{a}_i\|^2$, in \text{\ding{193}} we used the standard convergence proof of gradient descent for non-convex objectives \cite{bottou2018optimization}, where $\boldsymbol{w}^{(*)}_j$ is the local minimizer of objective function $f_j(.)$, and  finally, \text{\ding{194}} follows from (smoothness assumption) inequality $\left\|{\mathbf{g}}^{(0)}\right\|^2\leq 2L\left(f(\boldsymbol{w}^{(0)})-f(\boldsymbol{w}^{(*)})\right)$ (see \cite{gower2019sgd,stich2019error} for more details). This completes the proof.
}

\end{proof}

\begin{remark}
If we let $\eta\gamma =\frac{1}{L}\sqrt{\frac{m}{R\tau\left(q+1\right)}}$, and want to make sure that the condition in Eq.~(\ref{eq:lr-cnd-varred}) is satisfied simultaneously, we need to have 
\begin{align}
1\geq \frac{10m^2\tau^2}{\gamma^2R^2(q+1)}+\sqrt{\frac{m\tau\left(q+1\right)}{R}}
\end{align}
This inequality is a polynomial of degree 4 with respect to $R$, therefore characterizing exact solution could be difficult. So, by letting $\gamma\geq \sqrt{20}m$ we derive an necessary solution here as follows:
\begin{align}
    R\geq m\tau\left(\frac{q+1}{2}\right)
\end{align}
We note that if we solve this inequality such as Eq.~(\ref{eq:iidexact}) we are expecting to degrade the dependency on $q$. This condition requires having  $R=\left(\frac{q+1}{m\epsilon}\right)$ and $\tau=\left(\frac{1}{m\epsilon}\right)$.
\end{remark}
\begin{corollary}[Linear speed up with fix global learning rate]
Considering the condition $30\eta^2L^2\tau^2\leq 1$, we have $1-20\eta^2L^2=\Theta\left(1\right)$. Therefore, in Eq.~(\ref{eq:thm1-result-varredd}) if we set $\eta\gamma=O\left(\frac{1}{L}\sqrt{\frac{m}{R\tau\left(q+1\right)}}\right)$, $\gamma\geq m$ leads to:
{
\begin{align}
    &\frac{1}{R}\sum_{r=0}^{R-1}\left\|\nabla f({\boldsymbol{w}}^{(r)})\right\|_2^2\nonumber\\
    &\leq O\Big(\frac{L\sqrt{q+1 }\left(f(\boldsymbol{w}^{(0)})-f(\boldsymbol{w}^{(*)})\right)}{\sqrt{mR\tau}}+\frac{\sqrt{q+1 }\sigma^2}{\sqrt{mR\tau}}+\frac{m\sigma^2}{R\left(q+1\right)\gamma^2}+\frac{m
\sigma^2\tau}{\left(q+1\right)\gamma^2R^2}+\frac{m^2
G_q}{\left(q+1\right)\gamma^2R^2}\nonumber\\
&\quad+ \frac{L\sqrt{\tau}}{\gamma\sqrt{q+1}\sqrt{m} R^{1.5}}\sum_{j=1}^m\left[f_j(\boldsymbol{w}^{(0)}_j)-f_j(\boldsymbol{w}^{(*)}_j)\right]+\frac{16m\sqrt{m}\sqrt{\tau}\sigma^2}{ L\gamma^3 R^2(q+1)\sqrt{R(q+1)}}+\frac{Lm}{\gamma^2  R^2(q+1)}\left(f(\boldsymbol{w}^{(0)})-f(\boldsymbol{w}^{(*)})\right)\nonumber\\
&\quad+\frac{mG_q}{R\tau\gamma^2\left(q+1\right)}\Big),\nonumber
\end{align}}
then by letting $\gamma\geq m$ we improve the convergence rate of \cite{karimireddy2019scaffold} and \cite{reddi2020adaptive} with tuned global and local learning rates, showing that we can archive the error $\epsilon$ with $R=\Theta\left(\left(1+q\right)\epsilon^{-1}\right)$ and $\tau=\Theta\left(\frac{1}{m\epsilon}\right)$, which matches the communication and computational complexity of \cite{karimireddy2019scaffold} and \cite{liang2019variance}, which shows that obtained rate is tight. We highlight that the communication complexity of our algorithm is better than \cite{karimireddy2019scaffold} in terms of number of bits per iteration as we do not use additional control variable. 
\end{corollary}

\begin{remark}
We note that the conditions in Eq.~(\ref{eq:lr-cnd-varred}) can be rewritten as 
\begin{align}
1-10\eta^2(\eta\gamma)^2(q+1)L^4\tau^4-L(q+1)\eta\gamma\tau\geq 0\quad \& \quad 30\eta^2L^2\tau^2\leq 1 
\end{align}
which implies that the choice of $\eta\leq \frac{1}{L\gamma (q+1) \tau \sqrt{30}}$ satisfies both conditions for $\gamma\geq m$.
\end{remark}

\subsection{Main result for the PL/strongly convex  setting}

\begin{theorem}[Strongly convex or PL]\label{thm:scvx-varred}
For \texttt{FedCOMGATE}$(\tau, \eta, \gamma)$, for all $0\leq t\leq R\tau-1$,  under Assumptions \ref{Assu:1}, \ref{Assu:09}, \ref{Assu:2}, \ref{assum:009} and \ref{assum:pl} and if the learning rate satisfies \begin{align}
1-(q+1)L\eta\gamma\tau-\frac{10(q+1)\eta^2\tau^4L^4(\eta\gamma)^2}{1-\mu\tau\gamma\eta+{20\mu\gamma\eta^3L^2\tau^3}}\geq 0\quad \& \quad 30\eta^2L^2\tau^2\leq 1
\end{align}
and all local model parameters are initialized at the same point ${\boldsymbol{w}}^{(0)}$, we obtain:
\begin{align}
    &  \mathbb{E}\Big[f({\boldsymbol{w}}^{(R)})-f({\boldsymbol{w}}^{(*)})\Big]\nonumber\\
    &\leq{\left(1-\frac{\mu \eta\gamma\tau}{3}\right)}^R\left(f(\boldsymbol{w}^{(0)})-f(\boldsymbol{w}^{(*)})\right)\nonumber\\
&\quad+\frac{3}{\mu}\Big[{L^218\eta^2\tau\sigma^2}+ \frac{8L^4\eta^2\tau^2}{m}\sum_{j=1}^m\left\|\left(\boldsymbol{w}_j^{(0,0)}-\boldsymbol{w}_j^{(*)}\right)\right\|^2_2+{16L^3\tau^2\eta^2}\left(f(\boldsymbol{w}^{(0)})-f(\boldsymbol{w}^{(*)})\right)  \nonumber\\
&\quad+5L^4{\eta}^2\tau^3(\eta\gamma)^2(q+1)\frac{\sigma^2}{m}+5L^2{\eta^2
L^2}\tau^2(\eta\gamma)^2(q+1)G_q+\frac{(q+1)\eta\gamma L}{2}\frac{\sigma^2}{m}+\frac{L\eta\gamma G_q}{2\tau}\Big]. \label{eq:main_result_for_PL}
\end{align}
\end{theorem}

\begin{proof} To prove our claim we use the following lemma. The proof of this intermediate lemma is deferred to Appendix~\ref{sec:appendix:omitted}.

\begin{lemma}\label{lemm:pl-main-lemm}

With $30\eta^2L^2\tau^2\leq 1$, under Assumptions~\ref{Assu:1}, \ref{Assu:09}, \ref{Assu:2} and \ref{assum:009} we have:
\begin{align}
&\frac{1}{m}\sum_{j=1}^m\sum_{c=0,r}^{\tau-1}\mathbb{E}\left\|\boldsymbol{w}_j^{(c,r)}-\boldsymbol{w}^{(r)}\right\|^2
\nonumber\\
&
=\frac{1}{m}\sum_{j=1}^m\sum_{c=0,r}^{\tau-1}\mathbb{E}\left\|\eta\sum_{c=0}^{\tau-1}\tilde{\boldsymbol{d}}_{j}^{(c,r)}\right\|^2\nonumber\\
&\leq {36\eta^2\tau^2\sigma^2}+ \frac{8\eta^2}{m}\sum_{j=1}^m\sum_{c=0}^{\tau-1}\left\|\sum_{c=0,r=0}^{\tau-1}\left({\mathbf{g}}_j^{(c,0)}-{\mathbf{g}}^{(0)}\right)\right\|^2 \nonumber\\
&\quad+{10\eta^2
L^2}\tau^4(\eta\gamma)^2(q+1)\sum_{c=0,r-1}^{\tau-1}\Big\|\frac{1}{m}\sum_{j=1}^m{\mathbf{g}}_j^{(c,r-1)}\Big\|^2\nonumber\\
&\quad+{10\eta^2
L^2}\tau^4(\eta\gamma)^2(q+1)\frac{\sigma^2}{m}+{10\eta^2
L^2}\tau^3(\eta\gamma)^2(q+1)G_q+{20\eta^2
\tau^2}\sum_{c=0}^{\tau-1}\Big\|{\mathbf{g}}^{(r)}\Big\|^2
\end{align}
\end{lemma}

Now we proceed to prove the claim of Theorem~\ref{thm:scvx-varred}. Note that
\begin{align}
     \mathbb{E}&\Big[f({\boldsymbol{w}}^{(r+1)})-f({\boldsymbol{w}}^{(r)})\Big]\nonumber\\
     &\leq 
     \frac{1}{2}\gamma\eta\sum_{c=0,r}^{\tau-1}\left[-\|\nabla f({\boldsymbol{w}}^{(r)})\|_2^2-\|\sum_{j=1}^m\frac{1}{m}\nabla{f}_j(\boldsymbol{w}_j^{(c,r)})\|_2^2+\frac{L^2}{m}\sum_{j=1}^m\mathbb{E}\left\|\boldsymbol{w}_j^{(c,r)}-\boldsymbol{w}^{(r)}\right\|^2\right]\nonumber\\
     &\quad+\frac{\gamma^2 L}{2}\left[(q+1)\eta^2\tau\frac{\sigma^2}{m}+(q+1)\eta^2\tau\sum_{c=0}^{\tau-1}\left\|\frac{1}{m}\sum_{j=1}^m{\mathbf{g}}_j^{(c,r)}\right\|^2+\eta^2G_q\right]\nonumber\\
     &=-\frac{\tau\gamma\eta}{2}\|\nabla f({\boldsymbol{w}}^{(r)})\|_2^2+\frac{1}{2}\gamma\eta\sum_{c=0,r}^{\tau-1}\left[-\|\sum_{j=1}^m\frac{1}{m}\nabla{f}_j(\boldsymbol{w}_j^{(c,r)})\|_2^2+\frac{L^2}{m}\sum_{j=1}^m\mathbb{E}\left\|\boldsymbol{w}_j^{(c,r)}-\boldsymbol{w}^{(r)}\right\|^2\right]\nonumber\\
     &\quad+\frac{\gamma^2 L}{2}\left[(q+1)\eta^2\tau\frac{\sigma^2}{m}+(q+1)\eta^2\tau\sum_{c=0}^{\tau-1}\left\|\frac{1}{m}\sum_{j=1}^m{\mathbf{g}}_j^{(c,r)}\right\|^2+\eta^2G_q\right]
\end{align}
which leads to the following:
\begin{align}
    \mathbb{E}&\Big[f({\boldsymbol{w}}^{(r+1)})-f({\boldsymbol{w}}^{(r)})\Big]\nonumber\\
     &=-\frac{\tau\gamma\eta}{2}\|\nabla f({\boldsymbol{w}}^{(r)})\|_2^2-\frac{1}{2}\gamma\eta\sum_{c=0,r}^{\tau-1}\|\sum_{j=1}^m\frac{1}{m}\nabla{f}_j(\boldsymbol{w}_j^{(c,r)})\|_2^2\nonumber\\
     &\quad+\frac{1}{2}\gamma\eta L^2\frac{1}{m}\sum_{j=1}^m\sum_{c=0,r}^{\tau-1}\mathbb{E}\left\|\boldsymbol{w}_j^{(c,r)}-\boldsymbol{w}^{(r)}\right\|^2\nonumber\\
     &\quad+\frac{\gamma^2 L}{2}\left[(q+1)\eta^2\tau\frac{\sigma^2}{m}+(q+1)\eta^2\tau\sum_{c=0}^{\tau-1}\left\|\frac{1}{m}\sum_{j=1}^m{\mathbf{g}}_j^{(c,r)}\right\|^2+\eta^2G_q\right]\nonumber\\
     &\leq -\frac{\tau\gamma\eta}{2}\|\nabla f({\boldsymbol{w}}^{(r)})\|_2^2-\frac{1}{2}\gamma\eta\sum_{c=0,r}^{\tau-1}\|\sum_{j=1}^m\frac{1}{m}\nabla{f}_j(\boldsymbol{w}_j^{(c,r)})\|_2^2\nonumber\\
     &\quad+ 18L^2{\gamma\eta\eta^2\tau^2\sigma^2}+ L^2\gamma\eta\frac{4\eta^2}{m}\sum_{j=1}^m\sum_{c=0}^{\tau-1}\left\|\sum_{c=0,r=0}^{\tau-1}\left({\mathbf{g}}_j^{(c,0)}-{\mathbf{g}}^{(0)}\right)\right\|^2 \nonumber\\
&\quad+\gamma\eta{5\eta^2
L^4}\tau^4(\eta\gamma)^2(q+1)\sum_{c=0,r-1}^{\tau-1}\Big[\Big\|\frac{1}{m}\sum_{j=1}^m{\mathbf{g}}_j^{(c,r-1)}\Big\|^2\nonumber\\
&\quad+5L^2\gamma\eta{\eta^2
L^2}\tau^4(\eta\gamma)^2(q+1)\frac{\sigma^2}{m}+5L^2\gamma\eta{\eta^2
L^2}\tau^3(\eta\gamma)^2(q+1)G_q\Big]+\gamma\eta{10\eta^2
\tau^3}L^2\Big\|{\mathbf{g}}^{(r)}\Big\|^2\nonumber\\
     &\quad+\frac{\gamma^2 L}{2}\left[(q+1)\eta^2\tau\frac{\sigma^2}{m}+(q+1)\eta^2\tau\sum_{c=0}^{\tau-1}\left\|\frac{1}{m}\sum_{j=1}^m{\mathbf{g}}_j^{(c,r)}\right\|^2+\eta^2G_q\right]\nonumber\\
     &=-\left(\frac{\tau\gamma\eta}{2}-\gamma L^2\eta{10\eta^2
\tau^3}\right)\|\nabla f({\boldsymbol{w}}^{(r)})\|_2^2\nonumber\\
&\quad-\frac{1}{2}\gamma\eta\left(1-(q+1)L\eta\gamma\tau\right)\sum_{c=0,r}^{\tau-1}\|\sum_{j=1}^m\frac{1}{m}\nabla{f}_j(\boldsymbol{w}_j^{(c,r)})\|_2^2+\frac{\gamma\eta}{2}{10\eta^2
L^4}\tau^4(\eta\gamma)^2(q+1)\sum_{c=0,r-1}^{\tau-1}\|\sum_{j=1}^m\frac{1}{m}\nabla{f}_j(\boldsymbol{w}_j^{(c,r-1)})\|_2^2\nonumber\\
     &\quad+ {L^218\gamma\eta\eta^2\tau^2\sigma^2}+ L^2\gamma\eta\frac{4\eta^2}{m}\sum_{j=1}^m\sum_{c=0}^{\tau-1}\left\|\sum_{c=0,r=0}^{\tau-1}\left({\mathbf{g}}_j^{(c,0)}-{\mathbf{g}}^{(0)}\right)\right\|^2 \nonumber\\
&\quad+5L^4\gamma\eta{\eta^2}\tau^4(\eta\gamma)^2(q+1)\frac{\sigma^2}{m}+5L^2\gamma\eta{\eta^2
L^2}\tau^3(\eta\gamma)^2(q+1)G_q+\frac{(q+1)\eta^2\gamma^2 L}{2}\frac{\tau\sigma^2}{m}+\frac{L\eta^2\gamma^2G_q}{2}\nonumber\\
&\stackrel{\text{\ding{192}}}{\leq}-\left(\frac{\tau\gamma\eta}{2}-\gamma L^2\eta{10\eta^2
\tau^3}\right)\|\nabla f({\boldsymbol{w}}^{(r)})\|_2^2+\frac{\gamma\eta}{2}{10\eta^2
L^4}\tau^4(\eta\gamma)^2(q+1)\sum_{c=0,r-1}^{\tau-1}\|\sum_{j=1}^m\frac{1}{m}\nabla{f}_j(\boldsymbol{w}_j^{(c,r-1)})\|_2^2\nonumber\\
     &\quad+ {L^218\gamma\eta\eta^2\tau^2\sigma^2}+ L^2\gamma\eta\frac{4\eta^2}{m}\sum_{j=1}^m\sum_{c=0}^{\tau-1}\left\|\sum_{c=0,r=0}^{\tau-1}\left({\mathbf{g}}_j^{(c,0)}-{\mathbf{g}}^{(0)}\right)\right\|^2 \nonumber\\
&\quad+5L^4\gamma\eta{\eta^2}\tau^4(\eta\gamma)^2(q+1)\frac{\sigma^2}{m}+5L^2\gamma\eta{\eta^2
L^2}\tau^3(\eta\gamma)^2(q+1)G_q+\frac{\eta^2\gamma^2 L}{2}\frac{(q+1)\tau\sigma^2}{m}+\frac{L\eta^2\gamma^2G_q}{2}\label{eq:pl-midstep-fl}
\end{align}
where \text{\ding{192}} follows from 
\begin{align}
1-(q+1)L\eta\gamma\tau\geq 0
\end{align}
Next Eq.~(\ref{eq:pl-midstep-fl}) leads us to 
\begin{align}
        \mathbb{E}&\Big[f({\boldsymbol{w}}^{(r+1)})-f({\boldsymbol{w}}^{(*)})\Big]=a_{r+1}\nonumber\\
        &{\leq} \left(1-\mu\tau\gamma\eta+{20\mu\gamma\eta^3L^2\tau^3}\right)\left(f(\boldsymbol{w}^{(r)})-f(\boldsymbol{w}^{(*)})\right)+\frac{\gamma\eta}{2}{10\eta^2
L^4}\tau^4(\eta\gamma)^2(q+1)\sum_{c=0,r-1}^{\tau-1}\|\sum_{j=1}^m\frac{1}{m}\nabla{f}_j(\boldsymbol{w}_j^{(c,r-1)})\|_2^2\nonumber\\
&\quad+{L^218\gamma\eta\eta^2\tau^2\sigma^2}+L^2\gamma\eta\frac{4\eta^2}{m}\sum_{j=1}^m\sum_{c=0}^{\tau-1}\left\|\sum_{c=0,r=0}^{\tau-1}\left({\mathbf{g}}_j^{(c,0)}-{\mathbf{g}}^{(0)}\right)\right\|^2 \nonumber\\
&\quad+5L^4\gamma\eta{\eta^2}\tau^4(\eta\gamma)^2(q+1)\frac{\sigma^2}{m}+5L^2\gamma\eta{\eta^2
L^2}\tau^3(\eta\gamma)^2(q+1)G_q+\frac{(q+1)\eta^2\gamma^2 L}{2}\frac{\tau\sigma^2}{m}+\frac{L\eta^2\gamma^2G_q}{2}\nonumber\\
&\stackrel{\text{\ding{192}}}{=}\Delta a_r+\frac{\gamma\eta}{2}{10\eta^2
L^4}\tau^4(\eta\gamma)^2(q+1)e_{r-1}+c\nonumber\\
&\stackrel{(a)}{\leq }\Delta \left(\Delta a_{r-1}+\frac{\eta\gamma}{2}\left(1-(q+1)L\eta\gamma\tau\right)e_{r-1}+\frac{\gamma\eta}{2}{10\eta^2
L^4}\tau^4(\eta\gamma)^2(q+1)e_{r-2}+c\right)+\frac{1}{2}\gamma^2\eta^2{10\eta^2
L^4}\tau^4(\eta\gamma)^2(q+1)e_{r-1}+c\nonumber\\
&=\Delta^2 a_{r-1}+\frac{\eta\gamma}{2}\left(\Delta-\Delta(q+1)L\eta\gamma\tau-10\eta^2
L^4\tau^4(\eta\gamma)^2(q+1)\right)e_{r-1}+\frac{\Delta\gamma^2\eta^2}{2}{10\eta^2
L^4}\tau^4(\eta\gamma)^2(q+1)e_{r-2}\nonumber\\
&\qquad+(\Delta+1) c\nonumber\\
&\stackrel{(b)}{\leq}\Delta^2 a_{r-1}+\frac{\Delta\gamma\eta}{2}{10\eta^2
L^4}\tau^4(\eta\gamma)^2(q+1)e_{r-2}+c\Delta+c\nonumber\\ 
&=\Delta\left(\Delta a_{r-1}+\frac{\gamma\eta}{2}{10\eta^2
L^4}\tau^4(\eta\gamma)^2(q+1)e_{r-2}+c\right)+c\nonumber\\
&\stackrel{(d)}{\leq}\Delta\left(\Delta\left(\Delta a_{r-2}+\frac{\gamma\eta}{2}{10\eta^2
L^4}\tau^4(\eta\gamma)^2(q+1)e_{r-3}+c\right)+c\right)+c\nonumber\\
&\stackrel{(e)}{\leq}\Delta^r a_{0}+\Delta^{r-1}\frac{\gamma\eta}{2}{10\eta^2
L^4}\tau^4(\eta\gamma)^2(q+1)e_{-1}+\left(\Delta^{r-1}+\Delta^{r-2}+\ldots+1\right)c\nonumber\\
&\stackrel{(f)}{=}\Delta^r a_{0}+\left(\frac{1-\Delta^r}{1-\Delta}\right)c\nonumber\\
&\stackrel{\text{\ding{192}}}{=}\Delta^r\left(f(\boldsymbol{w}^{(0)})-f(\boldsymbol{w}^{(*)})\right)\nonumber\\
&\quad+\frac{1-\Delta^r}{1-\Delta}\Big[{L^218\gamma\eta\eta^2\tau^2\sigma^2}+ L^2\gamma\eta\frac{4\eta^2}{m}\sum_{j=1}^m\sum_{c=0}^{\tau-1}\left\|\sum_{c=0,r=0}^{\tau-1}\left({\mathbf{g}}_j^{(c,0)}-{\mathbf{g}}^{(0)}\right)\right\|^2 \nonumber\\
&\quad+5L^4\gamma\eta{\eta^2}\tau^4(\eta\gamma)^2(q+1)\frac{\sigma^2}{m}+5L^2\gamma\eta{\eta^2
L^2}\tau^3(\eta\gamma)^2(q+1)G_q+\frac{(q+1)\eta^2\gamma^2 L}{2}\frac{\tau\sigma^2}{m}+\frac{L\eta^2\gamma^2G_q}{2}\Big]\nonumber\\
&\leq \Delta^r\left(f(\boldsymbol{w}^{(0)})-f(\boldsymbol{w}^{(*)})\right)\nonumber\\
&\quad+\frac{1}{1-\Delta}\Big[{L^218\gamma\eta\eta^2\tau^2\sigma^2}+ L^2\gamma\eta\frac{4\eta^2}{m}\sum_{j=1}^m\sum_{c=0}^{\tau-1}\left\|\sum_{c=0,r=0}^{\tau-1}\left({\mathbf{g}}_j^{(c,0)}-{\mathbf{g}}^{(0)}\right)\right\|^2 \nonumber\\
&\quad+5L^4\gamma\eta{\eta^2}\tau^4(\eta\gamma)^2(q+1)\frac{\sigma^2}{m}+5L^2\gamma\eta{\eta^2
L^2}\tau^3(\eta\gamma)^2(q+1)G_q+\frac{(q+1)\eta^2\gamma^2 L}{2}\frac{\tau\sigma^2}{m}+\frac{L\eta^2\gamma^2G_q}{2}\Big]
\end{align}
where \text{\ding{192}} holds because of $\Delta=1-\mu\tau\gamma\eta+{20\mu\gamma\eta^3L^2\tau^3}$, and the following short hand notations:
\begin{align}
    a_r&=\mathbb{E}\Big[f({\boldsymbol{w}}^{(r)})-f({\boldsymbol{w}}^{(*)})\Big]\nonumber\\
    e_r&=\sum_{c=0,r}^{\tau-1}\|\sum_{j=1}^m\frac{1}{m}\nabla{f}_j(\boldsymbol{w}_j^{(c,r)})\|_2^2\nonumber\\
    c&={L^218\gamma\eta\eta^2\tau^2\sigma^2}+ L^2\gamma\eta\frac{4\eta^2}{m}\sum_{j=1}^m\sum_{c=0}^{\tau-1}\left\|\sum_{c=0,r=0}^{\tau-1}\left({\mathbf{g}}_j^{(c,0)}-{\mathbf{g}}^{(0)}\right)\right\|^2 \nonumber\\
&\quad+5L^4\gamma\eta{\eta^2}\tau^4(\eta\gamma)^2(q+1)\frac{\sigma^2}{m}+5L^2\gamma\eta{\eta^2
L^2}\tau^3(\eta\gamma)^2(q+1)G_q+\frac{\eta^2\gamma^2 L}{2}\frac{(q+1)\tau\sigma^2}{m}+\frac{L\eta^2\gamma^2G_q}{2}
\end{align}
(a) comes from reapplying the recursion. (b) is due to the condition 
\begin{align}
1-(q+1)L\eta\gamma\tau-\frac{10(q+1)\eta^2\tau^4L^4(\eta\gamma)^2}{\Delta}=1-(q+1)L\eta\gamma\tau-\frac{10(q+1)\eta^2\tau^4L^4(\eta\gamma)^2}{1-\mu\tau\gamma\eta+{20\mu\gamma\eta^3L^2\tau^3}}\geq 0
\end{align}
(d) comes from one step reapplying of recursion. (e) holds by repeating the recursion under the same condition of learning rate for $r-1$ times. Finally, (f) follows from $e_{-1}=0$, which leads the following bound:

\begin{align}
            \mathbb{E}&\Big[f({\boldsymbol{w}}^{(R)})-f({\boldsymbol{w}}^{(*)})\Big]\nonumber\\
        &{\leq}{\left(1-\mu \eta\gamma\tau\left(1-20\eta^2L^2\tau^2\right)\right)}^r\left(f(\boldsymbol{w}^{(0)})-f(\boldsymbol{w}^{(*)})\right)\nonumber\\
&\quad+\frac{1}{\mu \eta\gamma\tau\left(1-20\eta^2L^2\tau^2\right)}\Big[{L^218\gamma\eta\eta^2\tau^2\sigma^2}+ L^2\gamma\eta\frac{4\eta^2}{m}\sum_{j=1}^m\sum_{c=0}^{\tau-1}\left\|\sum_{c=0}^{\tau-1}\left({\mathbf{g}}_j^{(c,0)}-{\mathbf{g}}^{(0)}\right)\right\|^2 \nonumber\\
&\qquad+5L^4\gamma\eta{\eta^2}\tau^4(\eta\gamma)^2(q+1)\frac{\sigma^2}{m}+5L^2\gamma\eta{\eta^2
L^2}\tau^3(\eta\gamma)^2(q+1)G_q+\frac{(q+1)\eta^2\gamma^2 L}{2}\frac{\tau\sigma^2}{m}+\frac{L\eta^2\gamma^2G_q}{2}\Big]\nonumber\\
&={\left(1-\mu \eta\gamma\tau\left(1-20\eta^2L^2\tau^2\right)\right)}^r\left(f(\boldsymbol{w}^{(0)})-f(\boldsymbol{w}^{(*)})\right)\nonumber\\
&\quad+\frac{1}{\mu \left(1-20\eta^2L^2\tau^2\right)}\Big[{L^218\eta^2\tau\sigma^2}+ \frac{4L^2\eta^2}{m\tau}\sum_{j=1}^m\sum_{c=0}^{\tau-1}\left\|\sum_{c=0}^{\tau-1}\left({\mathbf{g}}_j^{(c,0)}-{\mathbf{g}}^{(0)}\right)\right\|^2 \nonumber\\
&\quad+5L^4{\eta}^2\tau^3(\eta\gamma)^2(q+1)\frac{\sigma^2}{m}+5L^2{\eta^2
L^2}\tau^2(\eta\gamma)^2(q+1)G_q+\frac{(q+1)\eta\gamma L}{2}\frac{\sigma^2}{m}+\frac{L\eta\gamma G_q}{2\tau}\Big]\nonumber\\
&\stackrel{\text{\ding{192}}}{\leq} {\left(1-\frac{\mu \eta\gamma\tau}{3}\right)}^r\left(f(\boldsymbol{w}^{(0)})-f(\boldsymbol{w}^{(*)})\right)\nonumber\\
&\quad+\frac{3}{\mu}\Big[{L^218\eta^2\tau\sigma^2}+ \underbrace{L^2\frac{4\eta^2}{m\tau}\sum_{j=1}^m\sum_{c=0}^{\tau-1}\left\|\sum_{c=0}^{\tau-1}\left({\mathbf{g}}_j^{(c,0)}-{\mathbf{g}}^{(0)}\right)\right\|^2}_{(\mathrm{V})} \nonumber\\
&\quad+5L^4{\eta}^2\tau^3(\eta\gamma)^2(q+1)\frac{\sigma^2}{m}+5L^2{\eta^2
L^2}\tau^2(\eta\gamma)^2(q+1)G_q+\frac{(q+1)\eta\gamma L}{2}\frac{\sigma^2}{m}+\frac{L\eta\gamma G_q}{2\tau}\Big]
\end{align}
where in \text{\ding{192}} we used the condition $30\eta^2L^2\tau^2\leq 1$.

{Finally we continue with bounding term $(\mathrm{V})$:
\begin{align}\label{simple}
    \frac{4L^2\eta^2}{m\tau}\sum_{j=1}^m\sum_{c=0}^{\tau-1}&\left\|\sum_{c=0}^{\tau-1}\left({\mathbf{g}}_j^{(c,0)}-{\mathbf{g}}^{(0)}\right)\right\|^2\nonumber\\
    &=\frac{4L^2\eta^2\tau}{m\tau}\sum_{j=1}^m\left\|\sum_{c=0}^{\tau-1}\left({\mathbf{g}}_j^{(c,0)}-{\mathbf{g}}_j^{(*)}+{\mathbf{g}}_j^{(*)}-{\mathbf{g}}^{(0)}\right)\right\|^2\nonumber\\
    &\leq \frac{8L^2\eta^2}{m}\sum_{j=1}^m\left\|\sum_{c=0}^{\tau-1}\left({\mathbf{g}}_j^{(c,0)}-{\mathbf{g}}_j^{(*)}\right)\right\|^2+\frac{8L^2\eta^2}{m}\sum_{j=1}^m\left\|\sum_{c=0}^{\tau-1}\left({\mathbf{g}}_j^{(*)}-{\mathbf{g}}^{(0)}\right)\right\|^2\nonumber\\
    &=\frac{8L^2\eta^2}{m}\sum_{j=1}^m\left\|\sum_{c=0}^{\tau-1}\left({\mathbf{g}}_j^{(c,0)}-{\mathbf{g}}_j^{(*)}\right)\right\|^2+\frac{8L^2\tau^2\eta^2}{m}\sum_{j=1}^m\left\|{\mathbf{g}}_j^{(*)}-{\mathbf{g}}^{(0)}\right\|^2\nonumber\\
    &\leq \frac{8L^2\eta^2\tau}{m}\sum_{j=1}^m\sum_{c=0}^{\tau-1}\left\|{\mathbf{g}}_j^{(c,0)}-{\mathbf{g}}_j^{(*)}\right\|^2+\frac{8L^2\tau^2\eta^2}{m}\sum_{j=1}^m\left\|{\mathbf{g}}_j^{(*)}-{\mathbf{g}}^{(0)}\right\|^2\nonumber\\
    &\stackrel{\text{\ding{192}}}{\leq} \frac{8L^4\eta^2\tau}{m}\sum_{j=1}^m\sum_{c=0}^{\tau-1}\left\|{\boldsymbol{w}}_j^{(c,0)}-{\boldsymbol{w}}_j^{(*)}\right\|^2+\frac{8L^2\tau^2\eta^2}{m}\sum_{j=1}^m\left\|{\mathbf{g}}_j^{(*)}-{\mathbf{g}}^{(0)}\right\|^2\nonumber\\
    &\stackrel{\text{\ding{193}}}{=}\frac{8L^4\eta^2\tau}{m}\sum_{j=1}^m\sum_{c=0}^{\tau-1}\left\|{\boldsymbol{w}}_j^{(c,0)}-{\boldsymbol{w}}_j^{(*)}\right\|^2+\frac{8L^2\tau^2\eta^2}{m}\sum_{j=1}^m\left\|{\mathbf{g}}^{(0)}\right\|^2\nonumber\\
    &=\frac{8L^4\eta^2\tau}{m}\sum_{j=1}^m\sum_{c=0}^{\tau-1}\left[\left\|{\boldsymbol{w}}_j^{(c,0)}-{\boldsymbol{w}}_j^{(*)}\right\|^2\right]+{8L^2\tau^2\eta^2}\left\|{\mathbf{g}}^{(0)}\right\|^2\nonumber\\
    &\stackrel{\text{\ding{194}}}{\leq} \frac{8L^4\eta^2\tau}{m}\sum_{j=1}^m\sum_{c=0}^{\tau-1}\left[\left(1-2\mu\eta\left(1-\eta L\right)\right)^{c}\left\|\left(\boldsymbol{w}_j^{(0,0)}-\boldsymbol{w}_j^{(*)}\right)\right\|^2_2+\frac{\eta\sigma^2}{\mu(1-\eta L)}\right]+{8L^2\tau^2\eta^2}\left\|{\mathbf{g}}^{(0)}\right\|^2\nonumber\\
    &\stackrel{\text{\ding{195}}}{\leq} \frac{8L^4\eta^2\tau^2}{m}\sum_{j=1}^m\left[\left\|\left(\boldsymbol{w}_j^{(0,0)}-\boldsymbol{w}_j^{(*)}\right)\right\|^2_2+\frac{\eta\sigma^2}{\mu(1-\eta L)}\right]+{8L^2\tau^2\eta^2}\left\|{\mathbf{g}}^{(0)}\right\|^2\nonumber\\
    &\stackrel{\text{\ding{196}}}{\leq} \frac{8L^4\eta^2\tau^2}{m}\sum_{j=1}^m\left\|\left(\boldsymbol{w}_j^{(0,0)}-\boldsymbol{w}_j^{(*)}\right)\right\|^2_2+\frac{8L^4\eta^3\tau^2\sigma^2}{\mu(1-\eta L)}+{16L^3\tau^2\eta^2}\left(f(\boldsymbol{w}^{(0)})-f(\boldsymbol{w}^{(*)})\right)
\end{align}

where \text{\ding{192}} comes from Assumption~\ref{Assu:1},  \text{\ding{193}} holds because at the  optimal local solution $\boldsymbol{w}_j^{*}$ of device $j$ we have $\mathbf{g}_j^{(*)}= \boldsymbol{0}$, \text{\ding{194}} comes from strong convexity  assumption for local cost functions where $\left\|{\boldsymbol{w}}_j^{(t,0)}-{\boldsymbol{w}}_j^{(*)}\right\|^2\leq \left(1-2\mu\eta\left(1-\eta L\right)\right)^{t}\left\|\left(\boldsymbol{w}_j^{(0,0)}-\boldsymbol{w}_j^{(*)}\right)\right\|^2_2+\frac{\eta\sigma^2}{\mu(1-\eta L)}$~\cite{needell2014stochastic}, \text{\ding{195}} holds due to the choice of learning rate $\eta$ such that $\left(1-2\mu\eta\left(1-\eta L\right)\right)^{c}\leq 1$, and  finally \text{\ding{196}} is due to smoothness assumption which implies $\left\|{\mathbf{g}}^{(0)}\right\|^2\leq 2L\left(f(\boldsymbol{w}^{(0)})-f(\boldsymbol{w}^{(*)})\right)$ holds at global optimal solution $\boldsymbol{w}^{(*)}$.}
\end{proof}

{
\begin{corollary}[Linear speed up]
To achieve linear speed up we set $\eta=\frac{1}{2L(q+1)\tau \gamma}$ and $\gamma\geq \sqrt{m\tau}$ in Eq.~(\ref{eq:main_result_for_PL}) which incurs:
\begin{align}
            \mathbb{E}\Big[f({\boldsymbol{w}}^{(R)})-f({\boldsymbol{w}}^{(*)})\Big]
&\leq \left(f(\boldsymbol{w}^{(0)})-f(\boldsymbol{w}^{(*)})\right)e^{-\left(\frac{ R}{6(q+1)\kappa}\right)}\nonumber\\
&\quad+\frac{3}{\mu}\Big[\frac{4.5\sigma^2}{\tau \gamma^2(q+1)^2}+ \frac{2L^2}{(q+1)^2\gamma^2 m}\sum_{j=1}^m\left\|\left(\boldsymbol{w}_j^{(0,0)}-\boldsymbol{w}_j^{(*)}\right)\right\|^2_2+\frac{2L\left(f(\boldsymbol{w}^{(0)})-f(\boldsymbol{w}^{(*)})\right)}{(q+1)^2\gamma^2} \nonumber\\
&\quad+\frac{\kappa\sigma^2}{(q+1)^2\gamma^2\left((q+1)\gamma\tau-0.5\right)}+\frac{5}{16(q+1)^3\tau\gamma^2}\frac{\sigma^2}{m}+\frac{5}{16(q+1)^3\tau^2\gamma^2}G_q\nonumber\\
&\quad+\frac{1}{4\tau}\frac{\sigma^2}{m}+\frac{G_q}{4(q+1)\tau^2}\Big]\label{eq:plspeedup} 
\end{align}
From Eq.~(\ref{eq:plspeedup}) we can see that to attain an $\epsilon$-accurate solution we can choose $$R=O\left(\kappa(q+1)\log\left(\frac{1}{\epsilon}\right)\right),\tau=O\left(\frac{1}{m\epsilon}\right),$$
as desired.
\end{corollary}
}

\newpage
\subsection{Main result for the general convex setting}

\begin{theorem}[Convex]\label{thm:cvx-noniid}
For a  convex function $f(\mathbf{\boldsymbol{w}})$, applying \texttt{FedCOMGATE}$(\tau, \eta, \gamma)$ (Algorithm~\ref{Alg:VRFLDL}) to optimize $\tilde{f}(\boldsymbol{w},\phi)=f(\boldsymbol{w})+\frac{\phi}{2}\left\|\boldsymbol{w}\right\|^2$,  for all $0\leq t\leq R\tau-1$,  under Assumptions~\ref{Assu:1},\ref{Assu:09}, \ref{Assu:2}, \ref{assum:009} if the learning rate satisfies \begin{align}
    1-(q+1)L\eta\gamma\tau-\frac{10(q+1)\eta^2\tau^4L^4(\eta\gamma)^2}{1-\mu\tau\gamma\eta+{20\mu\gamma\eta^3L^2\tau^3}}\geq 0\quad  \& \quad 30\eta^2L^2\tau^2\leq 1
\end{align}
and all the models are initialized with $\boldsymbol{w}^{(0)}$, with the choice of $\phi=\frac{1}{\sqrt{m\tau}}$ and $\eta=\frac{1}{2L\gamma\tau\left(1+q\right)}$ we obtain:
\begin{align}
    &  \mathbb{E}\Big[f({\boldsymbol{w}}^{(R)})-f({\boldsymbol{w}}^{(*)})\Big]\nonumber\\
    &\leq e^{-\frac{R}{6\left(1+{q}\right)L \sqrt{m\tau}}}\left(f(\boldsymbol{w}^{(0)})-f(\boldsymbol{w}^{(*)})\right)\nonumber\\
        &\quad+\Big[\frac{13.5\sqrt{m}\sigma^2}{(q+1)^2\gamma^2\sqrt{\tau}}+ \frac{6\sqrt{m\tau} L^2}{m (q+1)^2\gamma^2}\sum_{j=1}^m\left\|\left(\boldsymbol{w}_j^{(0,0)}-\boldsymbol{w}_j^{(*)}\right)\right\|^2_2+\frac{12L \sqrt{m\tau}}{\gamma^2(q+1)^2}\left(f(\boldsymbol{w}^{(0)})-f(\boldsymbol{w}^{(*)})\right).  \nonumber\\
&\quad+\frac{3\sqrt{m\tau }\kappa\sigma^2}{(q+1)^2\gamma^2\left((q+1)\gamma\tau-0.5\right)}+\frac{15\sigma^2}{16(q+1)^3\gamma^2 \sqrt{m\tau}}+\frac{15G_q\sqrt{m}}{16(q+1)^3\tau^{1.5}\gamma^2}+\frac{3\sigma^2}{4\sqrt{m\tau}}+\frac{3\sqrt{m}G_q}{4(q+1)\tau^{1.5}}\Big]\nonumber\\
        &\quad +\frac{1}{2\sqrt{m\tau}}\left\|\boldsymbol{w}^{(*)}\right\|^2 
\end{align}

\end{theorem}

\begin{proof}
Since $\tilde{f}(\boldsymbol{w}^{(r)},\phi)=f(\boldsymbol{w}^{(r)})+\frac{\phi}{2}\left\|\boldsymbol{w}^{(r)}\right\|^2$ is $\phi$-PL, according to Theorem~\ref{thm:scvx-varred}, we have:
\begin{align}
    \tilde{f}(\boldsymbol{w}^{(R)},\phi)-\tilde{f}(\boldsymbol{w}^{(*)},\phi)&={f}(\boldsymbol{w}^{(r)})+\frac{\lambda}{2}\left\|\boldsymbol{w}^{(r)}\right\|^2-\left({f}(\boldsymbol{w}^{(*)})+\frac{\lambda}{2}\left\|\boldsymbol{w}^{(*)}\right\|^2\right)\nonumber\\
    &\leq \left(1-\frac{\eta\gamma{\phi\tau}}{3}\right)^R\left(f(\boldsymbol{w}^{(0)})-f(\boldsymbol{w}^{(*)})\right)\nonumber\\
    &\quad+\frac{3}{\phi}\Big[{L^218\eta^2\tau\sigma^2}+ L^2\frac{4\eta^2}{m\tau}\sum_{j=1}^m\sum_{c=0}^{\tau-1}\left\|\sum_{c=0}^{\tau-1}\left({\mathbf{g}}_j^{(c,0)}-{\mathbf{g}}^{(0)}\right)\right\|^2 \nonumber\\
&\quad+5L^4{\eta}^2\tau^3(\eta\gamma)^2(q+1)\frac{\sigma^2}{m}+5L^2{\eta^2
L^2}\tau^2(\eta\gamma)^2(q+1)G_q+\frac{(q+1)\eta\gamma L}{2}\frac{\sigma^2}{m}+\frac{L\eta\gamma G_q}{2\tau}\Big]\label{eq:mid-cvx-nniid}
\end{align}
Next rearranging Eq.~(\ref{eq:mid-cvx-nniid}) and replacing $\mu$ with $\phi$, and using the short hand notation of
\begin{align}
    \mathcal{A}(\eta)&\triangleq \Big[{L^218\eta^2\tau\sigma^2}+ L^2\frac{4\eta^2}{m\tau}\sum_{j=1}^m\sum_{c=0}^{\tau-1}\left\|\sum_{c=0,r=0}^{\tau-1}\left({\mathbf{g}}_j^{(c,0)}-{\mathbf{g}}^{(0)}\right)\right\|^2 \nonumber\\
&\quad+5L^4{\eta}^2\tau^3(\eta\gamma)^2(q+1)\frac{\sigma^2}{m}+5L^2{\eta^2
L^2}\tau^2(\eta\gamma)^2(q+1)G_q+\frac{(q+1)\eta\gamma L}{2}\frac{\sigma^2}{m}+\frac{L\eta\gamma G_q}{2\tau}\Big]
\end{align} 

leads to the following error bound:
\begin{align}
    \tilde{f}(\boldsymbol{w}^{(R)},\phi)-f^*&\leq \left(1-\frac{\eta\gamma{\phi\tau}}{3}\right)^R\left(f(\boldsymbol{w}^{(0)})-f(\boldsymbol{w}^{(*)})\right)+\frac{3}{{\phi}}\mathcal{A}(\eta) +\frac{\phi}{2}\left(\left\|\boldsymbol{w}^{(*)}\right\|^2-\left\|\boldsymbol{w}^{(r)}\right\|^2\right)\nonumber\\
    &\leq e^{-\left(\frac{\eta\gamma{\phi\tau}}{3}\right)R}\left(f(\boldsymbol{w}^{(0)})-f(\boldsymbol{w}^{(*)})\right)+\frac{3}{{\phi}}\mathcal{A}(\eta) +\frac{\phi}{2}\left\|\boldsymbol{w}^{(*)}\right\|^2 
\end{align}
Next, if we set $\phi=\frac{1}{\sqrt{m\tau}}$ and $\eta=\frac{1}{2\left(1+q\right)L\gamma \tau}$, we obtain the following bound:
\begin{align}
        \tilde{f}(\boldsymbol{w}^{(R)},\phi)-f^*&\leq e^{-\frac{R}{6\left(1+{q}\right)L \sqrt{m\tau}}}\left(f(\boldsymbol{w}^{(0)})-f(\boldsymbol{w}^{(*)})\right)+3\sqrt{m\tau}\mathcal{A}(\frac{1}{2\left(1+{q}\right)L\gamma \tau}) +\frac{1}{2\sqrt{m\tau}}\left\|\boldsymbol{w}^{(*)}\right\|^2 \nonumber\\
        &=e^{-\frac{R}{6\left(1+{q}\right)L \sqrt{m\tau}}}\left(f(\boldsymbol{w}^{(0)})-f(\boldsymbol{w}^{(*)})\right)\nonumber\\
        &\quad+\Big[\frac{13.5\sqrt{m}\sigma^2}{(q+1)^2\gamma^2\sqrt{\tau}}+ \frac{3}{\sqrt{m}\left(q+1\right)^2\gamma^2\tau^{2.5}}\sum_{j=1}^m\sum_{c=0}^{\tau-1}\left\|\sum_{c=0,r=0}^{\tau-1}\left({\mathbf{g}}_j^{(c,0)}-{\mathbf{g}}^{(0)}\right)\right\|^2 \nonumber\\
&\quad+\frac{15\sigma^2}{16(q+1)^3\gamma^2 \sqrt{m\tau}}+\frac{15G_q\sqrt{m}}{16(q+1)^3\tau^{1.5}\gamma^2}+\frac{3\sigma^2}{4\sqrt{m\tau}}+\frac{3\sqrt{m}G_q}{4(q+1)\tau^{1.5}}\Big]\nonumber\\
        &\quad+\frac{1}{2\sqrt{m\tau}}\left\|\boldsymbol{w}^{(*)}\right\|^2 
\end{align}
{ Finally, using Eq.~(\ref{simple}) we obtain the bound:

\begin{align}
        \tilde{f}(\boldsymbol{w}^{(R)},\phi)-f^*&\leq e^{-\frac{R}{6\left(1+{q}\right)L \sqrt{m\tau}}}\left(f(\boldsymbol{w}^{(0)})-f(\boldsymbol{w}^{(*)})\right)\nonumber\\
        &\quad+\Big[\frac{13.5\sqrt{m}\sigma^2}{(q+1)^2\gamma^2\sqrt{\tau}}+ \frac{6 \sqrt{m \tau} L^2}{(q+1)^2 \gamma^2 m}\sum_{j=1}^m\left\|\left(\boldsymbol{w}_j^{(0,0)}-\boldsymbol{w}_j^{(*)}\right)\right\|^2_2+\frac{12L\sqrt{m \tau} }{\gamma^2(q+1)^2}\left(f(\boldsymbol{w}^{(0)})-f(\boldsymbol{w}^{(*)})\right) \nonumber\\
&\quad+\frac{3\sqrt{m\tau }\kappa\sigma^2}{(q+1)^2\gamma^2\left((q+1)\gamma\tau-0.5\right)}+\frac{15\sigma^2}{16(q+1)^3\gamma^2 \sqrt{m\tau}}+\frac{15G_q\sqrt{m}}{16(q+1)^3\tau^{1.5}\gamma^2}+\frac{3\sigma^2}{4\sqrt{m\tau}}+\frac{3\sqrt{m}G_q}{4(q+1)\tau^{1.5}}\Big]\nonumber\\
        &\quad+\frac{1}{2\sqrt{m\tau}}\left\|\boldsymbol{w}^{(*)}\right\|^2 
\end{align}
}
\end{proof}

{
\begin{corollary}\label{corol:cvx-nniid}
As a result of Theorem~\ref{thm:cvx-noniid}, for general convex functions with $\gamma\geq \sqrt{m\tau}$, to achieve the convergence error of ${\epsilon}$ we need to have $\tau=O\left(\frac{1}{m\epsilon^2}\right)$ and $R=O\left(\frac{L\left(1+q\right)}{\epsilon}\log\left(\frac{1}{\epsilon}\right)\right)$.
\end{corollary}}

\newpage
\section{Deferred Proofs}\label{sec:appendix:omitted}
\subsection{Proof of Lemma~\ref{lemma:var-red-main}}\label{sec:0}
We prove Lemma~\ref{lemma:var-red-main} in two steps. First, we prove the following lemma:
\begin{lemma}
Under Assumption~\ref{Assu:1} and \ref{Assu:2}, and the condition over learning rate $30\eta^2\tau^2L^2\leq 1$, we have the following inequality: 
\begin{align}
    \frac{1}{p}\sum_{j=1}^p\sum_{r=0}^{R-1}\sum_{c=0,r}^{\tau-1}\mathbb{E}\left\|\left({\boldsymbol{w}}^{(r)}-\boldsymbol{w}^{(c,r)}_j\right)\right\|^2&=\frac{1}{p}\sum_{j=1}^p\sum_{r=0}^{R-1}\sum_{c=0,r}^{\tau-1}\mathbb{E}\left\|\sum_{c=0,r}^{\tau-1}\tilde{\boldsymbol{d}}^{(c,r)}_j\right\|^2\nonumber\\
    &\leq {36R\eta^2\tau^2\sigma^2}+ {8\eta^2}C +{20\eta^2
\tau^2}\sum_{r=0}^{R-1}\sum_{c=0}^{\tau-1}\Big\|{\mathbf{g}}^{(r)}\Big\|^2\nonumber\\
&+\frac{10\eta^2
L^2\tau}{p}\sum_{j=1}^p\sum_{r=0}^{R-1}\sum_{c=0,r}^{\tau-1}\sum_{c=0,r-1}^{\tau-1}\Big\|{\boldsymbol{w}}^{(r)}-{\boldsymbol{w}}^{(r-1)}\Big\|^2
\end{align}
where $C=\frac{1}{p}\sum_{j=1}^p\sum_{c=0,r=0}^{\tau-1}\left\|\sum_{k=0,r=0}^{c}\left(\nabla{f}_j({\boldsymbol{w}}_j^{(k,r)})-\nabla{f}({\boldsymbol{w}}^{(r)})\right)\right\|^2$
\end{lemma}

First, we bound the term $\frac{1}{p}\sum_{j=1}^p\sum_{r=0}^{R-1}\sum_{c=0,r}^{\tau-1}\mathbb{E}\left\|\left({\boldsymbol{w}}^{(r)}-\boldsymbol{w}^{(c,r)}_j\right)\right\|^2$ for $r\geq 1$:

\begin{lemma}\label{lem:D2}
For $r\geq1$:
\begin{align}
    \frac{1}{p}\sum_{j=1}^p\mathbb{E}\left\|\sum_{c=0}^{\tau-1}\tilde{d}_j^{(c,r)}\right\|^2&\leq  18\sigma^2 \tau+\frac{1}{p}\sum_{j=1}^p\Big[6L^2\tau\Big[\sum_{c=0,r}^{\tau-1}\Big\|\Big[\boldsymbol{w}_{j}^{(c,r)}-{\boldsymbol{w}}^{(r)}\Big]\Big\|^2+\frac{1}{\tau}\sum_{c=0,r}^{\tau-1}\sum_{c=0,r-1}^{\tau-1}\Big\|{\boldsymbol{w}}^{(r)}-{\boldsymbol{w}}^{(r-1)}\Big\|^2\nonumber\\
&\quad+\sum_{c=0,r}^{\tau-1}\sum_{c=0,r-1}^{\tau-1}\Big\|{\boldsymbol{w}}^{(r-1)}-{\boldsymbol{w}}_j^{(c,r-1)}\Big\|^2+\sum_{c=0,r}^{\tau-1}\sum_{c=0,r-1}^{\tau-1}\frac{1}{p}\sum_{j=1}^p\Big\|{\boldsymbol{w}}_{j}^{(c,r-1)}-{\boldsymbol{w}}^{(r-1)}\Big\|^2\nonumber\\
&\quad+\sum_{c=0,r}^{\tau-1}\sum_{c=0,r-1}^{\tau-1}\Big\|{\boldsymbol{w}}^{(r-1)}-\boldsymbol{w}^{(r-1)}\Big\|^2+\frac{1}{L^2}\sum_{c=0,r-1}^{\tau-1}\Big\|{\mathbf{g}}^{(r-1)}\Big\|^2\Big]\Big]
\end{align}
\end{lemma}

\begin{proof}
\begin{align}
    \frac{1}{p}&\sum_{j=1}^p\mathbb{E}\left\|\sum_{c=0}^{\tau-1}\tilde{d}_j^{(c,r)}\right\|^2\nonumber\\
    &=\frac{1}{p}\sum_{j=1}^p\mathbb{E}\left\|\sum_{c=0}^{\tau-1}\left[\tilde{\mathbf{g}}_{j}^{(c,r)}+\frac{1}{\tau }\left(\frac{1}{p}\sum_{j=1}^p\sum_{c=0}^{\tau-1}\tilde{\mathbf{g}}_{j}^{(c,r-1)}-\sum_{c=0}^{\tau-1}\tilde{\mathbf{g}}_{j}^{(c,r-1)}\right)\right]\right\|^2\nonumber\\
    &=\frac{1}{p}\sum_{j=1}^p\mathbb{E}\Big\|\sum_{c=0}^{\tau-1}\Big[\left(\tilde{\mathbf{g}}_{j}^{(c,r)}-\mathbf{g}_{j}^{(c,r)}+\mathbf{g}_{j}^{(c,r)}\right)\nonumber\\
    &\quad+\frac{1}{\tau }\sum_{c=0}^{\tau-1}\left(\frac{1}{p}\sum_{j=1}^p\left(\tilde{\mathbf{g}}_{j}^{(c,r-1)}-{\mathbf{g}}_{j}^{(c,r-1)}+{\mathbf{g}}_{j}^{(c,r-1)}\right)-{\mathbf{g}}_{j}^{(c,r-1)}+{\mathbf{g}}_{j}^{(c,r-1)}-\tilde{\mathbf{g}}_{j}^{(c,r-1)}\right)\Big]\Big\|^2\nonumber\\
    &\leq \frac{2}{p}\sum_{j=1}^p\mathbb{E}\underbrace{\Big\|\sum_{c=0}^{\tau-1}\Big[\left(\tilde{\mathbf{g}}_{j}^{(c,r)}-\mathbf{g}_{j}^{(c,r)}\right)+\frac{1}{\tau }\sum_{c=0}^{\tau-1}\left(\frac{1}{p}\sum_{j=1}^p\left(\tilde{\mathbf{g}}_{j}^{(c,r-1)}-{\mathbf{g}}_{j}^{(c,r-1)}\right)+{\mathbf{g}}_{j}^{(c,r-1)}-\tilde{\mathbf{g}}_{j}^{(c,r-1)}\right)\Big]\Big\|^2}_{\mathrm{(I)}}\nonumber\\
        &\quad+ \frac{2}{p}\sum_{j=1}^p\underbrace{\Big\|\sum_{c=0}^{\tau-1}\Big[\mathbf{g}_{j}^{(c,r)}+\frac{2}{\tau }\sum_{c=0}^{\tau-1}\left(\frac{1}{p}\sum_{j=1}^p{\mathbf{g}}_{j}^{(c,r-1)}-{\mathbf{g}}_{j}^{(c,r-1)}\right)\Big]\Big\|^2}_{(\mathrm{II})}\label{eq:I-II}
\end{align}

\end{proof}
We first bound the term $(\mathrm{I})$ in Eq.~(\ref{eq:I-II}) with the following lemma:
\begin{lemma}
\begin{align}
    \frac{2}{p}\sum_{j=1}^p\mathbb{E}\Big\|\sum_{c=0}^{\tau-1}\Big[\left(\tilde{\mathbf{g}}_{j}^{(c,r)}-\mathbf{g}_{j}^{(c,r)}\right)+\frac{1}{\tau }\sum_{c=0}^{\tau-1}\left(\frac{1}{p}\sum_{j=1}^p\left(\tilde{\mathbf{g}}_{j}^{(c,r-1)}-{\mathbf{g}}_{j}^{(c,r-1)}\right)+{\mathbf{g}}_{j}^{(c,r-1)}-\tilde{\mathbf{g}}_{j}^{(c,r-1)}\right)\Big]\Big\|^2\leq 18\sigma^2 \tau
\end{align}
\end{lemma}
\begin{proof}
\begin{align}
    &\mathbb{E}\Big\|\sum_{c=0}^{\tau-1}\Big[\left(\tilde{\mathbf{g}}_{j}^{(c,r)}-\mathbf{g}_{j}^{(c,r)}\right)+\frac{1}{\tau }\sum_{c=0}^{\tau-1}\left(\frac{1}{p}\sum_{j=1}^p\left(\tilde{\mathbf{g}}_{j}^{(c,r-1)}-{\mathbf{g}}_{j}^{(c,r-1)}\right)+{\mathbf{g}}_{j}^{(c,r-1)}-\tilde{\mathbf{g}}_{j}^{(c,r-1)}\right)\Big]\Big\|^2\nonumber\\
    &\leq \mathbb{E}\Big\|\sum_{c=0}^{\tau-1}\left(\tilde{\mathbf{g}}_{j}^{(c,r)}-\mathbf{g}_{j}^{(c,r)}\right)\Big\|^2+\Big\|\frac{1}{\tau }\sum_{c=0}^{\tau-1}\sum_{c=0}^{\tau-1}\frac{1}{p}\sum_{j=1}^p\left(\tilde{\mathbf{g}}_{j}^{(c,r-1)}-{\mathbf{g}}_{j}^{(c,r-1)}\right)\Big\|^2+\Big\|\frac{1}{\tau }\sum_{c=0}^{\tau-1}\sum_{c=0}^{\tau-1}\left({\mathbf{g}}_{j}^{(c,r-1)}-\tilde{\mathbf{g}}_{j}^{(c,r-1)}\right)\Big\|^2\nonumber\\
    &= 3\left[\mathbb{E}\Big\|\sum_{c=0}^{\tau-1}\left(\tilde{\mathbf{g}}_{j}^{(c,r)}-\mathbf{g}_{j}^{(c,r)}\right)\Big\|^2+\Big\|\sum_{c=0}^{\tau-1}\frac{1}{p}\sum_{j=1}^p\left(\tilde{\mathbf{g}}_{j}^{(c,r-1)}-{\mathbf{g}}_{j}^{(c,r-1)}\right)\Big\|^2+\Big\|\sum_{c=0}^{\tau-1}\left({\mathbf{g}}_{j}^{(c,r-1)}-\tilde{\mathbf{g}}_{j}^{(c,r-1)}\right)\Big\|^2\right]\nonumber\\
    &\stackrel{\text{\ding{192}}}{=}3\left[\sum_{c=0}^{\tau-1}\mathbb{E}\Big\|\left(\tilde{\mathbf{g}}_{j}^{(c,r)}-\mathbf{g}_{j}^{(c,r)}\right)\Big\|^2+\sum_{c=0}^{\tau-1}\frac{1}{p}\sum_{j=1}^p\mathbb{E}\Big\|\left(\tilde{\mathbf{g}}_{j}^{(c,r-1)}-{\mathbf{g}}_{j}^{(c,r-1)}\right)\Big\|^2+\sum_{c=0}^{\tau-1}\mathbb{E}\Big\|\left({\mathbf{g}}_{j}^{(c,r-1)}-\tilde{\mathbf{g}}_{j}^{(c,r-1)}\right)\Big\|^2\right]\nonumber\\
    &\leq \tau\left(\sigma^2+\sigma^2+\sigma^2\right)\nonumber\\
    &=9\sigma^2 \tau
\end{align}
where \text{\ding{192}} follows from Assumption~\ref{Assu:2}.
\end{proof}

We bound the term $(\mathrm{II})$ in Eq.~(\ref{eq:I-II}) as follows:
\begin{lemma}\label{lemma:r>1}
For $r\geq 1$ we have:
\begin{align}
&\Big\|\sum_{c=0,r}^{\tau-1}\Big[\mathbf{g}_{j}^{(c,r)}+\frac{1}{\tau }\sum_{c=0,r}^{\tau-1}\left(\frac{1}{p}\sum_{j=1}^p{\mathbf{g}}_{j}^{(c,r-1)}-{\mathbf{g}}_{j}^{(c,r-1)}\right)\Big]\Big\|^2\\
&\leq 5L^2\Big[\tau\sum_{c=0,r}^{\tau-1}\Big\|\Big[\boldsymbol{w}_{j}^{(c,r)}-{\boldsymbol{w}}^{(r)}\Big]\Big\|^2+\sum_{c=0,r}^{\tau-1}\sum_{c=0,r-1}^{\tau-1}\Big\|{\boldsymbol{w}}^{(r)}-{\boldsymbol{w}}^{(r-1)}\Big\|^2\nonumber\\
&\quad+\tau\sum_{c=0,r-1}^{\tau-1}\Big\|{\boldsymbol{w}}_j^{(c,r-1)}-{\boldsymbol{w}}^{(r-1)}\Big\|^2+\tau\sum_{c=0,r-1}^{\tau-1}\frac{1}{p}\sum_{j=1}^p\Big\|{\boldsymbol{w}}_{j}^{(c,r-1)}-{\boldsymbol{w}}^{(r-1)}\Big\|^2\nonumber\\
&\quad+\tau\sum_{c=0,r-1}^{\tau-1}\Big\|{\boldsymbol{w}}_j^{(c,r-1)}-\boldsymbol{w}^{(r-1)}\Big\|^2+\tau\frac{1}{L^2}\sum_{c=0,r-1}^{\tau-1}\Big\|{\mathbf{g}}^{(r-1)}\Big\|^2\Big]
\end{align}
\end{lemma}
\begin{proof}
Adopting the notation $\mathbf{g}_j^{(r)}=\nabla{f}_j(\boldsymbol{w}^{(r)})$, we have:
\begin{align}
&\Big\|\sum_{c=0}^{\tau-1}\Big[\mathbf{g}_{j}^{(c,r)}+\frac{1}{\tau }\sum_{c=0}^{\tau-1}\left(\frac{1}{p}\sum_{j=1}^p{\mathbf{g}}_{j}^{(c,r-1)}-{\mathbf{g}}_{j}^{(c,r-1)}\right)\Big]\Big\|^2\nonumber\\
&= \Big\|\sum_{c=0}^{\tau-1}\Big[\mathbf{g}_{j}^{(c,r)}-{\mathbf{g}}_j^{(r)}+{\mathbf{g}}_j^{(r)}\nonumber\\
&\quad+\frac{1}{\tau}\sum_{c=0}^{\tau-1}\left(-{\mathbf{g}}_j^{(r-1)}+{\mathbf{g}}_j^{(r-1)}-{\mathbf{g}}_j^{(c,r-1)}\right)+\frac{1}{\tau }\sum_{c=0}^{\tau-1}\frac{1}{p}\sum_{j=1}^p\left({\mathbf{g}}_{j}^{(c,r-1)}-{\mathbf{g}}_j^{(r-1)}+{\mathbf{g}}_j^{(r-1)}\right)\Big]\Big\|^2\nonumber\\
&= \Big\|\sum_{c=0}^{\tau-1}\Big[\mathbf{g}_{j}^{(c,r)}-{\mathbf{g}}_j^{(r)}+{\mathbf{g}}_j^{(r)}-\frac{1}{\tau}\sum_{c=0}^{\tau-1}{\mathbf{g}}_j^{(r-1)}\nonumber\\
&\quad+\frac{1}{\tau}\sum_{c=0}^{\tau-1}\left({\mathbf{g}}_j^{(r-1)}-{\mathbf{g}}_j^{(c,r-1)}\right)+\frac{1}{\tau }\sum_{c=0}^{\tau-1}\frac{1}{p}\sum_{j=1}^p\left({\mathbf{g}}_{j}^{(c,r-1)}-{\mathbf{g}}_j^{(r-1)}\right)+{\mathbf{g}}^{(r-1)}\Big]\Big\|^2\nonumber\\
&\leq 5\Big[\Big\|\sum_{c=0,r}^{\tau-1}\Big[\mathbf{g}_{j}^{(c,r)}-{\mathbf{g}}_j^{(r)}\Big\|^2+\Big\|\sum_{c=0,r}^{\tau-1}\left({\mathbf{g}}_j^{(r)}-\frac{1}{\tau}\sum_{c=0,r-1}^{\tau-1}{\mathbf{g}}_j^{(r-1)}\right)\Big\|^2\nonumber\\
&\quad+\Big\|\frac{1}{\tau}\sum_{c=0,r}^{\tau-1}\sum_{c=0,r-1}^{\tau-1}\left({\mathbf{g}}_j^{(r-1)}-{\mathbf{g}}_j^{(c,r-1)}\right)\Big\|^2+\Big\|\frac{1}{\tau }\sum_{c=0,r}^{\tau-1}\sum_{c=0,r-1}^{\tau-1}\frac{1}{p}\sum_{j=1}^p\left({\mathbf{g}}_{j}^{(c,r-1)}-{\mathbf{g}}_j^{(r-1)}\right)\Big\|^2\nonumber\\
&\quad+\Big\|\sum_{c=0,r-1}^{\tau-1}{\mathbf{g}}^{(r-1)}\Big\|^2\Big]\nonumber\\
&\stackrel{\text{\ding{192}}}{=}5\Big[\Big\|\sum_{c=0,r}^{\tau-1}\Big[\mathbf{g}_{j}^{(c,r)}-{\mathbf{g}}_j^{(r)}\Big]\Big\|^2+\Big\|\sum_{c=0,r}^{\tau-1}\frac{1}{\tau}\sum_{c=0,r-1}^{\tau-1}\left({\mathbf{g}}_j^{(r)}-{\mathbf{g}}_j^{(r-1)}\right)\Big\|^2\nonumber\\
&\quad+\Big\|\frac{1}{\tau}\sum_{c=0,r}^{\tau-1}\sum_{c=0,r-1}^{\tau-1}\left({\mathbf{g}}_j^{(c,r-1)}-{\mathbf{g}}_j^{(r-1)}\right)\Big\|^2+\Big\|\frac{1}{\tau}\sum_{c=0,r}^{\tau-1}\sum_{c=0,r-1}^{\tau-1}\frac{1}{p}\sum_{j=1}^p\left({\mathbf{g}}_{j}^{(c,r-1)}-{\mathbf{g}}_j^{(r-1)}\right)\Big\|^2\nonumber\\
&\quad+\Big\|\sum_{c=0,r-1}^{\tau-1}{\mathbf{g}}^{(r-1)}\Big\|^2\Big]\label{eq:mid-bnd-II}
\end{align}
where \text{\ding{192}} holds due to $\frac{1}{\tau}\sum_{c=0,r}^{\tau-1}\left(\mathbf{g}_{j}^{(c,r)}-\frac{1}{\tau}\sum_{c=0,r-1}^{\tau-1}{\mathbf{g}}_j^{(r)}\right)=\sum_{c=0,r}^{\tau-1}\frac{1}{\tau}\sum_{c=0,r-1}^{\tau-1}\left({\mathbf{g}}_j^{(r)}-{\mathbf{g}}_j^{(r-1)}\right)$.
We continue with bounding Eq.~(\ref{eq:mid-bnd-II}):
\begin{align}
    &5\Big[\Big\|\sum_{c=0,r}^{\tau-1}\Big[\mathbf{g}_{j}^{(c,r)}-{\mathbf{g}}_j^{(r)}\Big]\Big\|^2+\Big\|\sum_{c=0,r}^{\tau-1}\frac{1}{\tau}\sum_{c=0,r-1}^{\tau-1}\left({\mathbf{g}}_j^{(r)}-{\mathbf{g}}_j^{(r-1)}\right)\Big\|^2\nonumber\\
&\quad+\Big\|\frac{1}{\tau}\sum_{c=0,r}^{\tau-1}\sum_{c=0,r-1}^{\tau-1}\left({\mathbf{g}}_j^{(c,r-1)}-{\mathbf{g}}_j^{(r-1)}\right)\Big\|^2+\Big\|\frac{1}{\tau}\sum_{c=0,r}^{\tau-1}\sum_{c=0,r-1}^{\tau-1}\frac{1}{p}\sum_{j=1}^p\left({\mathbf{g}}_{j}^{(c,r-1)}-{\mathbf{g}}_j^{(r-1)}\right)\Big\|^2\nonumber\\
&\quad+\Big\|\sum_{c=0,r-1}^{\tau-1}{\mathbf{g}}^{(r-1)}\Big\|^2\Big]\nonumber\\
&\stackrel{\text{\ding{193}}}{=}5\Big[\Big\|\sum_{c=0,r}^{\tau-1}\Big[\mathbf{g}_{j}^{(c,r)}-{\mathbf{g}}_j^{(r)}\Big]\Big\|^2+\Big\|\sum_{c=0,r}^{\tau-1}\frac{1}{\tau}\sum_{c=0,r-1}^{\tau-1}\left({\mathbf{g}}_j^{(r)}-{\mathbf{g}}_j^{(r-1)}\right)\Big\|^2\nonumber\\
&\quad+\Big\|\sum_{c=0,r-1}^{\tau-1}\left({\mathbf{g}}_j^{(c,r-1)}-{\mathbf{g}}_j^{(r-1)}\right)\Big\|^2+\Big\|\sum_{c=0,r-1}^{\tau-1}\frac{1}{p}\sum_{j=1}^p\left({\mathbf{g}}_{j}^{(c,r-1)}-{\mathbf{g}}_j^{(r-1)}\right)\Big\|^2\nonumber\\
&\quad+\Big\|\sum_{c=0,r-1}^{\tau-1}{\mathbf{g}}^{(r-1)}\Big\|^2\Big]\nonumber\\
&\leq 5\Big[\tau\sum_{c=0,r}^{\tau-1}\Big\|\Big[\mathbf{g}_{j}^{(c,r)}-{\mathbf{g}}_j^{(r)}\Big]\Big\|^2+\sum_{c=0,r}^{\tau-1}\sum_{c=0,r-1}^{\tau-1}\Big\|\left({\mathbf{g}}_j^{(r)}-{\mathbf{g}}_j^{(r-1)}\right)\Big\|^2\nonumber\\
&\quad+\sum_{c=0,r-1}^{\tau-1}\Big\|\left({\mathbf{g}}_j^{(c,r-1)}-{\mathbf{g}}_j^{(r-1)}\right)\Big\|^2+\sum_{c=0,r-1}^{\tau-1}\frac{1}{p}\sum_{j=1}^p\Big\|\left({\mathbf{g}}_{j}^{(c,r-1)}-{\mathbf{g}}_j^{(r-1)}\right)\Big\|^2\nonumber\\
&\quad+\tau\sum_{c=0,r-1}^{\tau-1}\Big\|{\mathbf{g}}^{(r-1)}\Big\|^2\Big]\nonumber\\
&\leq 5L^2\Big[\tau\sum_{c=0,r}^{\tau-1}\Big\|\Big[\boldsymbol{w}_{j}^{(c,r)}-{\boldsymbol{w}}^{(r)}\Big]\Big\|^2+\sum_{c=0,r}^{\tau-1}\sum_{c=0,r-1}^{\tau-1}\Big\|{\boldsymbol{w}}^{(r)}-{\boldsymbol{w}}^{(r-1)}\Big\|^2\nonumber\\
&\quad+\tau\sum_{c=0,r-1}^{\tau-1}\Big\|{\boldsymbol{w}}_j^{(c,r-1)}-{\boldsymbol{w}}^{(r-1)}\Big\|^2+\tau\sum_{c=0,r-1}^{\tau-1}\frac{1}{p}\sum_{j=1}^p\Big\|{\boldsymbol{w}}_{j}^{(c,r-1)}-{\boldsymbol{w}}^{(r-1)}\Big\|^2+\frac{\tau}{L^2}\sum_{c=0,r-1}^{\tau-1}\Big\|{\mathbf{g}}^{(r-1)}\Big\|^2\Big]
\end{align}
where \text{\ding{193}} is due to the fact that 
\begin{align}
&\Big\|\frac{1}{\tau}\sum_{c=0,r}^{\tau-1}\sum_{c=0,r-1}^{\tau-1}\left({\mathbf{g}}_j^{(c,r-1)}-{\mathbf{g}}_j^{(r-1)}\right)\Big\|^2+\Big\|\frac{1}{\tau}\sum_{c=0,r}^{\tau-1}\sum_{c=0,r-1}^{\tau-1}\frac{1}{p}\sum_{j=1}^p\left({\mathbf{g}}_{j}^{(c,r-1)}-{\mathbf{g}}_j^{(r-1)}\right)\Big\|^2\nonumber\\
 &=\sum_{c=0,r}^{\tau-1}\sum_{c=0,r-1}^{\tau-1}\Big\|\left({\mathbf{g}}_j^{(r)}-{\mathbf{g}}_j^{(r-1)}\right)\Big\|^2+\Big\|\sum_{c=0,r-1}^{\tau-1}\frac{1}{p}\sum_{j=1}^p\left({\mathbf{g}}_{j}^{(c,r-1)}-{\mathbf{g}}_j^{(r-1)}\right)\Big\|^2,
 \end{align}
as ${\mathbf{g}}_j^{(r)}-{\mathbf{g}}_j^{(r-1)}$ depends on argument in round $r-1$.
\end{proof}

\begin{lemma}\label{lemma:r=0}
For $r=0$, we have:
\begin{align}
    \frac{1}{p}\sum_{j=1}^p\mathbb{E}\left\|\sum_{c=0}^{\tau-1}\tilde{\boldsymbol{d}}^{(c,r)}_j\right\|^2\leq \frac{4}{p}\sum_{j=1}^p\left[\tau\sigma^2+ \tau L^2\sum_{c=0}^{\tau-1}\left\|\boldsymbol{w}^{(c,0)}_j-{\boldsymbol{w}}^{(c,0)}\right\|^2+\left\|\sum_{c=0}^{\tau-1}\left(\mathbf{g}_j^{(c,0)}-\mathbf{g}^{(c,0)}\right)\right\|^2+\tau\sum_{c=0}^{\tau-1}\left\|{\mathbf{g}}^{(0)}\right\|^2\right]
\end{align}
\end{lemma}
\begin{proof}
For $r=0$ we have:
\begin{align}
    \frac{1}{p}&\sum_{j=1}^p\mathbb{E}\left\|\sum_{c=0}^{\tau-1}\tilde{\boldsymbol{d}}^{(c,0)}_j\right\|^2\nonumber\\
    &=\frac{1}{p}\sum_{j=1}^p\mathbb{E}\left\|\sum_{c=0}^{\tau-1}\tilde{\mathbf{g}}^{(c,0)}_j\right\|^2\nonumber\\
    &=\frac{1}{p}\sum_{j=1}^p\mathbb{E}\left\|\sum_{c=0}^{\tau-1}\left(\tilde{\mathbf{g}}^{(c,0)}_j-{\mathbf{g}}^{(c,0)}_j+{\mathbf{g}}^{(c,0)}_j-{\mathbf{g}}^{(0)}_j+{\mathbf{g}}^{(0)}_j-{\mathbf{g}}^{(0)}+{\mathbf{g}}^{(0)}\right)\right\|^2\nonumber\\
    &\leq \frac{4}{p}\sum_{j=1}^p\left[\mathbb{E}\left\|\sum_{c=0}^{\tau-1}\left(\tilde{\mathbf{g}}^{(c,0)}_j-{\mathbf{g}}^{(c,0)}_j\right)\right\|^2+\left\|\sum_{c=0}^{\tau-1}\left(\mathbf{g}^{(c,0)}_j-{\mathbf{g}}^{(0)}_j\right)\right\|^2+\left\|\sum_{c=0}^{\tau-1}\left({\mathbf{g}}^{(0)}_j-{\mathbf{g}}^{(0)}\right)\right\|^2+\left\|\sum_{c=0}^{\tau-1}{\mathbf{g}}^{(0)}\right\|^2\right]\nonumber\\
    &\stackrel{\text{\ding{192}}}{=}\frac{4}{p}\sum_{j=1}^p\left[\sum_{c=0}^{\tau-1}\mathbb{E}\left\|\left(\tilde{\mathbf{g}}^{(c,0)}_j-{\mathbf{g}}^{(c,0)}_j\right)\right\|^2+\left\|\sum_{c=0}^{\tau-1}\left(\mathbf{g}^{(c,0)}_j-{\mathbf{g}}^{(0)}_j\right)\right\|^2+\left\|\sum_{c=0}^{\tau-1}\left({\mathbf{g}}^{(0)}_j-{\mathbf{g}}^{(0)}\right)\right\|^2+\left\|\sum_{c=0}^{\tau-1}{\mathbf{g}}^{(0)}\right\|^2\right]\nonumber\\
    &\leq \frac{4}{p}\sum_{j=1}^p\left[\sum_{c=0}^{\tau-1}\mathbb{E}\left\|\left(\tilde{\mathbf{g}}^{(c,0)}_j-{\mathbf{g}}^{(c,0)}_j\right)\right\|^2+\left\|\sum_{c=0}^{\tau-1}\left(\mathbf{g}^{(c,0)}_j-{\mathbf{g}}^{(0)}_j\right)\right\|^2+\tau\sum_{c=0}^{\tau-1}\left\|\left({\mathbf{g}}^{(0)}_j-{\mathbf{g}}^{(0)}\right)\right\|^2+\tau\sum_{c=0}^{\tau-1}\left\|{\mathbf{g}}^{(0)}\right\|^2\right]\nonumber\\
    &\leq \frac{4}{p}\sum_{j=1}^p\left[\sum_{c=0}^{\tau-1}\sigma^2+\tau\sum_{c=0}^{\tau-1}\left\|\left(\mathbf{g}^{(c,0)}_j-{\mathbf{g}}^{(0)}_j\right)\right\|^2+\left\|\sum_{c=0}^{\tau-1}\left({\mathbf{g}}^{(0)}_j-{\mathbf{g}}^{(0)}\right)\right\|^2+\tau\sum_{c=0}^{\tau-1}\left\|{\mathbf{g}}^{(0)}\right\|^2\right]\nonumber\\
    &=  \frac{4}{p}\sum_{j=1}^p\left[\tau\sigma^2+ \tau L^2\sum_{c=0}^{\tau-1}\left\|\boldsymbol{w}^{(c,0)}_j-{\boldsymbol{w}}^{(c,0)}\right\|^2+\left\|\sum_{c=0}^{\tau-1}\left(\mathbf{g}_j^{(c,0)}-\mathbf{g}^{(0)}\right)\right\|^2+\tau\sum_{c=0}^{\tau-1}\left\|{\mathbf{g}}^{(0)}\right\|^2\right]
\end{align}
where \text{\ding{192}} comes from i.i.d. mini-batch sampling.
\end{proof}

The rest of the proof comes from plugging both Lemmas~\ref{lemma:r>1} and \ref{lemma:r=0} in Eq.~(\ref{eq:I-II}) as shown below. 
\begin{proof}
\begin{align}
\boldsymbol{w}_j^{(c,r)}=\boldsymbol{w}_j^{(c-1,r)}-\eta \tilde{\boldsymbol{d}}_{j}^{(c,r)}=\ldots=\boldsymbol{w}^{(r)}-\eta\sum_{k=0}^{c-1}\tilde{\boldsymbol{d}}_{j}^{(c,r)}    
\end{align}
Now we can write: 
\begin{align}
    &\sum_{r=0}^{R-1}\sum_{c=0,r}^{\tau-1}\frac{1}{p}\sum_{j=1}^p\mathbb{E}\left\|\boldsymbol{w}_j^{(c,r)}-\boldsymbol{w}^{(r)}\right\|^2\nonumber\\
    &=\frac{\eta^2}{p}\sum_{r=0}^{R-1}\sum_{c=0,r}^{\tau-1}\sum_{j=1}^p\mathbb{E}\left\|\sum_{k=0}^{c-1}\tilde{\boldsymbol{d}}_{j}^{(c,r)}\right\|^2\nonumber\\
    &={\eta^2}\left[\sum_{c=0,r=0}^{\tau-1}\frac{1}{p}\sum_{j=1}^p\mathbb{E}\left\|\sum_{k=0}^{c-1}\tilde{\boldsymbol{d}}_{j}^{(c,0)}\right\|^2+\sum_{r=1}^{R-1}\sum_{c=0,r}^{\tau-1}\frac{1}{p}\sum_{j=1}^p\mathbb{E}\left\|\sum_{k=0}^{c-1}\tilde{\boldsymbol{d}}_{j}^{(c,r)}\right\|^2\right]\nonumber\\
    &\leq{\eta^2}\Big(\sum_{c=0,r=0}^{\tau-1}\frac{4}{p}\sum_{j=1}^p\left[\tau\sigma^2+ \tau L^2\sum_{c=0}^{\tau-1}\left\|\boldsymbol{w}^{(c,0)}_j-{\boldsymbol{w}}^{(c,0)}\right\|^2+\left\|\sum_{c=0}^{\tau-1}\mathbf{g}_j^{(c,0)}-\mathbf{g}^{(c,0)}\right\|^2+\tau\sum_{c=0}^{\tau-1}\left\|{\mathbf{g}}^{(0)}\right\|^2\right]\nonumber\\
    &\qquad +\sum_{r=1}^{R-1}\sum_{c=0,r}^{\tau-1}\Big[18\sigma^2 \tau+\frac{1}{p}\sum_{j=1}^p\Big[5L^2\Big[\tau\sum_{c=0,r}^{\tau-1}\Big\|\boldsymbol{w}_{j}^{(c,r)}-{\boldsymbol{w}}^{(r)}\Big\|^2+\sum_{c=0,r}^{\tau-1}\sum_{c=0,r-1}^{\tau-1}\Big\|{\boldsymbol{w}}^{(r)}-{\boldsymbol{w}}^{(r-1)}\Big\|^2\nonumber\\
&\quad+\tau\sum_{c=0,r-1}^{\tau-1}\Big\|{\boldsymbol{w}}^{(r-1)}-{\boldsymbol{w}}_j^{(c,r-1)}\Big\|^2+\tau\sum_{c=0,r-1}^{\tau-1}\frac{1}{p}\sum_{j=1}^p\Big\|{\boldsymbol{w}}_{j}^{(c,r-1)}-{\boldsymbol{w}}^{(r-1)}\Big\|^2\nonumber\\
&\quad+\frac{\tau}{L^2}\sum_{c=0,r-1}^{\tau-1}\Big\|{\mathbf{g}}^{(r-1)}\Big\|^2\Big]\Big]\Big) \nonumber\\
&= {\eta^2}\Big(\Big[\sum_{c=0,r=0}^{\tau-1}\frac{4\tau}{p}\sum_{j=1}^p\sigma^2+ L^2\sum_{c=0,r=0}^{\tau-1}\frac{4\tau}{p}\sum_{j=1}^p\sum_{c=0}^{\tau-1}\left\|\boldsymbol{w}^{(c,0)}_j-{\boldsymbol{w}}^{(c,0)}\right\|^2+\frac{4}{p}\sum_{j=1}^p\sum_{c=0}^{\tau-1}\left\|\sum_{c=0,r=0}^{\tau-1}{\mathbf{g}}_j^{(c,0)}-{\mathbf{g}}^{(0)}\right\|^2\nonumber\\
&\qquad+\sum_{c=0,r=0}^{\tau-1}\frac{4\tau}{p}\sum_{j=1}^p\sum_{c=0}^{\tau-1}\left\|{\mathbf{g}}^{(0)}\right\|^2\Big]\nonumber\\
&\qquad +\Big[18\sigma^2 \tau\sum_{r=1}^{R-1}\sum_{c=0,r}^{\tau-1}1+\Big[5
L^2\Big[\frac{\tau}{p}\sum_{j=1}^p\sum_{r=1}^{R-1}\sum_{c=0,r}^{\tau-1}\Big\|\Big[\boldsymbol{w}_{j}^{(c,r)}-{\boldsymbol{w}}^{(r)}\Big]\Big\|^2+\frac{1}{\tau}\sum_{r=1}^{R-1}\sum_{c=0,r}^{\tau-1}\sum_{c=0,r}^{\tau-1}\sum_{c=0,r-1}^{\tau-1}\Big\|{\boldsymbol{w}}^{(r)}-{\boldsymbol{w}}^{(r-1)}\Big\|^2\nonumber\\
&\quad+\frac{\tau}{p}\sum_{j=1}^p\sum_{r=1}^{R-1}\sum_{c=0,r}^{\tau-1}\sum_{c=0,r-1}^{\tau-1}\Big\|{\boldsymbol{w}}^{(r-1)}-{\boldsymbol{w}}_j^{(c,r-1)}\Big\|^2+\frac{\tau}{p}\sum_{j=1}^p\sum_{r=1}^{R-1}\sum_{c=0,r}^{\tau-1}\sum_{c=0,r-1}^{\tau-1}\Big\|{\boldsymbol{w}}_{j}^{(c,r-1)}-{\boldsymbol{w}}^{(r-1)}\Big\|^2\nonumber\\
&\quad+\tau^2\sum_{r=1}^{R-1}\frac{1}{L^2}\sum_{c=0,r-1}^{\tau-1}\Big\|{\mathbf{g}}^{(r-1)}\Big\|^2\Big]\Big]\Big) \nonumber\\
&={\eta^2}\Big(\Big[4\tau^2\sigma^2+ L^2\frac{4\tau^2}{p}\sum_{j=1}^p\sum_{c=0}^{\tau-1}\left\|\boldsymbol{w}^{(c,0)}_j-{\boldsymbol{w}}^{(c,0)}\right\|^2+\frac{4}{p}\sum_{j=1}^p\sum_{c=0}^{\tau-1}\left\|\sum_{c=0,r=0}^{\tau-1}{\mathbf{g}}_j^{(c,0)}-{\mathbf{g}}^{(0)}\right\|^2+4\tau^2\sum_{c=0,r=0}^{\tau-1}\left\|{\mathbf{g}}^{(0)}\right\|^2\Big]\nonumber\\
&\qquad +\Big[18\sigma^2 (R-1)\tau^2+\Big[5
L^2\Big[\frac{\tau^2}{p}\sum_{j=1}^p\sum_{r=1}^{R-1}\sum_{c=0,r}^{\tau-1}\Big\|\Big[\boldsymbol{w}_{j}^{(c,r)}-{\boldsymbol{w}}^{(r)}\Big]\Big\|^2+\tau\sum_{r=1}^{R-1}\sum_{c=0,r}^{\tau-1}\sum_{c=0,r-1}^{\tau-1}\Big\|{\boldsymbol{w}}^{(r)}-{\boldsymbol{w}}^{(r-1)}\Big\|^2\nonumber\\
&\qquad+\frac{\tau^2}{p}\sum_{j=1}^p\sum_{r=1}^{R-1}\sum_{c=0,r-1}^{\tau-1}\Big\|{\boldsymbol{w}}^{(r-1)}-{\boldsymbol{w}}_j^{(c,r-1)}\Big\|^2+\frac{\tau^2}{p}\sum_{j=1}^p\sum_{r=1}^{R-1}\sum_{c=0,r-1}^{\tau-1}\Big\|{\boldsymbol{w}}_{j}^{(c,r-1)}-{\boldsymbol{w}}^{(r-1)}\Big\|^2\nonumber\\
&\qquad+\tau^2\sum_{r=1}^{R-1}\frac{1}{L^2}\sum_{c=0,r-1}^{\tau-1}\Big\|{\mathbf{g}}^{(r-1)}\Big\|^2\Big]\Big]\Big)\label{eq:-0} 
\end{align}

Now we continue with bounding Eq.~(\ref{eq:-0}) with further simplification as follows:
\begin{align}
    &={\eta^2}\Big(\Big[4\tau^2\sigma^2+ L^2\frac{4\tau^2}{p}\sum_{j=1}^p\sum_{c=0}^{\tau-1}\left\|\boldsymbol{w}^{(c,0)}_j-{\boldsymbol{w}}^{(c,0)}\right\|^2+\frac{4}{p}\sum_{j=1}^p\sum_{c=0}^{\tau-1}\left\|\sum_{c=0}^{\tau-1}\left({\mathbf{g}}_j^{(c,0)}-{\mathbf{g}}^{(0)}\right)\right\|^2+4\tau^2\sum_{c=0,r=0}^{\tau-1}\left\|{\mathbf{g}}^{(0)}\right\|^2\Big]\nonumber\\
&\quad +18\sigma^2 (R-1)\tau^2+\frac{5
L^2\tau^2}{p}\sum_{j=1}^p\sum_{r=1}^{R-1}\sum_{c=0,r}^{\tau-1}\Big\|\boldsymbol{w}_{j}^{(c,r)}-{\boldsymbol{w}}^{(r)}\Big\|^2+{5
L^2\tau}\sum_{r=1}^{R-1}\sum_{c=0,r}^{\tau-1}\sum_{c=0,r-1}^{\tau-1}\Big\|{\boldsymbol{w}}^{(r)}-{\boldsymbol{w}}^{(r-1)}\Big\|^2\nonumber\\
&\quad+\frac{10
L^2\tau^2}{p}\sum_{j=1}^p\sum_{r=1}^{R-1}\sum_{c=0,r-1}^{\tau-1}\Big\|{\boldsymbol{w}}^{(r-1)}-{\boldsymbol{w}}_j^{(c,r-1)}\Big\|^2+{5
\tau^2}\sum_{r=1}^{R-1}\sum_{c=0,r-1}^{\tau-1}\Big\|{\mathbf{g}}^{(r-1)}\Big\|^2\Big) \nonumber\\
 &\stackrel{\text{\ding{192}}}{\leq}{\eta^2}\Big(\Big[18R\tau^2\sigma^2+ L^2\frac{4\tau^2}{p}\sum_{j=1}^p\sum_{c=0}^{\tau-1}\left\|\boldsymbol{w}^{(c,0)}_j-{\boldsymbol{w}}^{(c,0)}\right\|^2+\frac{4}{p}\sum_{j=1}^p\sum_{c=0}^{\tau-1}\left\|\sum_{c=0,r=0}^{\tau-1}{\mathbf{g}}_j^{(c,0)}-{\mathbf{g}}^{(0)}\right\|^2+4\tau^2\sum_{c=0,r=0}^{\tau-1}\left\|{\mathbf{g}}^{(0)}\right\|^2\Big]\nonumber\\
&\quad +\frac{5
L^2\tau^2}{p}\sum_{j=1}^p\sum_{r=1}^{R-1}\sum_{c=0,r}^{\tau-1}\Big\|\Big[\boldsymbol{w}_{j}^{(c,r)}-{\boldsymbol{w}}^{(r)}\Big]\Big\|^2+\frac{5
L^2\tau}{p}\sum_{j=1}^p\sum_{r=1}^{R-1}\sum_{c=0,r}^{\tau-1}\sum_{c=0,r-1}^{\tau-1}\Big\|{\boldsymbol{w}}^{(r)}-{\boldsymbol{w}}^{(r-1)}\Big\|^2\nonumber\\
&\quad+\frac{10
L^2\tau^2}{p}\sum_{j=1}^p\sum_{r=1}^{R-1}\sum_{c=0,r-1}^{\tau-1}\Big\|{\boldsymbol{w}}_{j}^{(c,r-1)}-{\boldsymbol{w}}^{(r-1)}\Big\|^2+{5
\tau^2}\sum_{r=1}^{R-1}\sum_{c=0,r-1}^{\tau-1}\Big\|{\mathbf{g}}^{(r-1)}\Big\|^2\Big) \nonumber\\
 &\stackrel{\text{\ding{193}}}{\leq}{\eta^2}\Big(\Big[18R\tau^2\sigma^2+ \frac{4}{p}\sum_{j=1}^p\sum_{c=0}^{\tau-1}\left\|\sum_{c=0,r=0}^{\tau-1}{\mathbf{g}}_j^{(c,0)}-{\mathbf{g}}^{(0)}\right\|^2+5\tau^2\sum_{c=0,r=0}^{\tau-1}\left\|{\mathbf{g}}^{(0)}\right\|^2\Big]\nonumber\\
&\quad +\frac{5
L^2\tau^2}{p}\sum_{j=1}^p\sum_{r=0}^{R-1}\sum_{c=0,r}^{\tau-1}\Big\|\Big[\boldsymbol{w}_{j}^{(c,r)}-{\boldsymbol{w}}^{(r)}\Big]\Big\|^2+\frac{5
L^2\tau}{p}\sum_{j=1}^p\sum_{r=0}^{R-1}\sum_{c=0,r}^{\tau-1}\sum_{c=0,r-1}^{\tau-1}\Big\|{\boldsymbol{w}}^{(r)}-{\boldsymbol{w}}^{(r-1)}\Big\|^2\nonumber\\
&\quad+\frac{10
L^2\tau^2}{p}\sum_{j=1}^p\sum_{r=0}^{R-1}\sum_{c=0,r}^{\tau-1}\Big\|{\boldsymbol{w}}_{j}^{(c,r)}-{\boldsymbol{w}}^{(r)}\Big\|^2+{5
\tau^2}\sum_{r=1}^{R-1}\sum_{c=0,r-1}^{\tau-1}\Big\|{\mathbf{g}}^{(r-1)}\Big\|^2\Big) \nonumber\\
&\stackrel{\text{\ding{194}}}{\leq}18R\eta^2\tau^2\sigma^2+ \frac{4\eta^2}{p}\sum_{j=1}^p\sum_{c=0}^{\tau-1}\left\|\sum_{c=0,r=0}^{\tau-1}{\mathbf{g}}_j^{(c,0)}-{\mathbf{g}}^{(0)}\right\|^2 +\frac{5\eta^2
L^2\tau}{p}\sum_{j=1}^p\sum_{r=0}^{R-1}\sum_{c=0,r}^{\tau-1}\sum_{c=0,r-1}^{\tau-1}\Big\|{\boldsymbol{w}}^{(r)}-{\boldsymbol{w}}^{(r-1)}\Big\|^2\nonumber\\
&\quad+\frac{15\eta^2
L^2\tau^2}{p}\sum_{j=1}^p\sum_{r=0}^{R-1}\sum_{c=0,r}^{\tau-1}\Big\|{\boldsymbol{w}}_{j}^{(c,r)}-{\boldsymbol{w}}^{(r)}\Big\|^2+{10\eta^2
\tau^2}\sum_{r=0}^{R-1}\sum_{c=0}^{\tau-1}\Big\|{\mathbf{g}}^{(r)}\Big\|^2 \label{eq:last-stp-to-bnd-drift}
\end{align}
where \text{\ding{192}} comes from $4\tau^2\sigma^2\leq 18\tau^2\sigma^2$, \text{\ding{193}} holds because of $4\tau^2\sum_{c=0,r=0}^{\tau-1}\left\|{\mathbf{g}}^{(0)}\right\|^2\leq 5\tau^2\sum_{c=0,r=0}^{\tau-1}\left\|{\mathbf{g}}^{(0)}\right\|^2$ and \text{\ding{194}} is due to 
$$5\eta^2\tau^2\sum_{c=0,r=0}^{\tau-1}\left\|{\mathbf{g}}^{(0)}\right\|^2+{5\eta^2
\tau^2}\sum_{r=1}^{R-1}\sum_{c=0,r-1}^{\tau-1}\Big\|{\mathbf{g}}^{(r-1)}\Big\|^2\leq {10\eta^2
\tau^2}\sum_{r=0}^{R-1}\sum_{c=0}^{\tau-1}\Big\|{\mathbf{g}}^{(r)}\Big\|^2.$$
Rearranging Eq.~(\ref{eq:last-stp-to-bnd-drift}) we obtain: 

\begin{align}
    \sum_{r=0}^{R-1}&\sum_{c=0,r}^{\tau-1}\frac{1}{p}\sum_{j=1}^p\mathbb{E}\left\|\boldsymbol{w}_j^{(c,r)}-\boldsymbol{w}^{(r)}\right\|^2\nonumber\\ 
    &\leq\frac{18R\eta^2\tau^2\sigma^2}{1-15\eta^2
L^2\tau^2}+ \frac{4\eta^2}{p\left(1-15\eta^2
L^2\tau^2\right)}\sum_{j=1}^p\sum_{c=0}^{\tau-1}\left\|\sum_{c=0,r=0}^{\tau-1}{\mathbf{g}}_j^{(c,0)}-{\mathbf{g}}^{(0)}\right\|^2 \nonumber\\
&+\frac{5\eta^2
L^2\tau}{p\left(1-15\eta^2
L^2\tau^2\right)}\sum_{j=1}^p\sum_{r=0}^{R-1}\sum_{c=0,r}^{\tau-1}\sum_{c=0,r-1}^{\tau-1}\Big\|{\boldsymbol{w}}^{(r)}-{\boldsymbol{w}}^{(r-1)}\Big\|^2\nonumber\\
&\quad+\frac{10\eta^2
\tau^2}{\left(1-15\eta^2
L^2\tau^2\right)}\sum_{r=0}^{R-1}\sum_{c=0}^{\tau-1}\Big\|{\mathbf{g}}^{(r)}\Big\|^2\nonumber\\
&\stackrel{\text{\ding{192}}}{\leq}{36R\eta^2\tau^2\sigma^2}+ \frac{36\eta^2}{p}\sum_{j=1}^p\sum_{c=0}^{\tau-1}\left\|\sum_{c=0}^{\tau-1}\left({\mathbf{g}}_j^{(c,0)}-{\mathbf{g}}^{(0)}\right)\right\|^2 \nonumber\\
&+\frac{10\eta^2
L^2\tau}{p}\sum_{j=1}^p\sum_{r=0}^{R-1}\sum_{c=0,r}^{\tau-1}\sum_{c=0,r-1}^{\tau-1}\Big\|{\boldsymbol{w}}^{(r)}-{\boldsymbol{w}}^{(r-1)}\Big\|^2\nonumber\\
&\quad+{20\eta^2
\tau^2}\sum_{r=0}^{R-1}\sum_{c=0}^{\tau-1}\Big\|{\mathbf{g}}^{(r)}\Big\|^2\label{eq:bndng-drift-ph1} 
\end{align}
where \text{\ding{192}} comes from the condition $1\geq 30\eta^2.
L^2\tau^2$
\end{proof}
\begin{lemma}\label{lemma:D.6}
Under Assumptions~\ref{Assu:1}, \ref{Assu:09}, \ref{Assu:2} and \ref{assum:009} we have: 
\begin{align}
    \frac{1}{p}\sum_{j=1}^p\sum_{r=0}^{R-1}\sum_{c=0,r}^{\tau-1}\sum_{c=0,r-1}^{\tau-1}\mathbb{E}_{\xi}\mathbb{E}_{Q}\Big\|{\boldsymbol{w}}^{(r)}-{\boldsymbol{w}}^{(r-1)}\Big\|^2&\leq \tau^3(\eta\gamma)^2(q+1)\sum_{r=1}^{R-1}\sum_{c=0,r-1}^{\tau-1}\Big[\Big\|\frac{1}{p}\sum_{j=1}^p{\mathbf{g}}_j^{(c,r-1)}\Big\|^2\Big]\nonumber\\
    &\quad+{\tau^3}R(\eta\gamma)^2(q+1)\frac{\sigma^2}{p}+\tau^2(\eta\gamma)^2(q+1)RG_q
\end{align}
\end{lemma}
\begin{proof}
\begin{align}
    \frac{1}{p}&\sum_{j=1}^p\sum_{r=0}^{R-1}\sum_{c=0,r}^{\tau-1}\sum_{c=0,r-1}^{\tau-1}\mathbb{E}_{\xi}\mathbb{E}_{Q}\Big\|{\boldsymbol{w}}^{(r)}-{\boldsymbol{w}}^{(r-1)}\Big\|^2\nonumber\\
    &=\sum_{r=0}^{R-1}\sum_{c=0,r}^{\tau-1}\sum_{c=0,r-1}^{\tau-1}\mathbb{E}_{\xi}\mathbb{E}_{Q}\Big\|{\boldsymbol{w}}^{(r)}-{\boldsymbol{w}}^{(r-1)}\Big\|^2\nonumber\\
    &=\tau\sum_{r=1}^{R-1}\sum_{c=0,r-1}^{\tau-1}\mathbb{E}_{\xi}\mathbb{E}_{Q}\Big\|{\boldsymbol{w}}^{(r)}-{\boldsymbol{w}}^{(r-1)}\Big\|^2\nonumber\\
    &=\tau(\eta\gamma)^2\sum_{r=1}^{R-1}\sum_{c=0,r-1}^{\tau-1}\mathbb{E}_{\xi}\mathbb{E}_{Q}\Big\|\frac{1}{p}\sum_{j=1}^pQ\left(\sum_{c=0,r-1}^{\tau-1}\tilde{\boldsymbol{d}}_j^{(c,r-1)}\right)\Big\|^2\nonumber\\
    &\stackrel{\text{\ding{192}}}{\leq}\tau(\eta\gamma)^2\sum_{r=1}^{R-1}\sum_{c=0,r-1}^{\tau-1}\mathbb{E}_{\xi}\left[\mathbb{E}_{Q}\Big\|Q\left(\frac{1}{p}\sum_{j=1}^p\sum_{c=0,r-1}^{\tau-1}\tilde{\boldsymbol{d}}_j^{(c,r-1)}\right)\Big\|^2+G_q\right]\nonumber\\
    &\stackrel{\text{\ding{193}}}{\leq}\tau(\eta\gamma)^2\sum_{r=1}^{R-1}\sum_{c=0,r-1}^{\tau-1}\mathbb{E}_{\xi}\Big[\mathbb{E}_{Q}\Big\|Q\left(\frac{1}{p}\sum_{j=1}^p\sum_{c=0,r-1}^{\tau-1}\tilde{\boldsymbol{d}}_j^{(c,r-1)}\right)-\mathbb{E}_{Q}\left[Q\left(\frac{1}{p}\sum_{j=1}^p\sum_{c=0,r-1}^{\tau-1}\tilde{\boldsymbol{d}}_j^{(c,r-1)}\right)\right]\Big\|^2\nonumber\\
    &\quad+\left\|\mathbb{E}_{Q}\left[Q\left(\frac{1}{p}\sum_{j=1}^p\sum_{c=0,r-1}^{\tau-1}\tilde{\boldsymbol{d}}_j^{(c,r-1)}\right)\right]\right\|^2+G_q\Big]\nonumber\\
    &=\tau(\eta\gamma)^2\sum_{r=1}^{R-1}\sum_{c=0,r-1}^{\tau-1}\mathbb{E}_{\xi}\Big[\mathbb{E}_{Q}\Big\|Q\left(\frac{1}{p}\sum_{j=1}^p\sum_{c=0,r-1}^{\tau-1}\tilde{\boldsymbol{d}}_j^{(c,r-1)}\right)-\left[\frac{1}{p}\sum_{j=1}^p\sum_{c=0,r-1}^{\tau-1}\tilde{\boldsymbol{d}}_j^{(c,r-1)}\right]\Big\|^2\nonumber\\
    &\quad+\left\|\left[\frac{1}{p}\sum_{j=1}^p\sum_{c=0,r-1}^{\tau-1}\tilde{\boldsymbol{d}}_j^{(c,r-1)}\right]\right\|^2+G_q\Big]\nonumber\\
    &\stackrel{\text{\ding{194}}}{\leq}\tau(\eta\gamma)^2\sum_{r=1}^{R-1}\sum_{c=0,r-1}^{\tau-1}\mathbb{E}_{\xi}\Big[q\Big\|\left[\frac{1}{p}\sum_{j=1}^p\sum_{c=0,r-1}^{\tau-1}\tilde{\boldsymbol{d}}_j^{(c,r-1)}\right]\Big\|^2+\left\|\left[\frac{1}{p}\sum_{j=1}^p\sum_{c=0,r-1}^{\tau-1}\tilde{\boldsymbol{d}}_j^{(c,r-1)}\right]\right\|^2+G_q\Big]\nonumber\\
    &=\tau(\eta\gamma)^2\sum_{r=1}^{R-1}\sum_{c=0,r-1}^{\tau-1}(q+1)\mathbb{E}_{\xi}\Big[\Big\|\left[\frac{1}{p}\sum_{j=1}^p\sum_{c=0,r-1}^{\tau-1}\tilde{\boldsymbol{d}}_j^{(c,r-1)}\right]\Big\|^2+G_q\Big]\nonumber\\
    &=\tau^2(\eta\gamma)^2(q+1)\sum_{r=1}^{R-1}\mathbb{E}_{\xi}\Big[\Big\|\left[\frac{1}{p}\sum_{j=1}^p\sum_{c=0,r-1}^{\tau-1}\tilde{\boldsymbol{d}}_j^{(c,r-1)}\right]\Big\|^2+G_q\Big]\nonumber\\
    &\stackrel{\text{\ding{195}}}{=}\tau^2(\eta\gamma)^2(q+1)\sum_{r=1}^{R-1}\mathbb{E}_{\xi}\Big[\Big\|\left[\frac{1}{p}\sum_{j=1}^p\sum_{c=0,r-1}^{\tau-1}\tilde{\mathbf{g}}_j^{(c,r-1)}\right]\Big\|^2+G_q\Big]\label{eq:mid-sdli}
    \end{align}
    where \text{\ding{192}} comes from Assumption~\ref{assum:009}, \text{\ding{193}} is due to the definition of variance, \text{\ding{194}} holds because of Assumption~\ref{Assu:09} and \text{\ding{195}} is because of $\frac{1}{p}\sum_{j=1}^p\delta^{(r,\tau)}=0$.
    We continue from Eq.~(\ref{eq:mid-sdli}) as follows:
    \begin{align}
    &=\tau^2(\eta\gamma)^2(q+1)\sum_{r=1}^{R-1}\mathbb{E}_{\xi}\Big[\Big\|\left[\frac{1}{p}\sum_{j=1}^p\sum_{c=0,r-1}^{\tau-1}\tilde{\mathbf{g}}_j^{(c,r-1)}\right]\Big\|^2\Big]+\tau^2(\eta\gamma)^2(q+1)RG_q\nonumber\\
    &=\tau^2(\eta\gamma)^2(q+1)\sum_{r=1}^{R-1}\text{Var}_{\xi}\Big(\left[\frac{1}{p}\sum_{j=1}^p\sum_{c=0,r-1}^{\tau-1}\tilde{\mathbf{g}}_j^{(c,r-1)}\right]\Big)+\tau^2(\eta\gamma)^2(q+1)\sum_{r=1}^{R-1}\left\|\mathbb{E}_{\xi}\frac{1}{p}\sum_{j=1}^p\sum_{c=0,r-1}^{\tau-1}\tilde{\mathbf{g}}_j^{(c,r-1)}\right\|^2\nonumber\\
    &\qquad\qquad+\tau^2(\eta\gamma)^2(q+1)RG_q\nonumber\\
    &=\tau^2(\eta\gamma)^2(q+1)\sum_{r=1}^{R-1}\text{Var}_{\xi}\Big(\left[\frac{1}{p}\sum_{j=1}^p\sum_{c=0,r-1}^{\tau-1}\tilde{\mathbf{g}}_j^{(c,r-1)}\right]\Big)+\tau^2(\eta\gamma)^2(q+1)\sum_{r=1}^{R-1}\left\|\frac{1}{p}\sum_{j=1}^p\sum_{c=0,r-1}^{\tau-1}{\mathbf{g}}_j^{(c,r-1)}\right\|^2\nonumber\\
    &\qquad\qquad+\tau^2(\eta\gamma)^2(q+1)RG_q\nonumber\\
    &\stackrel{\text{\ding{192}}}{=}\tau^2(\eta\gamma)^2(q+1)\sum_{r=1}^{R-1}\frac{1}{p^2}\sum_{j=1}^p\sum_{c=0,r-1}^{\tau-1}\text{Var}_{\xi}\Big(\left[\tilde{\mathbf{g}}_j^{(c,r-1)}\right]\Big)+\tau^2(\eta\gamma)^2(q+1)\sum_{r=1}^{R-1}\left\|\frac{1}{p}\sum_{j=1}^p\sum_{c=0,r-1}^{\tau-1}{\mathbf{g}}_j^{(c,r-1)}\right\|^2\nonumber\\
    &\qquad\qquad+\tau^2(\eta\gamma)^2(q+1)RG_q\nonumber\\
     &\stackrel{\text{\ding{193}}}{\leq} \tau^2(\eta\gamma)^2(q+1)\sum_{r=1}^{R-1}\frac{1}{p^2}\sum_{j=1}^p\sum_{c=0,r-1}^{\tau-1}\sigma^2+\tau^3(\eta\gamma)^2(q+1)\sum_{r=1}^{R-1}\frac{1}{p}\sum_{j=1}^p\sum_{c=0,r-1}^{\tau-1}\left\|{\mathbf{g}}_j^{(c,r-1)}\right\|^2+\tau^2(\eta\gamma)^2(q+1)RG_q\nonumber\\
     &=\tau^3(\eta\gamma)^2(q+1)\sum_{r=1}^{R-1}\frac{1}{p}\sum_{j=1}^p\sum_{c=0}^{\tau-1}\left\|{\mathbf{g}}_j^{(c,r-1)}\right\|^2+\tau^3(\eta\gamma)^2(q+1)R\frac{1}{p}\sigma^2+\tau^2(\eta\gamma)^2(q+1)RG_q
     \end{align}
     where \text{\ding{192}} comes from i.i.d. mini-batch sampling and \text{\ding{193}} is due to inequlity $\|\sum_{i=1}^n\mathbf{a}_i\|^2\leq n\sum_{i=1}^n\|\mathbf{a}_i\|^2$.
\end{proof}

Finally, by plugging Lemma~\ref{lemma:D.6} into Eq.~(\ref{eq:bndng-drift-ph1}), we obtain the following bound: 

\begin{align}
    \sum_{r=0}^{R-1}&\sum_{c=0,r}^{\tau-1}\frac{1}{p}\sum_{j=1}^p\mathbb{E}\left\|\boldsymbol{w}_j^{(c,r)}-\boldsymbol{w}^{(r)}\right\|^2\nonumber\\ 
    &\leq{36R\eta^2\tau^2\sigma^2}+ \frac{8\eta^2}{p}\sum_{j=1}^p\sum_{c=0}^{\tau-1}\left\|\sum_{c=0,r=0}^{\tau-1}{\mathbf{g}}_j^{(c,0)}-{\mathbf{g}}^{(0)}\right\|^2 \nonumber\\
&\quad+{10\eta^2
L^2\tau}\left[\tau^3(\eta\gamma)^2(q+1)\sum_{r=1}^{R-1}\sum_{c=0,r-1}^{\tau-1}\Big[\Big\|\frac{1}{p}\sum_{j=1}^p{\mathbf{g}}_j^{(c,r-1)}\Big\|^2\Big]+\tau^3R(\eta\gamma)^2(q+1)\frac{\sigma^2}{p}+\tau^2(\eta\gamma)^2(q+1)RG_q\right]\nonumber\\
&\quad+{20\eta^2
\tau^2}\sum_{r=0}^{R-1}\sum_{c=0}^{\tau-1}\Big\|{\mathbf{g}}^{(r)}\Big\|^2\nonumber\\
&={36R\eta^2\tau^2\sigma^2}+ \frac{8\eta^2}{p}\sum_{j=1}^p\sum_{c=0}^{\tau-1}\left\|\sum_{c=0,r=0}^{\tau-1}\left({\mathbf{g}}_j^{(c,0)}-{\mathbf{g}}^{(0)}\right)\right\|^2 \nonumber\\
&\quad+\Big[10\eta^2
L^2\tau^4(\eta\gamma)^2(q+1)\sum_{r=1}^{R-1}\sum_{c=0,r-1}^{\tau-1}\Big[\Big\|\frac{1}{p}\sum_{j=1}^p{\mathbf{g}}_j^{(c,r-1)}\Big\|^2\Big]\nonumber\\
&\quad+10\eta^2
L^2\tau^4R(\eta\gamma)^2(q+1)\frac{\sigma^2}{p}+{10\eta^2
L^2}\tau^3(\eta\gamma)^2(q+1)RG_q\Big]\nonumber\\
&\quad+{20\eta^2
\tau^2}\sum_{r=0}^{R-1}\sum_{c=0}^{\tau-1}\Big\|{\mathbf{g}}^{(r)}\Big\|^2
\end{align}

\newpage
\subsection{Proof of Lemma~\ref{lemm:pl-main-lemm}}
Similarly, using Lemmas~\ref{lem:D2} and \ref{lemma:r=0} for every communication round we can write:
\begin{align}
   & \frac{1}{p}\sum_{j=1}^p\sum_{c=0}^{\tau-1}\mathbb{E}\left\|\boldsymbol{w}_j^{(r,c)}-\boldsymbol{w}^{(r)}\right\|^2\nonumber\\
   &
   \leq {36\eta^2\tau^2\sigma^2}+ \frac{8\eta^2}{p}\sum_{j=1}^p\sum_{c=0}^{\tau-1}\left\|\sum_{c=0}^{\tau-1}\left({\mathbf{g}}_j^{(c,0)}-{\mathbf{g}}^{(0)}\right)\right\|^2 \nonumber\\
&+{10\eta^2
L^2\tau}\sum_{c=0,r}^{\tau-1}\sum_{c=0,r-1}^{\tau-1}\Big\|{\boldsymbol{w}}^{(r)}-{\boldsymbol{w}}^{(r-1)}\Big\|^2+{20\eta^2
\tau^2}\sum_{c=0}^{\tau-1}\Big\|{\mathbf{g}}^{(r)}\Big\|^2\nonumber\\
&\stackrel{\text{\ding{192}}}{\leq} {36\eta^2\tau^2\sigma^2}+ \frac{8\eta^2}{p}\sum_{j=1}^p\sum_{c=0}^{\tau-1}\left\|\sum_{c=0}^{\tau-1}\left({\mathbf{g}}_j^{(c,0)}-{\mathbf{g}}^{(0)}\right)\right\|^2 \nonumber\\
&+\Big[{10\eta^2
L^2}\tau^4(\eta\gamma)^2(q+1)\sum_{c=0,r-1}^{\tau-1}\Big[\Big\|\frac{1}{p}\sum_{j=1}^p{\mathbf{g}}_j^{(c,r-1)}\Big\|^2\Big]\nonumber\\
    &\quad+{10\eta^2
L^2}\tau^4(\eta\gamma)^2(q+1)\frac{\sigma^2}{p}+{10\eta^2
L^2}\tau^3(\eta\gamma)^2(q+1)G_q\Big]+{20\eta^2
\tau^2}\sum_{c=0}^{\tau-1}\Big\|{\mathbf{g}}^{(r)}\Big\|^2
\end{align}
where \text{\ding{192}} follows from Lemma~\ref{lemma:D.6} without summation over $r$.

\bibliographystyle{plain}
\bibliography{ref}

\end{document}